\documentclass[sigconf]{acmart}

\setcopyright{none} 
\acmYear{2021}

\renewcommand\footnotetextcopyrightpermission[1]{} 

\usepackage{amsmath}
\usepackage{amsfonts}
\usepackage{amsthm}
\usepackage{mathtools}
\usepackage[caption=false]{subfig}
\usepackage{algorithm}
\usepackage{algpseudocode}
\usepackage{multirow}
\usepackage{enumitem}
\usepackage{shortcuts}
\newtheorem{theorem}{Theorem}
\newtheorem{assumption}{Assumption}
\newtheorem{lemma}{Lemma}
\newtheorem{claim}{Claim}

\usepackage[title]{appendix}

\makeatletter
\newcommand{\mathleft}{\@fleqntrue\@mathmargin0pt}
\newcommand{\mathcenter}{\@fleqnfalse}
\makeatother

\begin{document}
\title{\codename: Privacy-preserving Optimization  for  Federated Learning} 

\author{Yanjun Zhang}
\orcid{0001-5611-3483}
\affiliation{%
	\institution{The University of Queensland}
	\city{St Lucia}
	\state{Queensland}
	\country{Australia}
	\postcode{4067}
}
\email{yanjun.zhang@uq.edu.au}

\author{Guangdong Bai} 
\authornote{The corresponding author.}
\affiliation{%
	\institution{The University of Queensland}
	\city{St Lucia}
	\state{Queensland}
	\country{Australia}
	\postcode{4067}
}
\email{g.bai@uq.edu.au}

\author{Xue Li}
\affiliation{%
	\institution{The University of Queensland}
	\city{St Lucia}
	\state{Queensland}
	\country{Australia}
	\postcode{4067}
}
\email{xueli@itee.uq.edu.au}

\author{Surya Nepal}
\affiliation{%
	\institution{Data61 CSIRO Australia}
	\city{MARSFIELD}
	\state{New South Wales}
	\country{Australia}
	\postcode{2122}
}
\email{Surya.Nepal@data61.csiro.au}

\author{Ryan K L Ko}
\affiliation{%
	\institution{The University of Queensland}
	\city{St Lucia}
	\state{Queensland}
	\country{Australia}
	\postcode{4067}
}
\email{ryan.ko@uq.edu.au}

\begin{abstract}
Federated learning enables multiple participants to collaboratively train a model without aggregating the training data.
Although the training data are kept within each participant and the local gradients can be securely synthesized, 
recent studies have shown that such privacy protection is insufficient.
The global model parameters that have to be shared for optimization are susceptible to leak  information about training data. 
In this work, we propose \codename (\short) that enhances privacy of federated learning by eliminating the sharing of global model parameters.
\short exploits the fact that a gradient descent optimization can start with a set of discrete 
points and converges to another set at the neighborhood of the global minimum of the objective function.
It lets the participants independently train on their local data, and securely share the sum of local gradients to benefit each other.
We formally demonstrate \short’s privacy enhancement over traditional FL. We prove that less information is exposed in CGD compared to that of traditional FL. 
\short also guarantees desired model accuracy. We theoretically establish a convergence rate for \short. We prove that the loss of the proprietary models learned for each participant against a model learned by aggregated training data is bounded. 
Extensive experimental results on two real-world datasets demonstrate the performance of \short is comparable with the centralized learning, 
with marginal differences on validation loss~(mostly within  0.05) and  accuracy~(mostly within  1\%). 
\end{abstract}

\maketitle
\pagestyle{plain} 

\section{Introduction}
The performance of machine learning largely relies on the availability of large representative datasets.
To take advantage of massive data owned by multiple entities, federated learning (FL) is proposed~\cite{konevcny2016federated, yang2019federated, konevcny2016federatedo}.
It enables participants to jointly train a global model without the necessity of sharing their datasets, demonstrating the potential to address the issues of data privacy and data ownership. 
It has been incorporated by popular machine learning tools such as TensorFlow~\cite{abadi2016tensorflow} and PyTorch~\cite{paszke2019pytorch}, 
and increasingly spread over various industries.

The privacy preservation of FL stems from its parallelization of the gradient descent optimization, which in essence is an application of stochastic gradient descent (SGD)~(or 
the mini-batch mode)~\cite{konevcny2016federated}.
During the training process, the participants work on the same intermediate global model via a coordinating server~(in the centralized FL)~\cite{bonawitz2017practical, mohassel2017secureml, abadi2016deep} or a peer-to-peer communication scheme~(in the decentralized FL)~\cite{kim2019blockchained, roy2019braintorrent}.
Each of them obtains the current model parameters, works out a local gradient based on the local data, and
disseminates it to update the global model synchronously~\cite{abadi2016deep,bonawitz2017practical} or asynchronously~\cite{hu2019fdml}.
This paradigm guarantees data locality, but has been found insufficient for data privacy:
although the local gradients can be securely synthesized via a variety of techniques such as differential privacy (DP)~\cite{abadi2016deep, shokri2015privacy,hu2019fdml,zhang2020differentially}, secure multi-party communication (MPC)~\cite{bonawitz2017practical, mohassel2017secureml,gascon2017privacy}, and homomorphic encryption (HE)~\cite{mohassel2017secureml, marc2019privacy, sharma2019confidential}, 
the global model parameters that have to be shared are still susceptible to information leakage(cf. Section \ref{sec:privacy} and~\cite{papernot2018sok, nasr2019comprehensive}). 

This work further decreases the dependency among 
participants by 
\emph{eliminating the explicit sharing of the central global model} which is the root cause of the information leakage~\cite{fredrikson2015model, nasr2019comprehensive}.  
We propose a new optimization algorithm named \codename (\short) that enables each participant to learn a proprietary global model.
The \short participants maintain 
their global models locally, which are strictly confined within 
themselves from the  beginning of and throughout the whole training process. 
We refer to these localized global models as \emph{confined models}, to distinguish them from the global model in traditional FL.

\short 
is inspired by an observation on the surface of the typical cost function.
The steepness of the first derivative decreases slower when approaching the minimum of the function, due to the small values in the Hessian (i.e.,  the second derivative) near the optimum~\cite{bottou2012stochastic}. 
This gives the function, when plotted, a flat valley bottom.
As such, a gradient descent algorithm $\mathcal{A}$, when applied on an objective function $F$, could start with a set of discrete points (referred to as a \emph{colony} and their distance is discussed later). 
Iteratively descending the colony using the joint gradient of the colony would lead $\mathcal{A}$ to the neighborhood of  $F$'s  minimum  in the ``flat valley bottom''.
The points in the colony would also end up with similar losses that are close to the loss of the minimum. 

\begin{figure*}[!t]
	\centering
	\subfloat[Gradient descent in traditional federated learning. Participants jointly work on the same global model $w$ using the descent computed by $f$. 
	Although the local gradients $g^a, g^b, g^c$ can be securely synthesized, by knowing $w$ and $f$, the adversary is able to derive information about the local raw data $\xi_a, \xi_b, \xi_c$.
	 ]{{\includegraphics[width=7cm]{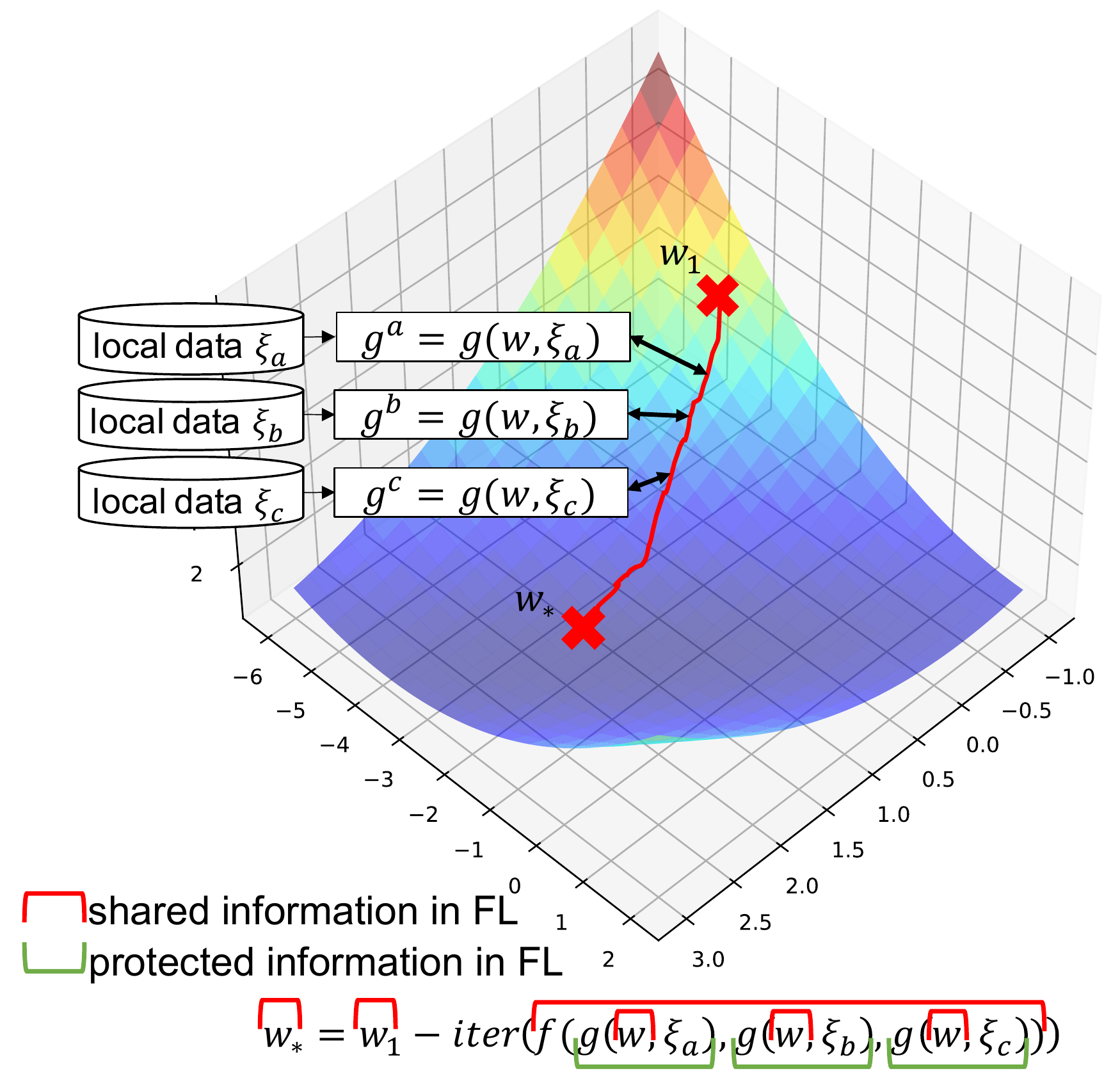} \label{fig:fl}}}%
	\qquad
\subfloat[\codename. Each participant strictly confines their own global models from their initialization~($w_0^a, w_0^b, w_0^c$) to optimal values~($w_*^a, w_*^b, w_*^c$).
The confined models descend in the same pace, and when \short converges, reach the bottom of the valley where the centralized model is located.  
Any two confined models keep the same distance throughout the training process.
In other words, $w^a, w^b, w^c$  would not become closer to each other during descending, 
preventing any participant from predicting models of others.
]{{\includegraphics[width=6.2cm]{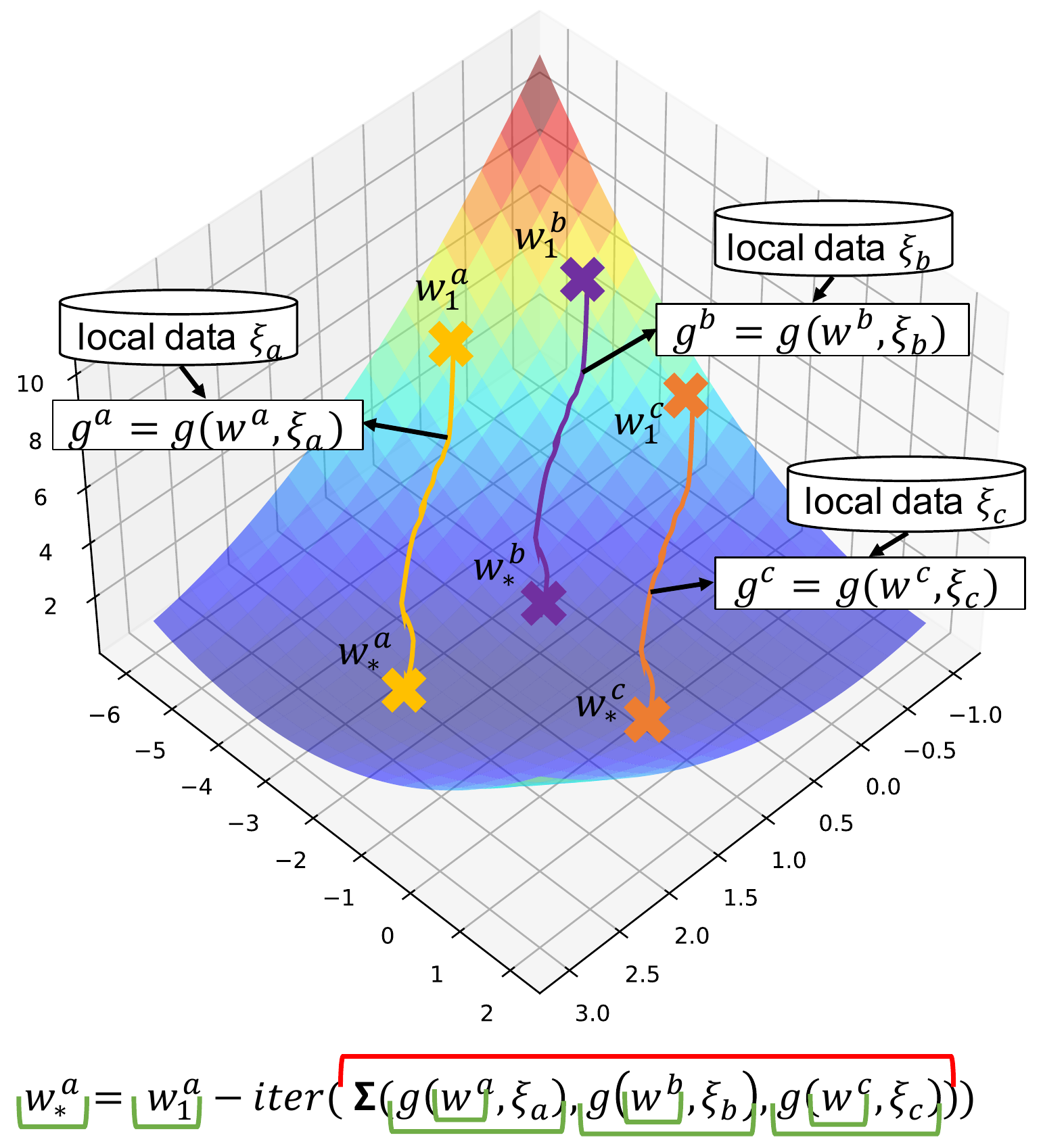} \label{fig:confined}}}%
	\caption{Comparison of \short and the gradient descent in  traditional FL. 
		This figure does not differentiate each iteration: the occurrences of traditional FL's global model $w_k$ and confined models $w_k^a, w_k^b, w_k^c$ in all iterations are represented by $w, w^a, w^b, w^c$, and $iter$ represents the sum-up of all iterations.
	}
	\label{fig:compgd}
\end{figure*}

In Figure~\ref{fig:compgd}, we illustrate a holistic comparison between the workflow of 
\short and that of a gradient decent in traditional FL.
In traditional FL, every participant updates the \emph{same} global model $w$ using their local gradients $g^a, g^b, g^c$.
In \short,  each participant $l$ first independently initializes the starting point of its confined model $w_1^l$.
Then, in every training iteration, participants independently compute the local gradient from their current confined model and local data, and then jointly work out the \emph{sum} of all local gradients and use it to update their confined models~(the equation in Figure \ref{fig:confined}).
By doing this, \short aims to \emph{enhance privacy without sacrificing much model accuracy}.
For the sake of simplicity, we refer to 
these two properties as \emph{privacy} and \emph{accuracy}. 
\begin{itemize}
	\item \textbf{Privacy.} 
	\short should ensure that, throughout the training process, neither local data of a participant nor intermediate results computed on them can be observed by other participants or an aggregator~(if any).
	\item \textbf{Accuracy.} The prediction made by any confined  model should approach the \emph{centralized model} that were to learn 
	centrally on the gathered data.
\end{itemize}
The desired privacy enhancement of \short stems from two aspects, i.e., \emph{secrecy of confined models} and \emph{secrecy of local gradients}. 
For the former, besides always hiding the confined models from each other, each participant independently initializes its $w_1^l$ at random. 
During the training process, any two confined models keep the same distance and never become closer to each other after descending, preventing any participant from predicting models of others.
To further boost the unpredictability, each participant could select its own interval range of initial weights to avoid leaking the average distance between the confined models.
\short withstands interval ranges differing by two orders of magnitude. 
For the latter, \short incorporates the secure addition operation on the local gradients to calculate their sum.
This has been proved to be viable through the additive secret sharing scheme, in which the sum of a set of secret values is collaboratively calculated without revealing any addends~\cite{bonawitz2017practical, bogdanov2008sharemind, lin2005efficient, tebaa2012homomorphic}.
A previous study~\cite{zhang2020privcoll} demonstrates it is efficient when applied to achieve decentralization in FL.
We prove that the adversary's observation in \short is only the sum of local gradients, and it conceals the extra indicative information that traditional FL would leak~(cf. Section~\ref{sec:privacy}).

We formally prove that \short ensures \emph{convergence}. 
It converges to confined models that are adjacent to the centralized model, and the adjacency is bounded~(cf. Section~\ref{sec:convergence}).
This merit guarantees the accuracy of \short. 
We further evaluate the accuracy performance of \short with two popular benchmark datasets, i.e., MNIST~\cite{lecun-mnisthandwrittendigit-2010} and CIFAR-10~\cite{krizhevsky2009learning}. 
Our experiments demonstrate that its accuracy closely approaches that of the centralized learning. 
When the confined models are initialized with the standard initialization scheme~(i.e., the Gaussian distribution of mean 0 and the variance 1), 
it achieves marginal differences on validation loss ~(mostly within  0.05) and  accuracy~(mostly within  1\%) 
and outperforms state-of-the-art federated learning with differential privacy.
Its accuracy performance remains stable even when the interval ranges of initial weights among participants differ by two orders of magnitude.

\paragraph{Contributions}
We summarize the main contributions as follows.

\begin{itemize}
	\item \textbf{\codename For Privacy-enhancing Decentralized Federated Learning.}
	We propose a new optimization algorithm \short for privacy-preserving decentralized FL. 
	\short  eliminates the explicit sharing of the global model and lets each participant learn a proprietary confined model. 
\short retains the merits of traditional FL such as algorithm independence.
Therefore, it can easily accommodate any FL schemes regardless of their underlying machine/deep learning algorithms.
	It also eliminates the necessity of a central coordinating server, 
	such that the optimization can be conducted in a fully decentralized manner.

	
	\item \textbf{Convergence Analysis.}
	We theoretically establish a convergence rate for  \short under 
	realistic assumptions on the loss function (such as convexity).
	We prove \short converges toward the centralized model as the number of iterations increases. 
	The distance between the trained confined models and the centralized model is bounded, and can be tuned by the hyper-parameter setting.

	\item \textbf{Enhanced Privacy Preservation Over Traditional FL.}
	With secrecy of both confined model and local gradients, \short  achieves enhanced privacy preservation over traditional FL.
	We prove that in \short,
given only the \emph{sum} of the local gradients, 
an \emph{honest-but-curious} white-box adversary, who may control $t$ out of $m$ participants~(where $t \leq m-2$) including the aggregator for secure addition operation~(if any), can learn no information other than their own inputs and the \emph{sum} of the local gradients from other honest parties, whereas in traditional FL, extra indicative information about local data can be obtained.



	\item \textbf{Functional Evaluations.} We implement \short and conduct experiments on two popular benchmark datasets MNIST and CIFAR-10.
	The results demonstrate that \short can closely approach the performance of centralized model on the validation loss and accuracy, 
	with marginal differences on validation loss ~(mostly within  0.05) and  accuracy~(mostly within  1\%).
	
\end{itemize}

\section{Background and Related Works} \label{sec:related}

\short is an optimization method based on the gradient decent. Therefore, in this section, we review the existing  techniques for gradient updates in the traditional FL.

\subsection{Stochastic gradient descent}
Stochastic gradient descent~(SGD)~\cite{robbins1951stochastic, bottou2018optimization} is an efficient variant of the gradient descent algorithm. It is extensively used for optimizing the objective function in machine learning and deep learning. 
Given a cost function $F$ with the parameter
$w$, SGD is defined by
\begin{equation}
	\label{equ:sgd}
	w_{k+1} \leftarrow w_k - \alpha_k \frac{1}{|\xi_{k}|} \nabla F(w_k, \xi_{k}),
\end{equation}
where $w_k$ are the parameter 
at the $k^{th}$ iteration, 
$\xi_{k} \in \xi$ is a randomly selected subset of the training samples at 
the $k^{th}$ iteration, and $\alpha_k$ is the learning rate. Equation \ref{equ:sgd} can generalize to 
the mini-batch update when $1 < |\xi_{k}| < |\xi|$, and to 
the batch update when $\xi_{k} = \xi$.

In 
FL, each local participant $l \in \mathcal{L}$ holds a subset of the training samples, denoted by 
$\xi_l$. To run SGD (or the mini-batch update), for each iteration, a random subset $\xi_{l,k} \subseteq \xi_l$ 
from a random participant $l$ is selected. The participant $l$ then computes the gradient with respect to $\xi_{l,k}$, which can be written as $\nabla F(w_k, \xi_{l,k})$, and shares the gradient with other participants (or a parameter server). All the participants (or the server) can thus take a gradient descent step by
\begin{equation}
	\label{equ:fdsgd}
	w_{k+1} \leftarrow w_k - \alpha_k \frac{1}{|\xi_{l,k}|} \nabla F(w_k, \xi_{l,k}).
\end{equation}

The gradients, 
if shared in plain text, 
are subject to information leakage of the local training data. For example, model-inversion attacks~\cite{melis2019exploiting, fredrikson2015model, shokri2017membership} are able to restore training data from the gradients. 
In the immediately following sections, we summarize the existing privacy-preserving methods for synthesizing the local gradients, which fall into two broad categories, i.e., \emph{secure aggregation} and \emph{learning with differential privacy}.

\subsection{Secure aggregation}
\label{sec:HE/MPC}
Secure aggregation typically employs cryptographic mechanisms such as homomorphic encryption (HE)~\cite{chen2018privacy, marc2019privacy, sharma2019confidential} and/or secure multiparty computation (MPC)~\cite{ gascon2017privacy, gascon2016secure, liu2017oblivious, mohassel2017secureml, bonawitz2017practical, zhang2020privcoll} to securely evaluate the gradient $\nabla F(w_k, \xi_{l,k})$ without revealing local data. 
With $\nabla F(w_k, \xi_{l,k})$, all the participants can thus take a gradient descent step by Equation~\ref{equ:fdsgd}.

Some existing studies fall into this category. For instance,
Bonawitz et al. 
~\cite{bonawitz2017practical} present a secure aggregation protocol that allows a server to compute the sum of  user-held data vectors, which can be used to aggregate user-provided model updates for a deep neural network.
Mohassel et al.~\cite{mohassel2017secureml} propose a secure two-party computation (2PC) protocol that supports secure arithmetic operations on shared decimal numbers for calculating the gradient updates using SGD.

Existing FL frameworks employing HE/MPC are mainly based on federated SGD.  All the participants in it share and update one and the same global model, and this  is  subject to membership inference, as revealed by Nasr et al.~\cite{nasr2019comprehensive}.

\subsection{Learning with differential privacy}
\label{sec:DP}
Another line of studies that approaches to privacy-preserving 
FL is through differential privacy (DP) mechanism~\cite{zheng2019bdpl, hagestedt2019mbeacon, abadi2016deep, song2013stochastic, shokri2015privacy, dwork2014algorithmic, hu2019fdml, zhang2020differentially, geyer2017differentially}.
The common practice of achieving differential privacy is based on additive noise calibrated to $\nabla F$’s sensitivity $\mathcal{S}^2_{\nabla F}$. 
As such, 
a differentially private learning framework 
can be achieved by updating parameters with perturbed gradients at each iteration, for example, to update parameters as
\begin{equation}
	\label{equ:dp}
w_{k+1} \leftarrow w_k - \alpha_k \frac{1}{|\xi_{l,k}|} (\nabla F(w_k, \xi_{l,k})+ \mathcal{N}(0, \mathcal{S}^2_{\nabla F} \cdot \sigma^2)),
\end{equation}
where $\mathcal{N}(0, \mathcal{S}^2_{\nabla F} \cdot \sigma^2)$ is the Gaussian distribution (a commonly used noise distribution in differentially private learning frameworks~\cite{dwork2014algorithmic}) with mean $0$ and standard deviation $\mathcal{S}_{\nabla F} \cdot \sigma$.

The privacy loss is accumulated with repeated access to the data during training epochs~\cite{abadi2016deep}. There is also an inherent tradeoff between privacy and utility of the trained model.


In summary, in all of the above approaches, the global model has to be shared with each participant, leading to the leakage of information. 
This motivates \short's design to eliminate the explicit sharing of the central global model.

\section{\codename}

\short optimizes an objective function in FL with multiple local datasets.
It starts with a colony of discrete points, and then uses the combination of their gradients to lead the optimization to another colony of points at the neighborhood of
the global optimum. 
In this section, we formalize this problem and present the 
workflow of \short optimization.

\subsection{Problem formulation}
\label{sec:problem}
\subsubsection{Optimization objective} Consider a centralized dataset $\xi = \{(x_i, y_i)\}^n_{i=1}$ consisting of $n$ training samples.
The goal of machine learning is to find a model parameter $w$ such that the overall loss, which is measured by the distance between the model prediction $h(w,x_i)$ and the label $y_i$ for each $(x_i,y_i) \in \xi$, is minimized.
This is reduced to solving the following problem
\begin{equation}
	\arg \min_w \frac{1}{n} \sum_{i=1}^{n} F (w;\xi) + \lambda z(w),
\end{equation}
where $F(w;\xi)$ is the loss function, and $z(w)$ is the regularizer for $w$. We use $w_*$ to denote the optimal solution of centralized training~(i.e., the centralized model).

In the context of FL, we have a system of $m$ local participants, each of which holds a private dataset $\xi_l \subseteq \xi $ $(l \in [1,m])$ 
consisting of a part of the training dataset.
The part could be a part of training samples, a part of features that have common entries, or both.
Assume the training takes $T$ iterations, and let $w^l_{k}$ denote the
confined model of participant $l$ at the $k^{th}$ iteration, where $k \in [1,T]$.
Let
\begin{equation}
	\label{equ:gl}
g^l(w^l_{k}, \xi_l)=\frac{1}{|\xi_{l}|}  \nabla F(w^l_{k}, \xi_{l})
\end{equation}
represent the local gradient with respect to $\xi_l$.
We use $w^l_*$ to denote the final confined model of participant $l$ when \short converges.
The objective of \short is to make $w^l_*$ located at a neighborhood of the centralized model $w_*$ within a bounded gap. 

\subsubsection{Attacker setting} \label{sec:attacker}
We assume an  \emph{honest-but-curious} white-box adversary\footnote{A white-box adversary knows the internals of the training algorithms such as the neural network architecture, and can observe the intermediate computations during the training iterations.} who may control $t$ out of $m$ participants (where $t \leq m-2$), including the aggregator for secure addition operation (if any).

\subsection{\short optimization}
\label{sec:optimization}

\begin{figure}
		\centering
	\subfloat[Centralized Training]{{\includegraphics[width=5.5cm]{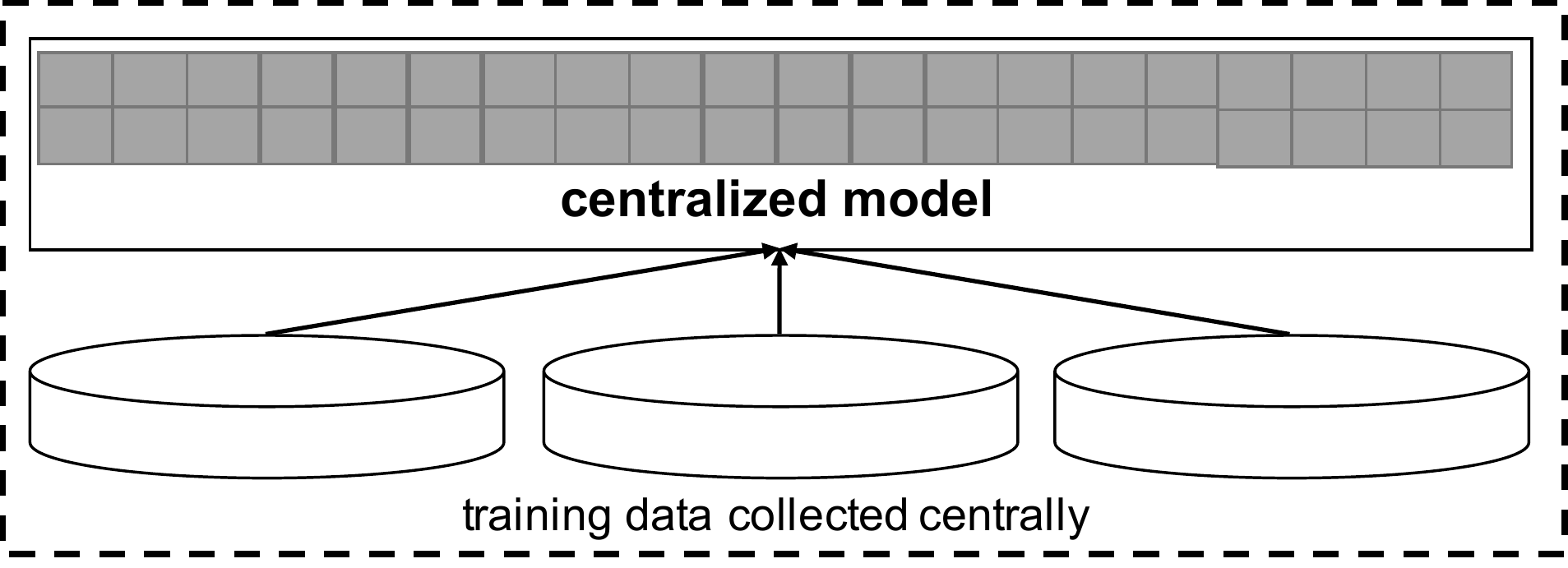} \label{fig:central}}}%
\qquad
	\subfloat[Traditional Federated Learning]{{\includegraphics[width=8cm]{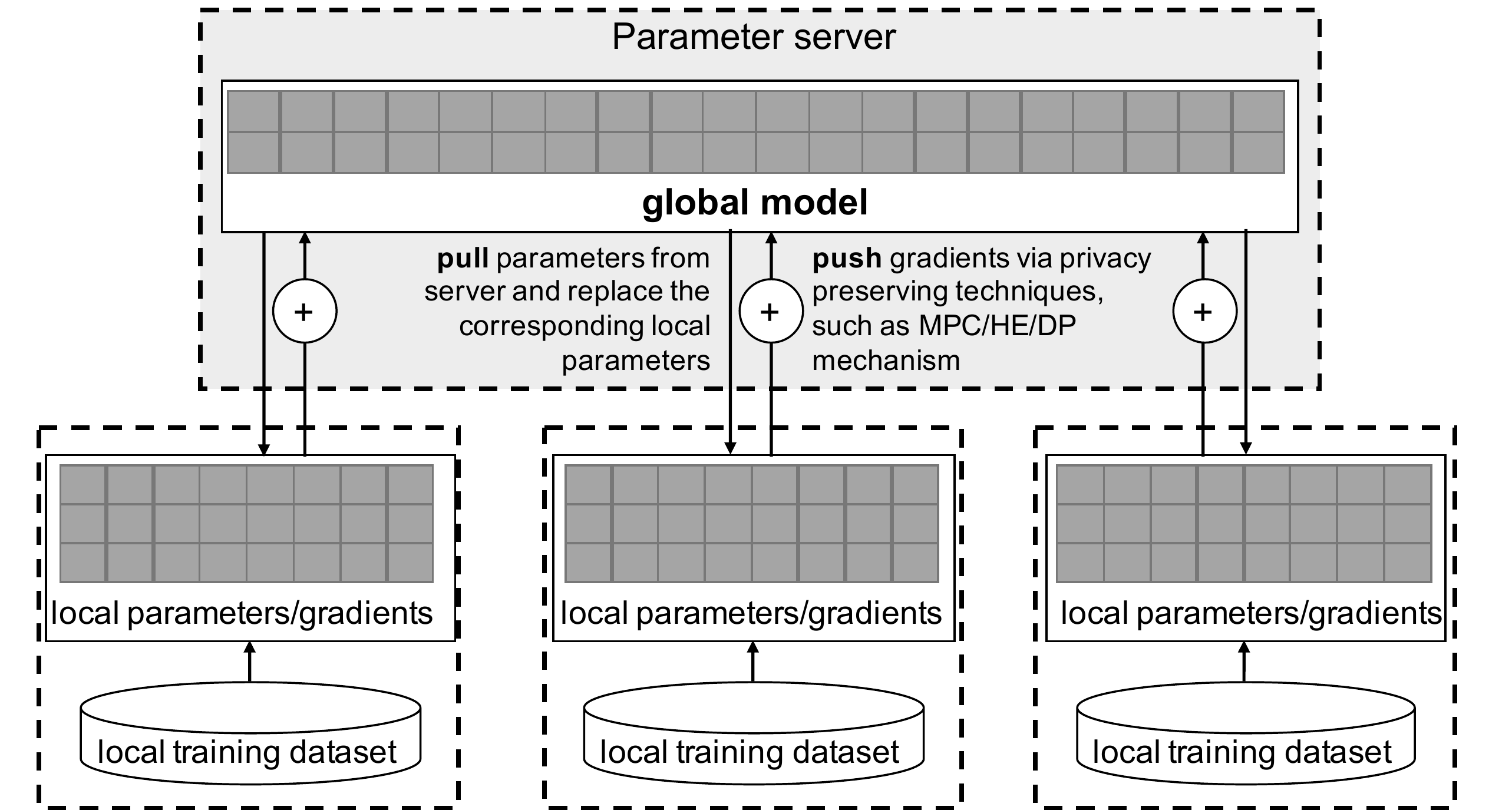} \label{fig:existing}}}%
	\qquad
	\subfloat[\short ]{{\includegraphics[width=8cm]{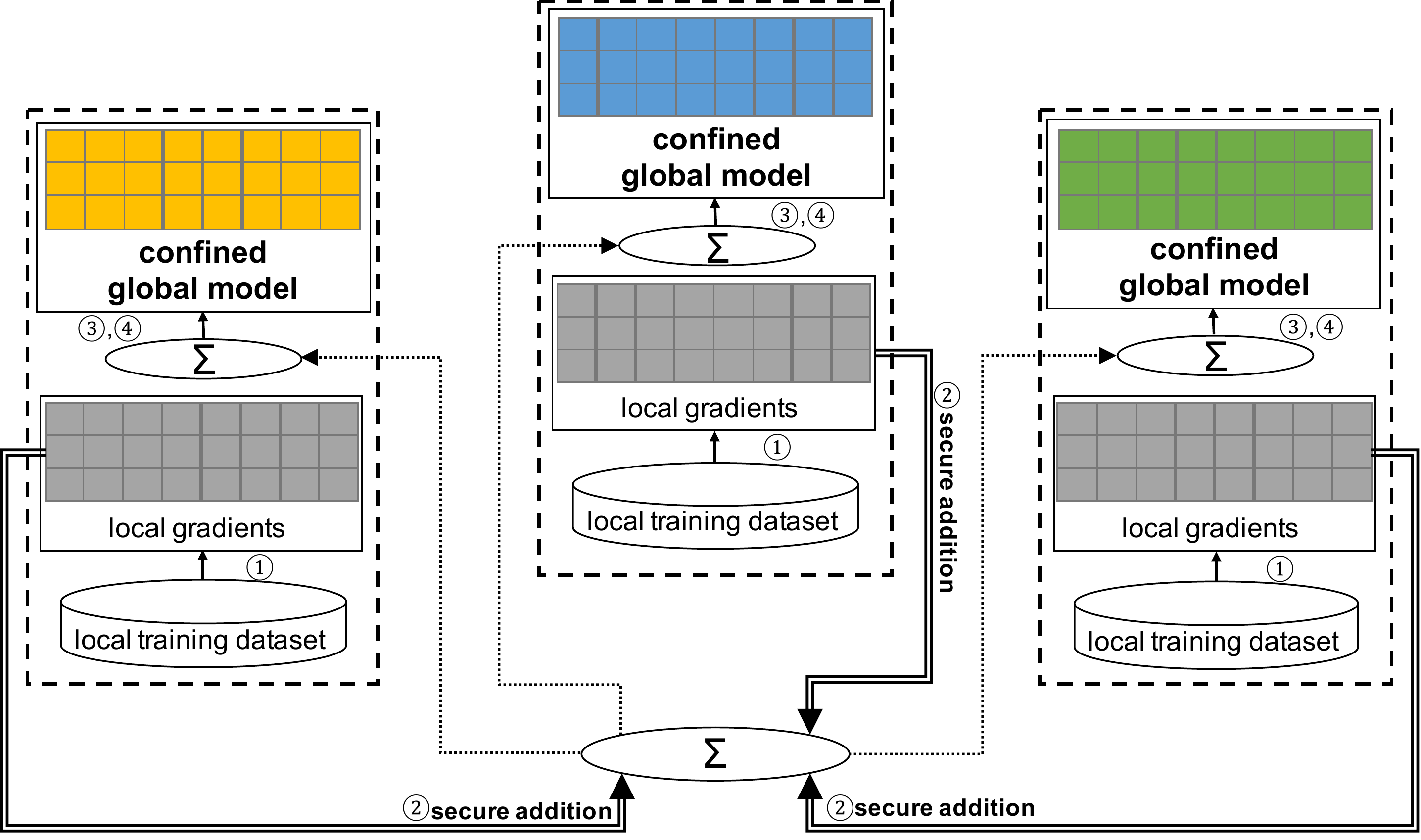} \label{fig:cgd}}}%
	\caption{Architectural comparison of the centralized training, traditional federated learning, and \short.}%
\label{fig:differ}	
\vspace{-0.5cm}
\end{figure}

Figure \ref{fig:differ} illustrates the architecture of \short (Figure \ref{fig:cgd}), with a comparison to the centralized training~(Figure \ref{fig:central}) and traditional FL~(Figure \ref{fig:existing}).
In the centralized training, the datasets of all participants are gathered for training a single model.
In the traditional FL, every participant owns its local training dataset, and updates the \emph{same} global model $w_{k}$ via a parameter server using its local model/gradients.
The local gradients $\nabla F(w_k, \xi_{l,k})$ can be protected via either secure aggregation~\cite{bonawitz2017practical, mohassel2017secureml, zhang2020privcoll,  marc2019privacy, sharma2019confidential, liu2017oblivious} or differential privacy mechanisms~\cite{abadi2016deep, song2013stochastic, shokri2015privacy, dwork2014algorithmic, hu2019fdml, zhang2020differentially}. 
This process can be decentralized by replacing the parameter server with a peer-to-peer communication mechanism~\cite{kim2019blockchained, roy2019braintorrent}. 
In \short, each participant learns its own confined model~(represented by different colors), i.e., each $w^l_{k}$ 
is \emph{different and private}.
The model updating in \short synthesizes the information from all training samples by summing up the local gradients, 
while in federated SGD~(Figure \ref{fig:existing}), each iteration takes into account only a subset of training samples.


To better position \short, we summarize the related studies in the literature in Table~\ref{tab:differ}.
We use federated SGD to represent the SGD or the mini-batch update 
in the FL, including both the plain SGD in which the local gradient/model is shared in plaintext,
and privacy-preserving SGD via secure aggregation or differential privacy mechanisms. 
\short guarantees that each confined model, when \short converges, is at the neighborhood of the centralized model, retaining the model accuracy (column 5 in Table \ref{tab:differ} and proved in Section \ref{sec:convergence}).
It also achieves desired privacy preservation compared to the traditional FL (column 6 in Table \ref{tab:differ} and detailed in Section \ref{sec:privacy}).


\begin{table*}[ht]
	\centering
	\resizebox{1\textwidth}{!}{
		\begin{tabular}{|p{0.07\linewidth} | p{0.17\linewidth} | p{0.13\linewidth} | p{0.23\linewidth} |p{0.22\linewidth} |p{0.24\linewidth}|}
			\hline
			\multicolumn{2}{|c|}{\multirow{2}{*}{Technique}}                                         & \multicolumn{2}{c|}{Methodology}                                                                                                                                                                     & \multicolumn{2}{c|}{Desired properties}                                                                                                                                                                                                                                                                                                                           \\ \cline{3-6}
			\multicolumn{2}{|c|}{}                                                                   & Architecture                                                                       & Model update                                                                                            & Model Accuracy                                                                                                                                                                                                      & Privacy                                                                                                                                     \\ \hline
			\multirow{3}{*}{\shortstack[l]{Federated \\ SGD  }} & Plain SGD~\cite{robbins1951stochastic, bottou2018optimization}                         & \multirow{3}{*}{\shortstack[l]{All the particip-\\ants  jointly  learn\\ one and the same \\ global  model.}} & \multirow{3}{*}{\shortstack[l]{Update the model  using  the \\ gradient  computed on a subset \\ of  training samples  in  each \\ iteration.}} & Theoretically guaranteed convergence~\cite{robbins1951stochastic, bottou2018optimization}.                                                                                    & The sharing of local parameters/gradients is subject to model inversion attack~\cite{melis2019exploiting, fredrikson2015model, shokri2017membership} .                                                                            \\ \cline{2-2} \cline{5-6}
			& Privacy-preserving SGD via secure aggregation~\cite{bonawitz2017practical, mohassel2017secureml, zhang2020privcoll,  marc2019privacy, sharma2019confidential, liu2017oblivious}  &                                                                                    &                                                                                                                 & As above, since the mechanism guarantees that the computation result from ciphertext is the same from plaintext~\cite{bonawitz2017practical,mohassel2017secureml}.& The sharing of one and the same global model (even though the local gradients are protected) is still subject to information leakage (refer to~\cite{nasr2019comprehensive} and Section \ref{sec:privacy}). \\ \cline{2-2} \cline{5-6}
			& Privacy-preserving SGD via differential privacy~\cite{abadi2016deep, song2013stochastic, shokri2015privacy, dwork2014algorithmic, hu2019fdml, zhang2020differentially}  &                                                                                    &                                                                                                                 & Additive noise (in most cases) affects the model accuracy~\cite{dwork2014algorithmic}.                                                                                                    &  The privacy cost is accumulated with repeated accesses to the dataset~\cite{abadi2016deep}.                                                      \\ \hline
			\multicolumn{2}{|l|}{\codename}                                           & Each participant learns and confines a different global model.                     & The model update in each  iteration embraces the information from all the training samples by summing up the local gradients.            & Theoretically guaranteed convergence with bounded gap to the centralized model.                                                                                                                         & Boosted privacy preservation by eliminating the sharing of the global model.                                                    \\ \hline
		\end{tabular}
	}	
	\caption{A summary of differences between the federated SGD and \short.}
	\label{tab:differ}
\end{table*}

Algorithm \ref{alg:cgd} outlines \short optimization for training with the confined model $w_k^l$.
In general, the optimization process consists of the following  steps.
\begin{itemize}
\item \textbf{Initialization. } Each  participant $l$ randomizes its own $w_1^l$. A default setting is to sample based on the Gaussian distribution of mean 0 and the variance 1, which is the standard weight initialization scheme used in most machine learning approaches~\cite{glorot2010understanding, krizhevsky2017imagenet, mishkin2015all}.
To prevent the colluding participants from inferring others' points by the knowledge of the average distance among the confined models, we introduce a hyper-parameter $\delta^l$ to control the interval range of initial weights $w_1^l$, i.e., $w_1^l \sim   N[-\delta^l, \delta^l]$, and allows each participant $l$ to independently choose its own $\delta^l$.

\item \textbf{Step 1. } At each iteration $k$, every participant computes the local gradient $g^l(w^l_{k}, \xi_l)$  with respect to its current confined model $w^l_{k}$ and 
own dataset $\xi_l$ using Equation \ref{equ:gl}. 
\item \textbf{Step 2. } Securely compute $\sum\limits_{l=1}^m g^l(w^l_{k}, \xi_l)$ which is later used for 
calculating $w^l_{k+1}$ (double lines in Figure \ref{fig:cgd}).


This step presumes that computational tools exist for securely evaluating the sum of a set of secret values to avoid releasing local gradient $g^l(w^l_{k}, \xi_l)$ in plain text.
We refer to Section \ref{sec:privacy} for more detail.
\item \textbf{Step 3. } A scalar stepsize $\alpha_k > 0$ is chosen given an iteration number $k\in[1,T]$.
\item \textbf{Step 4. } Every participant takes a descent step on its own $w_k^l$ with $\alpha_k \sum\limits_{l=1}^m g^l(w^l_{k}, \xi_l)$, 
i.e., $w_{k+1}^l \leftarrow w_k^l - \alpha_k \sum\limits_{l=1}^m g^l(w^l_{k}, \xi_l)$.

\end{itemize}


\begin{algorithm}
	\caption{\codename Optimization}
	\label{alg:cgd}
	\begin{algorithmic}[1]
		\State{\textbf{Input:} Local training data $\xi_l$ ($l \in [1, m]$), }
		\State{\qquad number of training iterations $T$}
		\State{\textbf{Output:} Confined global model parameters $w_*^l$ ($l \in [1, m]$) }
		\State{\textbf{Initialize:} $k\leftarrow 1$, each participant $l$ randomizes its own $w_1^l$}
		\While{$k \leq T$}
		\State{\textbf{for all} participants $l \in [1, m]$ \textbf{do in parallel}}
		\State{\qquad Compute the local gradient: $g^l(w^l_{k}, \xi_l)$ }
		\State{\qquad Securely evaluate the sum: $\sum\limits_{l=1}^{m}g^l(w^l_{k}, \xi_l)$ }
		\State{\qquad Choose a stepsize: $\alpha_k$}
		\State{\qquad Set the new iterate as: $w_{k+1}^l \leftarrow w^l_{k} - \alpha_k \sum\limits_{l=1}^{m}g^l(w^l_{k}, \xi_l)$}
		\State{\textbf{end for}}
		\State{$k \leftarrow k+1 $}		
		\EndWhile
	\end{algorithmic}
\end{algorithm}






\section{Convergence Analysis}
\label{sec:convergence}

In this section, we conduct a formal convergence analysis on \short.
Convergence analysis has been extensively used in the literature~\cite{hu2019fdml, hsieh2017gaia, ho2013more} to prove the correctness of optimization algorithms.
Through the analysis, we demonstrate the bound of the distance between an arbitrary $w_*^l$ learned by \short and the centralized model $w_*$.
The analysis is centered around a \emph{regret} function $\mathcal{R}$, which is the difference between the \short's training loss 
and the loss of the 
centralized model, defined as
\begin{equation}
	\label{equ:regret}
\mathcal{R} = \frac{1}{T} \sum_{k=1}^{T} (F(w_k^l) - F(w_*))
\end{equation}


\subsection{Assumptions}
We make the following  assumptions on the loss function $F$.
They all are common assumptions in convergence analyses of most gradient-based methods, and satisfied in a variety of widely used cost functions~\cite{bottou2018optimization}, such as mean squared error (MSE) and cross entropy.

\begin{assumption}
	\label{assum:Lip}
(Lipschitz continuity). 
The loss function $F$: $\mathbb{R}^d \rightarrow \mathbb{R}$ is continuously differentiable and the gradient function of $F$, namely, $\nabla F$: $\mathbb{R}^d \rightarrow \mathbb{R}^d$, is Lipschitz continuous with Lipschitz constant $L > 0$, 
\begin{equation}
\|\nabla F(w) - \nabla F(\widehat{w})\|_2 \leq L\|w-\widehat{w}\|_2 \text{\quad for all } \{w,\widehat{w}\} \subset \mathbb{R}^d.
\end{equation}
\end{assumption}
Intuitively, this assumption ensures that the gradient of $F$ does not change arbitrarily 
in the course of descending, such that the gradient can be a 
proper indicator 
towards the optimum~\cite{bottou2018optimization}. 

\begin{assumption}
	\label{assum:convex}
	(Strong convexity).
	The loss function $F$: $\mathbb{R}^d \rightarrow \mathbb{R}$ is strongly convex such that
	\begin{equation}
		F(\widehat{w}) \geq F(w) + \nabla F(w)^T(\widehat{w}-w)  \text{\quad for all } \{w,\widehat{w}\} \subset \mathbb{R}^d.
	\end{equation}
\end{assumption}

A useful fact from Assumption \ref{assum:convex} for our analysis is
\begin{equation}
	\label{equ:convex}
(F(w)-F(w_*)) \leq \langle w-w_*, \nabla F(w) \rangle  \text{\quad for all } w \subset \mathbb{R}^d,
\end{equation}
where $\langle\cdot{,}\cdot\rangle$ denotes the inner product operation.

In addition, we adopt the same assumption on a bounded solution space used in related studies~\cite{bottou2018optimization, hu2019fdml}.
\begin{assumption}
	\label{assum:sol}
(Bounded solution space). The set of $\{w_k^l\}|_{k \in [1,T]}$ 
is contained in an open set over which $F$ is bounded below a scalar $F_{inf}$, such that
\begin{enumerate}[label=(\alph*)]
\item  there exists a $D>0$, s.t.,	$\|w_k^l-w_*\|_2^2 \leq D^2$ for all $k$, and \label{assum:sola}
\item there exists a $G>0$, s.t., $\|g^l(w_k^l, \xi_l)\|_2^2 \leq G^2$ for all $w_k^l \subset \mathbb{R}^d$.  ~\label{assum:solb} 
\end{enumerate}
\end{assumption}

\subsection{Main Theorem}

We first present our main theorem below, and leave its proof to Section~\ref{sec:proof}. 
It demonstrates our main result on the converge rate of \short.
\begin{theorem}
	\label{theorem:main}
Given a cost function satisfying Assumptions \ref{assum:Lip} - \ref{assum:sol}, and a learning rate of $\alpha_k = \frac{\alpha}{(k+\mu T)^2}$ ($0<\mu <1$), the \short optimization gives the regret $\mathcal{R}$  
\begin{equation}
	\mathcal{R} =O(\epsilon + \frac{1}{\mu T +1}+ \frac{\ln |\mu T +1|}{T})
\end{equation}
where $\epsilon = m \| \displaystyle \mathop{\mathbb{E}}_{j\in m}(w_1^l-w_1^j)\|$. 
\end{theorem}
The theorem implies the following two remarks. 
\begin{itemize}
	\item \textbf{Convergence rate.} 
	Both $\frac{1}{\mu T +1}$ and $\frac{\ln |\mu T +1|}{T}$ approach 0 as $T$ increases,  implying that
	\short will converge toward the optimum. 
	The convergence rate can be adjusted by the parameter $\mu$~(
	the effect of $\mu$ 
	is investigated in Section \ref{sec:mu}).
	\item \textbf{Bounded optimality gap. }
	When \short converges, the  
	gap between the confined 
	models and the centralized model is bounded by $\epsilon$.
	\short uses the initialization parameter $\delta^l$ 
	to determine the range of $w_1^l$  in the way of
	\begin{equation}
		\label{equ:delta}
		w_1^l = \delta^l\cdot \mathit{Rand}^l  \text{\quad for all } l \in (1, m),
	\end{equation}
	where $\mathit{Rand}$ is the initialization scheme. A standard $\mathit{Rand}$ used in machine learning is to apply random sampling from the Gaussian distribution of mean 0 and the variance 1, and a standard $\delta$ is $\frac{1}{\sqrt n}$ (where $n$ is the sample size)~\cite{glorot2010understanding}. 
	In \short, each participant 
	determines its own $\mathit{Rand}^l$ and $\delta^l$ 
	independently to avoid leaking the average distance among the confined models.
	Our experiment finds that \short keeps robust~(in terms of 
	validation accuracy) 
	even when the participants select their $\delta^l$s uniformly at random in a range of $(\frac{10}{\sqrt n}, \frac{0.1}{\sqrt n})$ (cf. Section \ref{sec:init}). 
		
\end{itemize}

\subsection{Proof of Theorem \ref{theorem:main}} \label{sec:proof}

Our proof aims to identify an upper bound of  $\mathcal{R}$.
To this end, we consider the trend of the distance from $w_{k}^l$ to $w_*$, which would shrink as $k$ increases, if there is a bound existing. 
Since $w_{k}^l$ is updated using $\sum\limits_{j=1}^{m}g^j(w_k^j, \xi_j)$ (i.e., the descent in  \short), and $w_*$ is obtained by $\nabla F$ (i.e., the descent in the centralized training),  the trend of the distance therefore should be related to the deviation between these two.
Exploring this leads to the following lemma which describes this relationship.

\begin{lemma}
	\label{lemma:inner}
	Let $S_{k+1} = \frac{1}{2} \|w_{k+1}^l - w_*\|_2^2$ and $S_{k} = \frac{1}{2} \|w_{k}^l - w_*\|_2^2$.  Let $\nabla F(\cdot) = \frac{1}{n} \sum_{i=1}^{n}  \nabla F(\cdot, \xi_{i})$.  
	We have
\end{lemma}
	\begin{multline*}
	\langle w_{k}^l - w_*, \nabla F(w_k^l) \rangle =
	\frac{1}{2} \alpha_k \|\sum\limits_{j=1}^{m}g^j(w_k^j, \xi_j)\|_2^2 
	- \frac{1}{\alpha_k} (S_{k+1}-S_k)\\ 
	- \langle w_{k}^l - w_*, \sum\limits_{j=1}^{m}g^j(w_k^j, \xi_j) - \nabla F(w_k^l)\rangle. 
\end{multline*}


\begin{proof}
	\begin{equation}
		\label{equ:lemma}
		\begin{aligned}
			& S_{k+1} - S_k = \frac{1}{2} \bigg( \|w_{k+1}^l - w_*\|_2^2 -  \|w_{k}^l - w_*\|_2^2 \bigg) \\
			&= \frac{1}{2} \bigg( \|w_k^l - \alpha_k \sum\limits_{j=1}^{m}g^j(w_k^j, \xi_j)  - w_*\|_2^2 - \|w_{k}^l - w_*\|_2^2 \bigg) \\
			&= \frac{1}{2} \| \alpha_k \sum\limits_{j=1}^{m}g^j(w_k^j, \xi_j)\|_2^2 - \alpha_k \langle w_{k}^l - w_*, \sum\limits_{j=1}^{m}g^j(w_k^j, \xi_j)\rangle\\
			&= \frac{1}{2} \| \alpha_k \sum\limits_{j=1}^{m}g^j(w_k^j, \xi_j)\|_2^2 \\
			& \qquad - \alpha_k \langle w_{k}^l - w_*, \sum\limits_{j=1}^{m}g^j(w_k^j, \xi_j) + \nabla F(w_k^l) - \nabla F(w_k^l) \rangle\\
			& = \frac{1}{2} \| \alpha_k \sum\limits_{j=1}^{m}g^j(w_k^j, \xi_j)\|_2^2 -  \alpha_k \langle w_{k}^l - w_*, \nabla F(w_k^l) \rangle\\
			& \qquad  \qquad \qquad- \alpha_k \langle w_{k}^l - w_*, \sum\limits_{j=1}^{m}g^j(w_k^j, \xi_j) - \nabla F(w_k^l) \rangle
		\end{aligned}
	\end{equation}
	Dividing the above equation by $\alpha_k$, we can 
	prove the lemma.
\end{proof}

In the following, we give the proof of Theorem \ref{theorem:main}.
It calculates a function that is greater than $\mathcal{R}$ based on the convexity of the objective function~(Inequation~\ref{equ:conv}). 
The function can be decomposed into three terms based on Lemma \ref{lemma:inner}~(Equation~\ref{equ:threeterms}).
We then explore the boundedness of each term, and taking these bounds together concludes the proof.

\begin{proof}
	By the definition 
	of the regret function~(Equation \ref{equ:regret}) and  Equation \ref{equ:convex} in Assumption \ref{assum:convex}, 
	we have
	\begin{equation} 
		\label{equ:conv}
		\mathcal{R} = \frac{1}{T} \sum_{k=1}^{T} (F(w_k^l) - F(w_*)) \leq \frac{1}{T} \sum_{k=1}^{T} \langle w_{k}^l - w_*, \nabla F(w_k^l) \rangle
	\end{equation}
	Applying Lemma \ref{lemma:inner} to Inequation \ref{equ:conv} and multiplying 
	it by $T$, we have
	\begin{equation}
		\label{equ:threeterms}
		\begin{aligned}
			T \cdot R & \leq  \sum_{k=1}^{T} \bigg( 	\frac{1}{2} \alpha_k \|\sum\limits_{j=1}^{m}g^j(w_k^j, \xi_j)\|_2^2 \\
			& \quad - \frac{1}{\alpha_k} (S_{k+1}-S_k)- \langle w_{k}^l - w_*, \sum\limits_{j=1}^{m}g^j(w_k^j, \xi_j) - \nabla F(w_k^l)\rangle \bigg) \\
			& \leq \sum_{k=1}^{T} \frac{1}{2} \alpha_k \|\sum\limits_{j=1}^{m}g^j(w_k^j, \xi_j)\|_2^2 - \sum_{k=1}^{T} \frac{1}{\alpha_k} (S_{k+1}-S_k)\\
			&  \quad - \sum_{k=1}^{T}  \langle w_{k}^l - w_*, \sum\limits_{j=1}^{m}g^j(w_k^j, \xi_j) - \nabla F(w_k^l)\rangle
		\end{aligned}
	\end{equation}
	Inequation \ref{equ:threeterms} can be decomposed into three terms.
  The first two terms $\sum_{k=1}^{T} \frac{1}{2} \alpha_k \|\sum\limits_{j=1}^{m}g^j(w_k^j, \xi_j)\|_2^2$ and $-\sum_{k=1}^{T} \frac{1}{\alpha_k} (S_{k+1}-S_k)$ sums up the model updates throughout the training iterations. 
	The third term $- \sum_{k=1}^{T}  \langle w_{k}^l - w_*, \sum\limits_{j=1}^{m}g^j(w_k^j, \xi_j) - \nabla F(w_k^l)\rangle$ measures the gap of the gradients between  \short and the centralized training.
	
	Next, we 
	explore the boundedness of each term. 	
	For the first term, we have
	\begin{align}
		\sum_{k=1}^{T} \frac{1}{2} \alpha_k \|\sum\limits_{j=1}^{m}g^j(w_k^j, \xi_j)\|_2^2 & \leq \sum_{k=1}^{T} \frac{1}{2} \frac{\alpha}{(k+\mu T)^2} m^2 G^2 ~\label{equ:firstbound}\\
		&=\frac{\alpha m^2 G^2}{2} \sum_{k=1}^{T} \frac{1}{(k+\mu T)^2} < \alpha m^2 G^2 ~\label{equ:basel}.
	\end{align}
	Inequation \ref{equ:firstbound} is 
	based on Assumption \ref{assum:sol}\ref{assum:solb}, and Inequation \ref{equ:basel} is 
	based on the solution to the Basel problem that $\sum_{x=1}^{\infty} \frac{1}{x^2} < 2$. 
	
	For the second term, we have
	\begin{align}
		- \sum_{k=1}^{T} \frac{1}{\alpha_k} (S_{k+1}-S_k) & = \sum_{k=1}^{T} \frac{1}{\alpha_k} (S_k-S_{k+1}) \\
		&= \sum_{k=1}^{T} \frac{1}{\alpha_k} \bigg( \frac{1}{2} \|w_{k}^l - w_*\|_2^2 - \frac{1}{2} \|w_{k+1}^l - w_*\|_2^2 \bigg)\\
		& \leq \sum_{k=1}^{T} \frac{1}{2\alpha_k}  \|(w_{k}^l - w_*)- (w_{k+1}^l - w_*)\|_2^2 ~\label{equ:reversetria}\\
		& = \sum_{k=1}^{T} \frac{1}{2\alpha_k} \|w_{k}^l - w_{k+1}^l \|_2^2
		=  \sum_{k=1}^{T} \frac{1}{2\alpha_k}  \|\alpha_k \sum\limits_{j=1}^{m}g^j(w_k^j, \xi_j)\|_2^2\\
		& = \sum_{k=1}^{T} \frac{1}{2} \alpha_k \|\sum\limits_{j=1}^{m}g^j(w_k^j, \xi_j)\|_2^2 < \alpha m^2 G^2 ~\label{equ:second}
	\end{align}
	Inequation \ref{equ:reversetria} follows reverse triangle inequality, and Inequation \ref{equ:second} reuses the result of the first term (cf. Equation \ref{equ:basel}).
	
	Determining the bound of the third term is slightly complex. We list it as the following claim, and prove it soon after the proof of Theorem~\ref{theorem:main}.
	\begin{claim}
		\label{claim:term3}
		\begin{equation}
		\begin{aligned}
	&- \sum_{k=1}^{T}\langle w_{k}^l - w_*, \sum\limits_{j=1}^{m}g^j(w_k^j, \xi_j) - \nabla F(w_k^l)\rangle \\
	&< DL \|T\sum_{j=1}^{m}(w_1^l-w_1^j)\| + 2m^2GDL(\frac{T}{\mu T +1} + \ln |\mu T +1|)
\end{aligned}
		\end{equation}
	\end{claim}

	Combining Inequations \ref{equ:threeterms}, \ref{equ:basel}, \ref{equ:second} and Claim \ref{claim:term3}, and dividing by T we obtain
\begin{align*}
	\mathcal{R} &< \frac{2 \alpha m^2 G^2}{T} + DL \|\sum_{j=1}^{m}(w_1^l-w_1^j)\| + 2m^2GDL(\frac{1}{\mu T +1} + \frac{\ln |\mu T +1|}{T}) \\
	& = O(\epsilon + \frac{1}{\mu T +1}+ \frac{\ln |\mu T +1|}{T}), \text{concluding the proof.}
\end{align*}
\end{proof}

\paragraph{Proof of Claim \ref{claim:term3}}
	\begin{proof}
	\mathleft
\[- \sum_{k=1}^{T}\langle w_{k}^l - w_*, \sum\limits_{j=1}^{m}g^j(w_k^j, \xi_j) - \nabla F(w_k^l)\rangle \]
\mathcenter
\begin{align}
	&=\langle w_{k}^l - w_*, \sum_{k=1}^{T} \big(\nabla F(w_k^l) - \sum\limits_{j=1}^{m}g^j(w_k^j, \xi_j)\big)\rangle \\
	&\leq \|w_{k}^l - w_*\|\cdot\|\sum_{k=1}^{T} \big(\nabla F(w_k^l) - \sum\limits_{j=1}^{m}g^j(w_k^j, \xi_j)\big)\| ~\label{equ:tria}\\
	&\leq \|w_{k}^l - w_*\|\cdot L \| \sum_{k=1}^{T} \sum_{j=1}^{m} (w_k^l -  w_k^j)\| ~\label{equ:lip}\\
	&\leq DL\|\sum_{j=1}^{m}(w_1^l-w_1^j) + ... +  \sum_{j=1}^{m}(w_T^l-w_T^j)\| ~\label{equ:assu3a}
\end{align}
\begin{equation}
	\begin{aligned}
		= DL \|&T\sum_{j=1}^{m}(w_1^l-w_1^j)  \\
		&-\sum_{k=1}^{T-1} \sum_{j=1}^{m} \alpha_{(T-k)} k \big(\nabla F(w_{(T-k)}^l) -  \sum\limits_{j=1}^{m}g^j(w_{(T-k)}^j, \xi_j)\big) \| ~\label{equ:rew}
	\end{aligned}
\end{equation}
\begin{equation}
	\begin{aligned}
		\leq  DL & \bigg(\|T\sum_{j=1}^{m}(w_1^l-w_1^j)\|  \\
		&+ \| \sum_{k=1}^{T-1} \sum_{j=1}^{m} \alpha_{(T-k)} k \big(\nabla F(w_{(T-k)}^l) -  \sum\limits_{j=1}^{m}g^j(w_{(T-k)}^j, \xi_j)\big) \| \bigg)~\label{equ:rewnorm}
	\end{aligned}
\end{equation}
\begin{align}
	&\leq  DL \|T\sum_{j=1}^{m}(w_1^l-w_1^j)\| + DL\cdot2m^2G \sum_{k=1}^{T-1} \alpha_{(T-k)} k \quad \quad \quad \quad ~\label{equ:assu3b} \\
	& = DL \|T\sum_{j=1}^{m}(w_1^l-w_1^j)\| + 2m^2GDL \sum_{k=1}^{T-1} \frac{k}{((1+\mu)T-k)^2}~\label{equ:insertalpha} \\
	& < DL \|T\sum_{j=1}^{m}(w_1^l-w_1^j)\| + 2m^2GDL(\frac{T}{\mu T +1} + \ln |\mu T +1|) ~\label{equ:integral}
\end{align}

Inequations \ref{equ:tria} and \ref{equ:rewnorm} are from triangle inequality.
Inequation \ref{equ:lip} is from the fact $\nabla F(w_k^l) = \frac{1}{n} \sum\limits_{i=1}^{n}  \nabla F(w_k^l, \xi_{i}) = \sum\limits_{j=1}^{m} g^j(w_k^l, \xi_j)$ and Assumption \ref{assum:Lip}'s blockwise Lipschitz-continuity.
Inequation \ref{equ:assu3a} is from Assumption \ref{assum:sol}\ref{assum:sola} and represents $\sum\limits_{k=1}^{T}(\cdot) $ by a summand sequence.
Equation \ref{equ:rew} comes from the fact
\begin{multline*}
	w_k^l-w_k^j = (w_1^l - \alpha_1 \nabla F(w_1^l) - ... - \alpha_k \nabla F(w_k^l)) \\
	-(w_1^j - \alpha_1 \sum\limits_{j=1}^{m}g^j(w_1^j, \xi_j) - ... - \alpha_k \sum\limits_{j=1}^{m}g^j(w_k^j, \xi_j))\\
	= (w_1^l - w_1^j) - \alpha_1 (\nabla F(w_1^l) - \sum\limits_{j=1}^{m}g^j(w_1^j, \xi_j)) \\
	-...- \alpha_k(\nabla F(w_k^l) -  \sum\limits_{j=1}^{m}g^j(w_k^j, \xi_j)).
\end{multline*}
Inequation \ref{equ:assu3b} follows Assumption \ref{assum:sol}\ref{assum:solb} from which we obtain $\|\nabla F(w_{(T-k)}^l) -  \sum\limits_{j=1}^{m}g^j(w_{(T-k)}^j, \xi_j)\| \leq 2mG$.
Equation \ref{equ:insertalpha} is obtained by 
applying $\alpha_{(T-k)} = \frac{\alpha}{(T-k +\mu T)^2}$.
Inequation \ref{equ:integral} is from the following fact.

Since\[\int_a^b \frac{k}{(c-k)^2}\, dk = \frac{b}{c-b} + \ln |c-b| - \frac{a}{c-a} - \ln |c-a|,\]
we have
\begin{align*}
	&\int_1^{T-1} \frac{k}{((1+\mu)T-k)^2}\, dk = \frac{T-1}{(1+\mu)T-(T-1)} \\
	&+ \ln |(1+\mu)T-(T-1)| - \frac{1}{(1+\mu)T-1} - \ln |(1+\mu)T-1|\\
	&<\frac{T-1}{(1+\mu)T-(T-1)}+ \ln |(1+\mu)T-(T-1)| \text{ \quad (with $T\geq2$) } \\
	&< \frac{T}{\mu T +1} + \ln |\mu T +1|
\end{align*}	
	\end{proof}

\section{Privacy Preservation}
\label{sec:privacy}

The participants in \short have to share the sum of local gradients, i.e., $ \sum\limits_{j=1}^{m}g^j(w^j_{k}, \xi_j)$.
A straightforward way is to let each participant release its local gradient $g^l(w^l_{k}, \xi_l)$, but it may leak information about $\xi_l$ or $w^l_k$~\cite{zhang2020privcoll}.
To address this, we incorporate the secure addition operation~\cite{bonawitz2017practical, bogdanov2008sharemind, lin2005efficient, tebaa2012homomorphic} on the local gradients to calculate their sum without releasing each of them.
We make use of the \emph{additive secret sharing scheme} proposed by Bogdanov et al.~\cite{bogdanov2008sharemind}, which uses additive sharing over $\mathbb{Z}_{2^{32}}$ for securely evaluating addition operations in a multiparty computation environment.
It guarantees the secrecy of the addends even though the majority ($m-2$ out of $m$) of participants are compromised.
An brief introduction of the additive secret sharing scheme can be found in Appendix \ref{apx:secret}.

In the rest of this section, we explore the privacy preservation of \short.
We demonstrate CGD’s privacy enhancement over traditional FL. 
We prove that less information is exposed in CGD compared to that of traditional FL.


\subsection{Information exposed in \short}
Recall that the involved parties are a set $\mathcal{L}$ of $m$ participants denoted with logical identities $l \in [1, m]$, and $t$ is the adversarial threshold ($t \leq m-2$) (Section \ref{sec:attacker}).
Let $c$ be any subset of $\mathcal{L}$ that includes the compromised and colluding parties.

We demonstrate that, during optimization in \short, given only the sum of the local gradients which are computed on different confined models, the adversary can learn no information other than their own inputs and the sum of the local gradients from other honest parties, i.e., $\sum g_k^l(w^l_{k}, \xi_l)\mid_{l \in \mathcal{L} \setminus  c}$.

Our analysis is based on the simulation paradigm~\cite{goldreich2019play}.
It compares what an adversary can do in a real protocol execution  to what it can do in an ideal scenario, which is secure by definition.
The adversary in the ideal scenario, is called the simulator.
An indistinguishability between adversary's view in real and ideal scenarios guarantees that it can learn nothing more than their own inputs and the information required by the simulator for the simulation.
A brief introduction of the simulation paradigm is given in Appendix \ref{apx:sim}.

To facilitate the understanding on our analysis, we 
first present the used notations.
Denote  $g_k^{\mathcal{L}^\prime} = \{g_k^l(w^l_{k}, \xi_l)\}_{l \in \mathcal{L}^\prime}$ as the local gradients of any subset of participants  $\mathcal{L}^\prime \subseteq \mathcal{L}$ 
at $k^{th}$ iteration.
Let $VIEW_{real} (g_k^\mathcal{L}, t, \mathcal{P}, c)$ denote their combined views 
from the execution of a real protocol $\mathcal{P}$.
Let $VIEW_{ideal} (g_k^c, z, t, \mathcal{F}_p, c)$ denote the views of $c$ from an ideal execution that securely computes a 
function $\mathcal{F}_p$, where $z$ is the information required by the simulator $\mathcal{S}$ in the ideal execution for simulation.

The following theorem shows that when executing \short with the threshold $t$, the joint view of the participants in $c$ 
can be simulated by their own inputs
and the sum of the local gradients from the remaining honest nodes, i.e., $\sum g_k^l\mid_{l \in \mathcal{L} \setminus  c}$. 
Therefore, $\sum g_k^l\mid_{l \in \mathcal{L} \setminus  c}$ is the only information that the adversary can learn during the execution.

\begin{theorem}
	\label{theorem_party}
	When executing \short with the threshold $t$, there exists a simulator $\mathcal{S}$ such that for $\mathcal{L}$ and $c$, with $c \subseteq \mathcal{L}$ and $\lvert c \lvert \leq t$, the output of $\mathcal{S}$ from $VIEW_{ideal}$ is perfectly indistinguishable from the output of  $VIEW_{real} $, namely
	\[ VIEW_{real} (g_k^\mathcal{L}, t, \mathcal{P}, c) \equiv VIEW_{ideal} (g_k^c, z, t, \mathcal{F}_p, c)\]
	where
	\[z = \sum g_k^l(w^l_{k}, \xi_l)\mid_{l \in \mathcal{L} \setminus  c}.\]
\end{theorem}

\begin{proof}
	We  define  $\mathcal{S}$  
	through each training iteration as:
	
	$\text{\textbf{SIM}}_1$: ${SIM}_1$ is the simulator  for the first training iteration.
	
	Since the inputs of the parties in $c$ do not depend on the inputs of the honest parties in $\mathcal{L} \setminus c$,
		${SIM}_1$ can produce a perfect simulation by running $c$ on their true inputs, and $\mathcal{L} \setminus c$ on a set of pseudorandom vectors  $\eta_1^{\mathcal{L} \setminus c} =  \{\eta_1^l\}_{l \in \mathcal{L} \setminus c}$ in a way that
	\[\sum \eta_1^{\mathcal{L} \setminus c} = \sum \eta_1^l\mid_{l \in \mathcal{L} \setminus c} = \sum g_1^\mathcal{L} - \sum g_1^c = \sum g_1^l\mid_{l \in \mathcal{L} \setminus  c}.\]
	
	Since each  $g_1^l(w^l_{1}, \xi_l)$ is computed from its respective confined model $w_1^l$ which is randomized in the initialization, the pseudorandom vectors $\eta_1^{m \setminus c} $ generated by ${SIM}_1$ for the inputs of all parties in $\mathcal{L} \setminus  c$, and the joint view of $c$ in $VIEW_{ideal}$, will be identical to that in $VIEW_{real}$, namely		
	\[ (\sum \eta_1^{\mathcal{L} \setminus c} + \sum g_1^c )\equiv \sum g_1^\mathcal{L},\]
	and the information required by  ${SIM}_1$ is $z = \sum g_1^l\mid_{l \in \mathcal{L} \setminus  c}$.
	
	$\text{\textbf{SIM}}_{k+1} ( k\geq 1)$: ${SIM}_{k+1}$  is the simulator
	 for the $(k+1)^{th}$ training iteration.
	
	In $Real$ execution, $\sum g_{k+1}^\mathcal{L}$ is computed as
	\begin{equation}
		\label{equ:real}
		\begin{aligned}
			\sum g_{k+1}^\mathcal{L} & = \sum g_{k+1}^l (w^l_{k+1}, \xi_l)\mid_{l \in \mathcal{L}} = \sum g_{k+1}^l (w^l_{k} - \alpha_k \sum g_{k}^\mathcal{L}, \xi_l)\mid_{l \in \mathcal{L}}\\
			&=\sum g_{k+1}^l \big(w^l_{1} -\sum_{i=1}^{k} (\alpha_i \sum g_{i}^\mathcal{L}), \xi_l\big)\mid_{l \in \mathcal{L}}.
		\end{aligned}
	\end{equation}
	
	In $Ideal$ execution, since each $g_{k+1}^l(w^l_{k+1}, \xi_l)\mid_{l \in \mathcal{L}}$ is also computed from randomized $w_1^l$, ${SIM}_{k+1}$ can 
	produce a perfect simulation by running the parties $\mathcal{L} \setminus  c$ on  a set of pseudorandom vectors  $\eta_{k+1}^{\mathcal{L} \setminus c} =  \{\eta_{k+1}^l\}_{l \in \mathcal{L} \setminus c}$ in a way that
	\[\sum \eta_{k+1}^{\mathcal{L} \setminus c} = \sum \eta_{k+1}^l\mid_{l \in \mathcal{L} \setminus c} = \sum g_{k+1}^\mathcal{L} - \sum g_{k+1}^c = \sum g_{k+1}^l\mid_{l \in \mathcal{L} \setminus  c}.\]
	
	As such, the joint view of 
	$c$ in  $VIEW_{ideal}$, will be identical to that in $VIEW_{real}$
	\[ (\sum \eta_{k+1}^{\mathcal{L} \setminus c} + \sum g_{k+1}^c )\equiv \sum g_{k+1}^\mathcal{L},\]
	and the information required by  ${SIM}_{k+1}$  is $z = \sum g_{k+1}^l\mid_{l \in \mathcal{L} \setminus  c}$.

	By summarizing ${SIM}_{1}$ and ${SIM}_{k+1}$, the output of the simulator $VIEW_{ideal}$ of each training iteration is perfectly indistinguishable from the output of $VIEW_{real}$, and knowledge of $z$ is sufficient for the simulation, completing the proof.
	
\end{proof}

\subsection{Information exposed in traditional FL}
In this section, we demonstrate information exposed in traditional FL, 
including plain federated SGD, secure aggregated federated SGD, and differentially private federated SGD.

Let
\begin{equation}
	\label{equ:gltrafl}
	g^l_k(w_{k}, \xi_l)=\frac{1}{|\xi_{l}|}  \nabla F(w_{k}, \xi_{l}) = \frac{1}{|x_{l}|}  \nabla F(w_{k}, x_{l}, y_l) ,
\end{equation}
be the local gradient with respect to training dataset $\xi_l = (x_l, y_l)$. In traditional FL, $w_{k}$ is the public global model shared among the participants. 

For the sake of simplicity, we assume SGD is not generalized to mini-batch update, i.e., we have $\frac{1}{|x_{l}|} =1$. Then, Equation \ref{equ:gltrafl} can be written as,
\begin{equation}
	\label{equ:gltrasgd}
	g^l_k(w_{k}, \xi_l)=   \nabla F(w_{k}, x_{l}, y_l) ,
\end{equation}

According to the chain rule in calculus,
$\nabla F(w_{k}, x_{l}, y_l)$ is computed as $\frac{\partial{F(h(x_{l}w_{k}), y_l)}}{\partial{h(x_{l}w_{k})}}\frac{\partial{h(x_{l}w_{k})}}{\partial{(x_{l}w_{k})}}\frac{\partial{(x_{l}w_{k})}}{\partial{w_{k}}}$,
where $x_{l}w_{k}$ is matrix multiplication of training samples $x_{l}$ and $w_{k}$,
and $h$ is the hypothesis function which is determined by the learning model.
For example, in logistic regression, $h$ is usually a sigmoid function,
while in neural network, $h$ is a composite function that is known as forward propagation.
Let $\Delta(x_{l}w_{k}, y_l) = \frac{\partial{F}}{\partial{h}}\frac{\partial{h}}{\partial{(x_{l}w_{k})}}$,
and $\frac{\partial{(x_{l}w_{k})}}{\partial{w_{k}}}$ equals to $x_{l}^T$.
Then, Equation \ref{equ:gltrafl} can be written as
\begin{equation}
	\label{equ:gltrafldelta}
g^l_k(w_{k}, \xi_l) =  x_{l}^T \Delta(x_{l}w_{k}, y_{l}),
\end{equation}

\paragraph{Plain federated SGD}
In plain federated SGD, $w_k$ is updated as the following (by combing Equation \ref{equ:sgd} and \ref{equ:gltrafldelta})
\begin{equation}
	w_{k+1} \leftarrow  w_k - \alpha_k  x_{l}^T \Delta(x_{l}w_{k}, y_l),
\end{equation}
in which the local gradient  $ x_{l}^T \Delta(x_{l}w_{k}, y_l)$ is shared among the participants. As such, by knowing both $ x_{l}^T \Delta(x_{l}w_{k}, y_l)$ and $w_{k}$, the adversary is able to derive indicative information about $(x_l, y_l)$. For example, in linear regression, since  $ x_{l}^T \Delta(x_{l}w_{k}, y_l) = x_{l}^T (x_{l}w_{k}-y_l)$, the adversary is able to obtain 
\{$x_{l}^T x_{l}$, $x_{l}^T y_{l}$\}. 

\paragraph{Secure aggregated federated SGD}
In this category of traditional FL~\cite{bonawitz2017practical, mohassel2017secureml, zhang2020privcoll,  marc2019privacy, sharma2019confidential, liu2017oblivious}, 
the local gradients is protected by secure aggregation, and the global model $w_k$ is updated as
\begin{equation}
	\label{equ:secureagg}
	w_{k+1} \leftarrow  w_k - \alpha_k \sum\limits_l  g^l_k(w_{k}, \xi_l)\mid_{l \in \mathcal{D}},
\end{equation}
where $\mathcal{D} \subseteq \mathcal{L}$.
By combing Equation \ref{equ:secureagg}  and \ref{equ:gltrafldelta}, we have
\begin{equation}
		\label{equ:secureaggdelta}
		w_{k+1} \leftarrow  w_k - \alpha_k \sum\limits_l   x_{l}^T \Delta(x_{l}w_{k}, y_l) \mid_{l \in \mathcal{D}},
\end{equation}
in which the \emph{aggregated} gradient, $\sum\limits_l   x_{l}^T \Delta(x_{l}w_{k}, y_l) $, is shared among the participants. 

As the global model $w_k$ is also shared, 
by observing the changes of  the aggregated gradient during training iterations, i.e., $w_k - w_{k+1}$, the adversary is still able to obtain 
indicative information about $(x_l, y_l)$.

Let $x^{\mathcal{L}^\prime}$, $y^{\mathcal{L}^\prime}$ respectively denote the concatenated matrix of training samples $\{x_l\}_{l\in \mathcal{L}^\prime}$, and labels $\{y_l\}_{l\in \mathcal{L}^\prime}$ of any subset of participants $\mathcal{L}^\prime \subseteq \mathcal{L}$.
The following theorem shows that when executing secure aggregated federated SGD with the threshold $t$, the joint view of the participants in $c$ 
can be simulated by 
(1) the sum of the local gradients from the remaining honest nodes in $\mathcal{D}$, that is, $\sum g_k^l(w_{k}, \xi_l)\mid_{l \in \mathcal{D} \setminus  c}$ (2) and indicative information about $(x_l, y_l)$ in  $\mathcal{D}\setminus  c$, that is, ${x^{\mathcal{D} \setminus  c}}^T \Delta(x^{\mathcal{D} \setminus  c}w_{k}, y^{\mathcal{D} \setminus  c})$.
For example, in linear regression,
as ${x^{\mathcal{D} \setminus  c}}^T \Delta(x^{\mathcal{D} \setminus  c}w_{k}, y^{\mathcal{D} \setminus  c}) = {x^{\mathcal{D} \setminus  c}}^T (x^{\mathcal{D} \setminus  c}w_{k} - y^{\mathcal{D} \setminus  c})$,
the adversary is able to simulate   $\{{x^{\mathcal{D} \setminus  c}}^T x^{\mathcal{D} \setminus  c}, {x^{\mathcal{D} \setminus  c}}^T y^{\mathcal{D} \setminus  c}\}$.

\begin{theorem}
	\label{theorem_mpc}
	When executing secure aggregated federated SGD with the threshold $t$, there exists a simulator $\mathcal{S}$ such that for $\mathcal{L}$, $\mathcal{D}$ and $c$, with $\mathcal{D}\subseteq \mathcal{L}$, $c \subseteq \mathcal{L}$ and $\lvert c \lvert \leq t$, the output of $\mathcal{S}$ from $VIEW_{ideal}$ is perfectly indistinguishable from the output of  $VIEW_{real} $, namely
	\begin{align*}
		VIEW_{real} &(\sum\limits_l  g^l_k(w_{k}, \xi_l)\mid_{l \in \mathcal{D}}, t, \mathcal{P}, c) \\
		&\equiv VIEW_{ideal} (g^l_k(w_{k}, \xi_l)\mid_{l \in c}, z_1, z_2, t, \mathcal{F}_p, c)
	\end{align*}
	where
	\[z_1 = \sum g_k^l(w_{k}, \xi_l)\mid_{l \in \mathcal{D} \setminus  c}, z_2 = {x^{\mathcal{D} \setminus  c}}^T \Delta(x^{\mathcal{D} \setminus  c}w_{k}, y^{\mathcal{D} \setminus  c})\]
\end{theorem}

\begin{proof}
	$\text{\textbf{SIM}}_{k} ( k\geq 1)$: ${SIM}_{k}$  is the simulator
	for the $k^{th}$ training iteration.
	
	In $Real$ execution, $\sum g_{k}(w_{k}, \xi_l)\mid_{l \in \mathcal{D}}$ is computed as (with Equation \ref{equ:secureagg} and \ref{equ:secureaggdelta})
	\begin{equation}
		\label{equ:realmpc}
		\begin{aligned}
			\sum g^l_{k}(w_{k}, \xi_l)\mid_{l \in \mathcal{D}} = \sum\limits_l   x_{l}^T \Delta(x_{l}w_{k}, y_l) \mid_{l \in \mathcal{D}} = \frac{1}{\alpha} (w_k-w_{k+1})
		\end{aligned}
	\end{equation}
	
	In $Ideal$ execution, since $w_k$ is shared among the participants, by computing $w_k-w_{k+1}$, ${SIM}_{k}$ can
	produce a perfect simulation by running the parties $\mathcal{D} \setminus  c$ on
	\[ \sum g^l_{k}(w_{k}, \xi_l)\mid_{l \in \mathcal{D}\setminus  c}, \textit{or, } {x^{\mathcal{D} \setminus  c}}^T \Delta(x^{\mathcal{D} \setminus  c}w_{k}, y^{\mathcal{D} \setminus  c}),\]
	
	As such, the joint  view of 
	$c$ in  $VIEW_{ideal}$, will be identical to that in $VIEW_{real}$, since
	\[\sum g^l_{k}(w_{k}, \xi_l)\mid_{l \in \mathcal{D}\setminus  c}+\sum g^l_{k}(w_{k}, \xi_l)\mid_{l \in   c} \equiv \sum  g^l_{k}(w_{k}, \xi_l)\mid_{l \in \mathcal{D}}\]
	and,
	\begin{align*}
		&{x^{\mathcal{D} \setminus  c}}^T \Delta(x^{\mathcal{D} \setminus  c}w_{k}, y^{\mathcal{D} \setminus  c}) +  {x^{c}}^T \Delta(x^{c}w_{k}, y^{c})\\
		& \equiv {x^{\mathcal{D} }}^T \Delta(x^{\mathcal{D} }w_{k}, y^{\mathcal{D} }) \\
		& \equiv \sum\limits_l  x_{l}^T \Delta(x_{l}w_{k}, y_l) \mid_{l \in \mathcal{D}}
	\end{align*}
	Thus the information required by  ${SIM}_{k}$  is $z_1 = \sum g^l_{k}(w_{k}, \xi_l)\mid_ {l \in {\mathcal{D}\setminus  c}}$ and $z_2 = {x^{\mathcal{D} \setminus  c}}^T \Delta(x^{\mathcal{D} \setminus  c}w_{k}, y^{\mathcal{D} \setminus  c})$.
	
	Together, we have  $\{\sum g_k^l(w_{k}, \xi_l)\mid_{l \in \mathcal{D} \setminus  c}, {x^{\mathcal{D} \setminus  c}}^T \Delta(x^{\mathcal{D} \setminus  c}w_{k}, y^{\mathcal{D} \setminus  c})\}$ being the  information that the adversary can learn during the execution.
\end{proof}

\paragraph{Differentially private federated SGD}
In most differentially private federated SGD, the local gradients are protected by additive noise mechanism as
\begin{equation}
	w_{k+1} \leftarrow w_k - \alpha_k  (g^l_k(w_{k}, \xi_l)+ \mathcal{N}_k),
\end{equation}
where $\mathcal{N}_k$ denote the noise added at the $k^{th}$ iteration. By combing Equation \ref{equ:gltrafldelta}, it can be written as
\begin{equation}
	w_{k+1} \leftarrow w_k - \alpha_k ( x_{l}^T \Delta(x_{l}w_{k}, y_l)+ \mathcal{N}_k),
\end{equation}

The information exposed among participants is $x_{l}^T \Delta(x_{l}w_{k}, y_l)+ \mathcal{N}_k$, and the additive noise $\mathcal{N}_k$ prevent one from directly deriving $x_{l}^T \Delta(x_{l}w_{k}, y_l)$ by subtracting $w_k$ and $w_{k+1}$.
However, with repeated access to the datasets  during training epochs, $\epsilon$ (the parameter  of privacy loss)  accumulates, i.e., privacy degrades,  as the effect of added noise being canceled out~\cite{dwork2014algorithmic, abadi2016deep}.

\subsection{Enhanced privacy over traditional FL}
Traditional FL requires all participants to update the same global model during the training process. Every participant thus sees the identical intermediate results, as the same aggregated gradients are shared. This is the root cause of most privacy threats against FL. \short breaks the mode of single global model, by introducing random variation among the proprietary global models of the participants. The variation hides each global model from other participants, such that the privacy is enhanced in general.

\begin{table*}[ht]
	\caption{A summary of the exposed information in traditional FL and \short }
		\resizebox{\textwidth}{!}{
			\begin{tabular}{|p{0.15\linewidth}|p{0.33\linewidth}|p{0.27\linewidth}|p{0.35\linewidth}|}
				\hline
				\textbf{Techniques}
				& \textbf{The observation of an honest-but-curious white-box adversary  during training iterations}
				& \textbf{Exposed indicative information about $(x_l,y_l)$  (in the example of linear regression)}
				& \textbf{Boosted privacy of \short over traditional FL}           \\ \hline
				   &&& \\ [-0.8em]
				\textbf{Plain federated SGD}
				& Local gradients: $g_k^l(w_{k}, \xi_l)$, which equals $x_{l}^T \Delta(x_{l}w_{k}, y_l)$
				& \{$x_{l}^T x_{l}$, $x_{l}^T y_{l}$\}
				& In plain federated SGD,  indicative information about the local training dataset can be observed. \\ \hline
				    &&&  \\ [-1em]
				\textbf{Secure aggregated federated SGD}
				& Sum of the local gradients from the remaining honest participants: $\sum g_k^l(w_{k}, \xi_l)\mid_{l \in \mathcal{D} \setminus  c}$, and with a shared $w_k$, it equals  ${x^{\mathcal{D} \setminus  c}}^T \Delta(x^{\mathcal{D} \setminus  c}w_{k}, y^{\mathcal{D} \setminus  c})$ (Theorem \ref{theorem_mpc}).
				& $\{{x^{\mathcal{D} \setminus  c}}^T x^{\mathcal{D} \setminus  c}$, ${x^{\mathcal{D} \setminus  c}}^T y^{\mathcal{D} \setminus  c}\}$
				&  With the shared global $w_k$, indicative information about the concatenated training datasets from  honest participants  can be observed. 
				\\ \hline
				    &&&  \\ [-0.8em]
				\textbf{Differentially private federated SGD (additive noise based mechanism)}
				& Perturbed local gradients: $g_k^l(w_{k}, \xi_l)+ \mathcal{N}_k$, which equals $x_{l}^T \Delta(x_{l}w_{k}, y_l)+ \mathcal{N}_k$
				&  Not applicable with the single access to $(x_l,y_l)$. However, with repeated access to $(x_l,y_l)$ during epochs, the adversary is able to get a more accurate estimate of $x_{l}^T \Delta(x_{l}w_{k}, y_l)$, which can be used to obtain \{$x_{l}^T x_{l}$, $x_{l}^T y_{l}$\}.
				& The decay of privacy with increasing number of training epochs is one of the limitations of most additive-noise based differentially private learning.
				\\ \hline
				    &&&  \\ [-0.8em]
				\textbf{\short}
				&  Sum of the local gradients from the remaining honest participants: $\sum g_k^l(w^l_{k}, \xi_l)\mid_{l \in \mathcal{L} \setminus  c}$  (Theorem \ref{theorem_party}).
				&  Not applicable.
				&  In \short, each local gradient is computed from a different and private $w_k^l$, and only the sum of local gradients is exposed during the training. As such, (1) the indicative information exposed in the secure aggregation cannot be derived in \short; (2) the randomness introduced in each $w_k^l$ hides the information about local gradient throughout the training, and thus privacy does not decay.\\ \hline
				
			\end{tabular}	
		}
	\label{tab:priv}
\end{table*}
		
In Table \ref{tab:priv}, we summarize the  privacy enhancement  from the perspective of adversary's observation, i.e, the exposed information to the adversary during the optimization.
In traditional FL, the sharing  of global model $w_k$, even if the local gradients are protected, is still subject to information leakage about the original dataset.
By eliminating the sharing of $w_k$ and letting each participant confine its own $w_k^l$, \short achieves boosted privacy over traditional privacy-preserving FL.
(1) Compared to secure aggregated FL in which indicative information about original datasets can be observed via the sharing of $w_k$ (Theorem \ref{theorem_mpc}), the variation among $w_k^l$ in \short prevents such information from disclosure, and guarantees that only the sum of local gradients is exposed during the optimization (Theorem \ref{theorem_party}). 
(2) Compared to differential privacy mechanism in which privacy decays with the increasing training epochs, the variation introduced to each $w_k^l$  hides the local gradients throughout the whole training process, and thus retains privacy regardless of the number of training epochs.




\section{Case study: \codename for a N-layer neural network}
	\label{sec:cgdnn}
\begin{figure}[!t]
	\centering
	\includegraphics[width=2.5in]{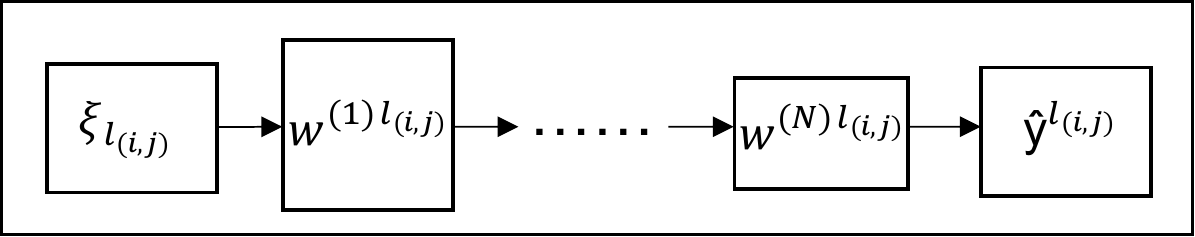}
	\caption{Local data and confined model of a N-layer neural network in participant $l_{(i,j)}$ ($i \in m^h$, $j \in m^v$).}
	\label{fig:cgdnn}
	\vspace{-0.1cm}
\end{figure}

\short can be applicable to any machine learning algorithms that use gradient descent for optimization. In this section,  
we apply it to a N-layer neural network to demonstrate its usability. 
We assume that a centralized dataset $\xi$ is horizontally and vertically partitioned and distributed to $m$ participants where $m = (m^h \times m^v)$, i.e., the number of horizontal partitions multiplied by the number of vertical partitions.
The participant $l$ owns a private part of the training dataset, denoted by $\xi_{l_{(i,j)}}$ ($i \in [1, m^h]$, $j \in [1, m^v]$), 
as well as its confined model parameters, denoted by $w_k^{(1)l_{(i,j)}}$,...,$w_k^{(N)l_{(i,j)}}$. Since each participant owns different confined models and proportion of the dataset, the training prediction $\widehat{y}^{l_{(i,j)}}$  is also different and kept confined in its owner~(shown in Figure \ref{fig:cgdnn}).

Algorithm \ref{alg:cgdnn} presents the detailed algorithm. 
The participant $l$ first randomly initializes its confined model $w_1^{(1)l_{(i,j)}}$,...,$w_1^{(N)l_{(i,j)}}$ (line 3). 
The size of the model in the first layer  is $w^{(1)l_{(i,j)}} \in \mathbb{R}^{d_{l_j}\times H_1}$, where $d_{l_j}$ is the number of features in $\xi_{l_{(i,j)}}$, and the size of models in the remaining layers 
is $w^{(r)l_{(i,j)}} \in \mathbb{R}^{H_{r-1}\times H_r}$ ($r \in [2,N]$).
It is possible that different participants have different size of  $w^{(1)l_{(i,j)}}$ because the number of features $d_{l_j}$ held by each participant may differ, 
while 
$w^{(r)l_{(i,j)}}$ ($r \in [2,N]$) keep the same size in each participant.

Next, we detail the training process in each iteration. 
In the forward propagation, each participant separately computes the output of each layer $a_k^{(1)l_{(i,j)}},...,a_k^{(N)l_{(i,j)}} $ based on its own confined models  (line 5 to 11).
In the backward propagation, each participant solely computes the local gradient of each layer $g_k^{(N)l_{(i,j)}},...g_k^{(1)l_{(i,j)}}$ which is computed from its own private dataset and confined model (line 12 to 18). 
Then, they securely evaluate the sum of local gradients from the $N^{th}$ layer to the $2^{nd}$ layer (line 19 to 21). 
For the first layer gradient, since the size of $w^{(1)l_{(i,j)}}$ can be different,  the sum of the local gradients $g_k^{(1)l_j}$ is therefore taken among the vertically partitioned participant groups $l_j = \{l_{(1,j)},...,l_{(m^h,j)}\}$ (line 22 to 24).
In the descent, after choosing a learning rate $\alpha_k$ (line 25), the first layer is updated by taking a descent step of $\alpha_k g_k^{(1)l_j}$ (line 26 to 28), and the remaining layers are updated by taking a descent step of $\alpha_k g_k^{(r)} $ ($r \in [2,N]$) (line 29 to 31). 

\begin{algorithm}[]
	\caption{\codename of a N-layer fully-connected neural network for $m = (m^h \times m^v)$ participants}
	\label{alg:cgdnn}
	\begin{algorithmic}[1]
		\State{\textbf{Input:} Local training data $\xi_{l_{(i,j)}}$ ($i \in m^h$, $j \in m^v$), activation functions of $N$ layers: $\sigma^{(1)}...\sigma^{(N)}$, cost function $J$, the number of  training iterations $T$}
		\State{\textbf{Output:} Confined global model parameters of $N$ layers: $w_*^{(1)l_{(i,j)}}$,...,$w_*^{(N)l_{(i,j)}}$ ($i \in m^h$, $j \in m^v$) }
		\State{\textbf{Initialize:} $k\leftarrow 1$, each participant $l_{(i,j)}$ randomizes its own $w_1^{(1)l_{(i,j)}}$,...,$w_1^{(N)l_{(i,j)}}$}
		\While{$k \leq T$}
		\Statex{\textbf{Forward propagation} }
		\State{\quad\textbf{for all} participants $l_{(i,j)}$ \textbf{do in parallel}}
		\State{\qquad  $z_k^{(1)l_{(i,j)}} \leftarrow \xi_{l_{(i,j)}} w_k^{(1)l_{(i,j)}}$ }
		\State{\qquad  $a_k^{(1)l_{(i,j)}} \leftarrow \sigma^{(1)}(z_k^{(1)l_{(i,j)}}) $ }
		\State{\qquad  $z_k^{(2)l_{(i,j)}} \leftarrow a_k^{(1)l_{(i,j)}} w_k^{(2)l_{(i,j)}}$}
		\State{\qquad  \qquad\qquad\qquad...... }
		\State{\qquad  $a_k^{(N)l_{(i,j)}} \leftarrow \sigma^{(N)}( z_k^{(N)l_{(i,j)}})) $, $\widehat{y}^{l_{(i,j)}} \leftarrow a_k^{(N)l_{(i,j)}} $  }
		\State{\quad\textbf{end for}}
		\Statex{\textbf{Backward propagation} }
\Statex\quad{\textbf{Calculate the local gradients of each layer} }
\State\quad{\textbf{for all} participants $l_{(i,j)}$ \textbf{do in parallel}}
\Statex{
	\qquad\qquad $\delta_k^{(N)l_{(i,j)}} \leftarrow  \frac{\partial J(\widehat{y}^{l_{(i,j)}} , y^{l_{(i,j)}}) }{\partial a_k^{(N)l_{(i,j)}}} \frac{\partial a_k^{(N)l_{(i,j)}}}{\partial z_k^{(N)l_{(i,j)}}}$ \\\qquad\qquad $g_k^{(N)l_{(i,j)}} \leftarrow \delta_k^{(N)l_{(i,j)}} \frac{\partial  z_k^{(N)l_{(i,j)}}}{\partial w_k^{(N)l_{(i,j)}}}$ }
\State\quad\qquad{\textbf{for}  $r= N-1,...,1$ \textbf{do}}
\State\qquad\qquad{$\delta_k^{(r)l_{(i,j)}} \leftarrow  \delta_k^{(r+1)l_{(i,j)}} \frac{\partial z_k^{(r+1)l_{(i,j)}}}{\partial a_k^{(r)l_{(i,j)}}} \frac{\partial a_k^{(r)l_{(i,j)}}}{\partial  z_k^{(ra)l_{(i,j)}}}$}
\State\qquad\qquad{$g_k^{(r)l_{(i,j)}} \leftarrow \delta_k^{(r)l_{(i,j)}}  \frac{\partial  z_k^{(r)l_{(i,j)}}}{\partial w_k^{(r)l_{(i,j)}}}$}
\State\quad\qquad{\textbf{end for}}		
\State{\quad\textbf{end for}}
\Statex\quad{\textbf{Securely evaluate the sum of local gradients} }
\State\quad{\textbf{for all} participants $l_{(i,j)}$ \textbf{do in parallel}}
\State{
	\qquad\qquad $g_k^{(r)} \leftarrow \sum_{i,j=1}^{m^h,m^v}g_k^{(r)l_{(i,j)}}$, ($r \in [2,N]$)}  
\State{\quad\textbf{end for}}
\State{\quad\textbf{for all} groups $l_j = \{l_{(1,j)},...,l_{(m^h,j)}\}$ \textbf{do in parallel}}
\State{\qquad  Evaluate the sum of first layer:  $g_k^{(1)l_j} \leftarrow \sum_{i=1}^{m^h}g_k^{(1)l_{(i,j)}}$ }
\State{\quad\textbf{end for}}
		\Statex{\textbf{Descent} }
\State{\quad Choose a stepsize: $\alpha_k$}
\State{\quad\textbf{for all} groups $l_j = \{l_{(1,j)},...,l_{(m^h,j)}\}$ \textbf{do in parallel}}
\State{
	\qquad\qquad\qquad $w_{k+1}^{(1)l_{(i,j)}} \leftarrow w_{k}^{(1)l_{(i,j)}} - \alpha_k g_k^{(1)l_j} $}
\State{\quad\textbf{end for}}
\State{\quad\textbf{for all} participants $l_{(i,j)}$ \textbf{do in parallel}}
\State{
	\qquad\qquad\qquad $w_{k+1}^{(r)l_{(i,j)}} \leftarrow w_{k}^{(r)l_{(i,j)}} - \alpha_k g_k^{(r)} $, ($r \in [2,N]$)}
\State{\quad\textbf{end for}}
\State{$k \leftarrow k+1 $}		
\EndWhile
\end{algorithmic}
\end{algorithm}


\section{Experiments} \label{sec:experiment}

We implement \short and evaluate its  performance of model accuracy. 
It is implemented using C++, and we use Eigen library \cite{eigenweb} to handle matrix operations and ZeroMQ library \cite{hintjens2013zeromq} for distributed messaging.
Our evaluation focuses on the performance of \short in terms of validation loss and accuracy, and the influence of the factors~(the parameters $\delta$ and $\mu$) on its performance.

\subsection{Performance Evaluation} \label{subsec:performace}
We evaluate validation loss and accuracy on two popular benchmark datasets: MNIST~ \cite{lecun-mnisthandwrittendigit-2010} and CIFAR-10~\cite{krizhevsky2009learning}.
We compare \short with the training on aggregated data~(referred to as the \emph{centralized training}), the training on each participant's local data only~(referred to as the \emph{local training}), and FL with differential privacy~(referred to as the \emph{DP-FL}).
\short is expected to outperform the local training and DP-FL, and approach that of the centralized training.
The settings of all trainings are listed below.
\begin{itemize}
	\item \textbf{\short training.} The model is trained using Algorithm \ref{alg:cgd} on the dataset which is partitioned and distributed to $m$ participants, where $m = (m^h \times m^v)$. 
	We observe that the performance of all confined models are proximal, and thus we report the worst performance in this section. 
	\item \textbf{Centralized training}~(the baseline). We take the performance of centralized training as the baseline. In this setting, the training is on the entire dataset using the batch update.
	\item \textbf{Local training.} In this setting, the data and model are distributed to $m$ participants in the same way as in \short.
 Each participant separately trains their local models from their local data, without sharing the gradients.
    \item \textbf{DP-FL}. We take the scheme proposed by Abadi et al.~\cite{abadi2016deep}, one of the state-of-the-art differentially private FL schemes, in our experiments.
        Due to the different setting of network architecture and hyper-parameters, its centralized baseline is slightly different from ours.
        Therefore, we report the margin between its performance and the performance of its baseline.
\end{itemize}

\subsubsection{MNIST}
Our first set of experiments are conducted on the standard MNIST dataset which is a benchmark for handwritten digit recognition.
It has $60,000$ training samples and $10,000$ test samples, each with $784$ features representing $28 \times 28$ pixels in the image.
We use a fully-connect neural network (FNN) 
with ReLU of $256$ units and softmax of 10 classes 
with cross-entropy loss.

\begin{figure}[]%
	\centering
	\subfloat[$m=(10 \times 7)$]{{\includegraphics[width=4cm]{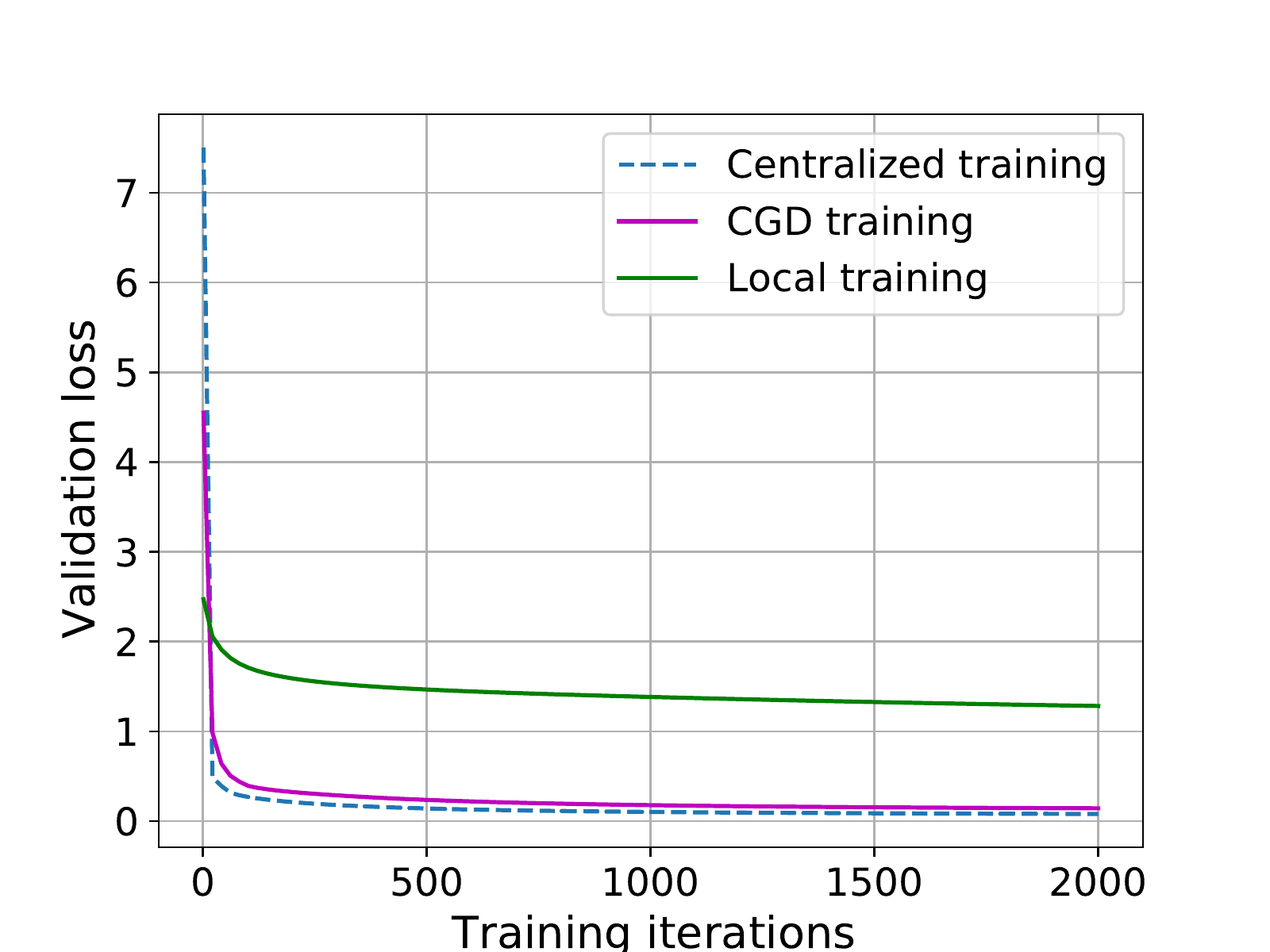} \label{fig:fixrate_10*7_loss}}}%
	\subfloat[$m=(10 \times 7)$]{{\includegraphics[width=4cm]{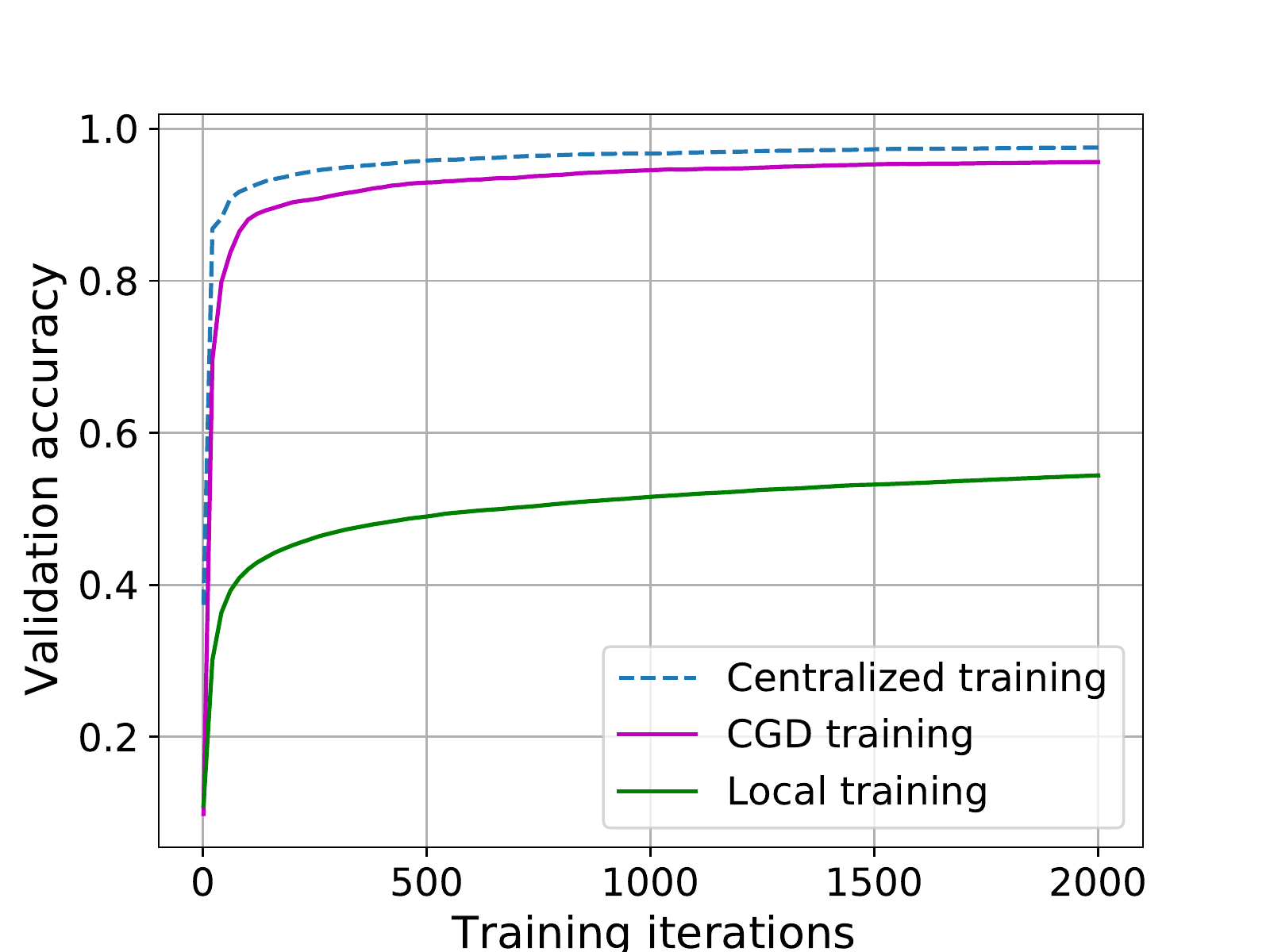} \label{fig:fixrate_10*7_acc}}}%
\quad
	\subfloat[$m=(100 \times 49)$]{{\includegraphics[width=4cm]{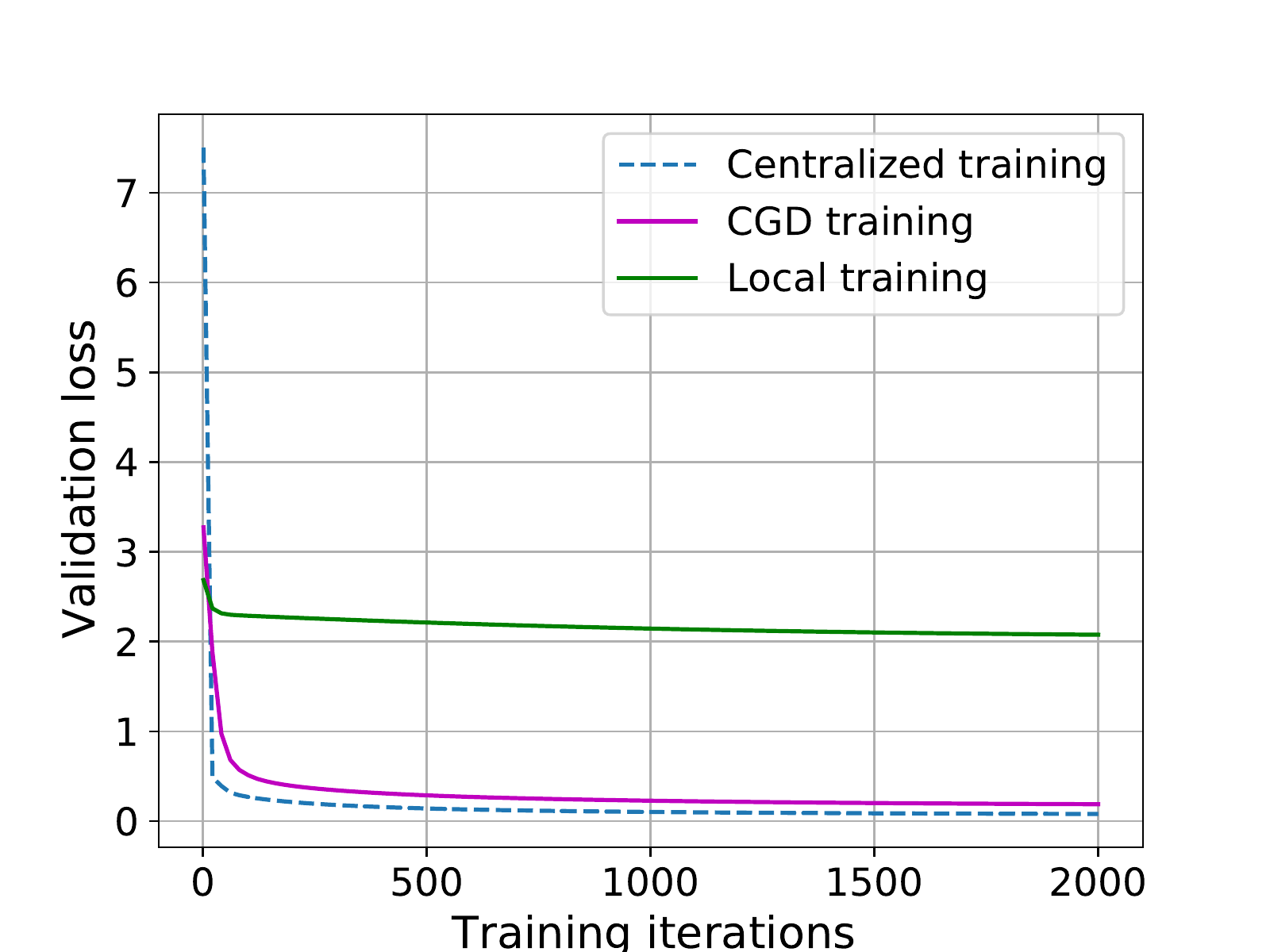} \label{fig:fixrate_100*49_loss}}}%
	\subfloat[$m=(100 \times 49)$]{{\includegraphics[width=4cm]{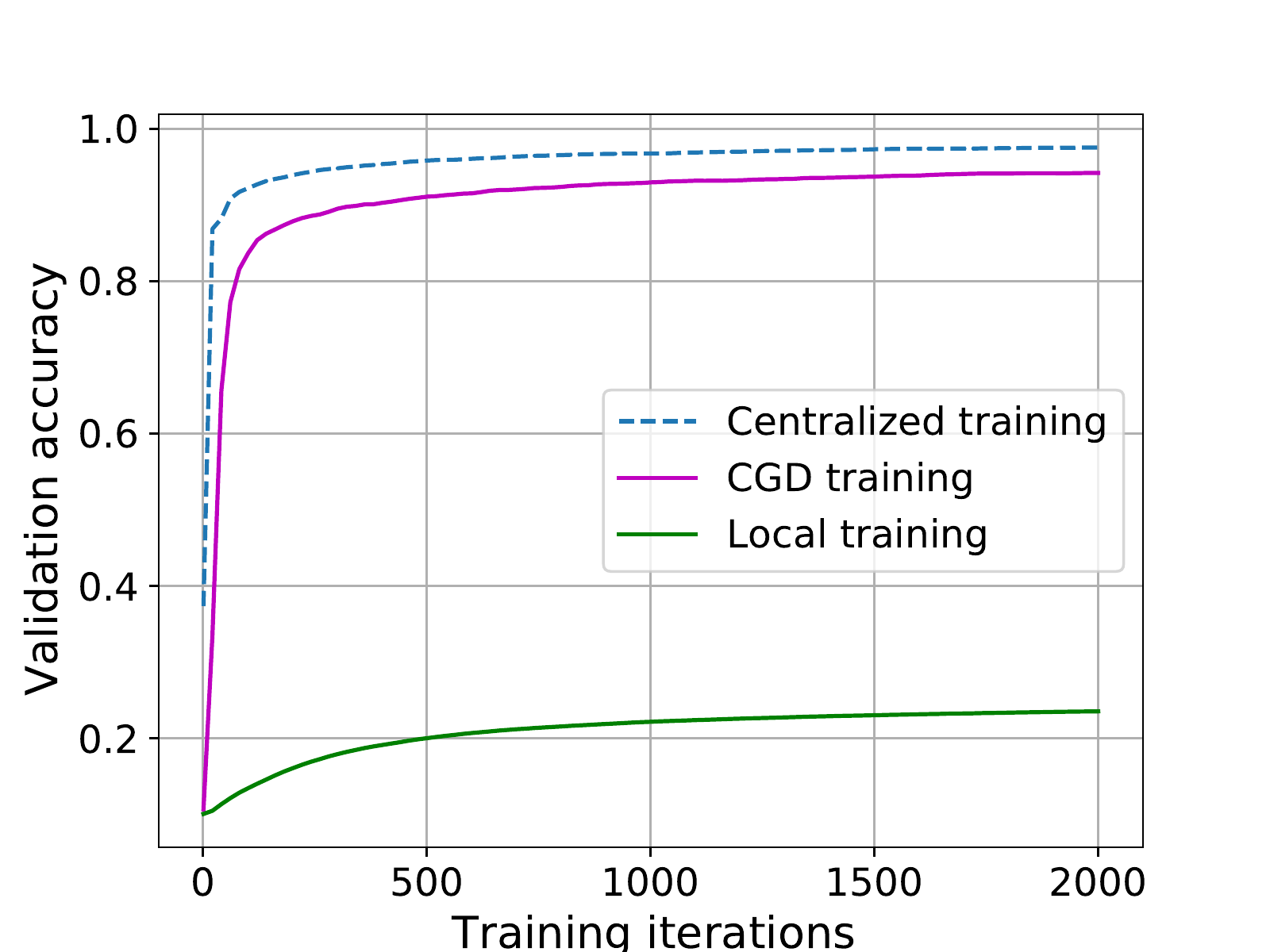} \label{fig:fixrate_100*49_acc}}}%
\quad
	\subfloat[$m=(1000 \times 112)$]{{\includegraphics[width=4cm]{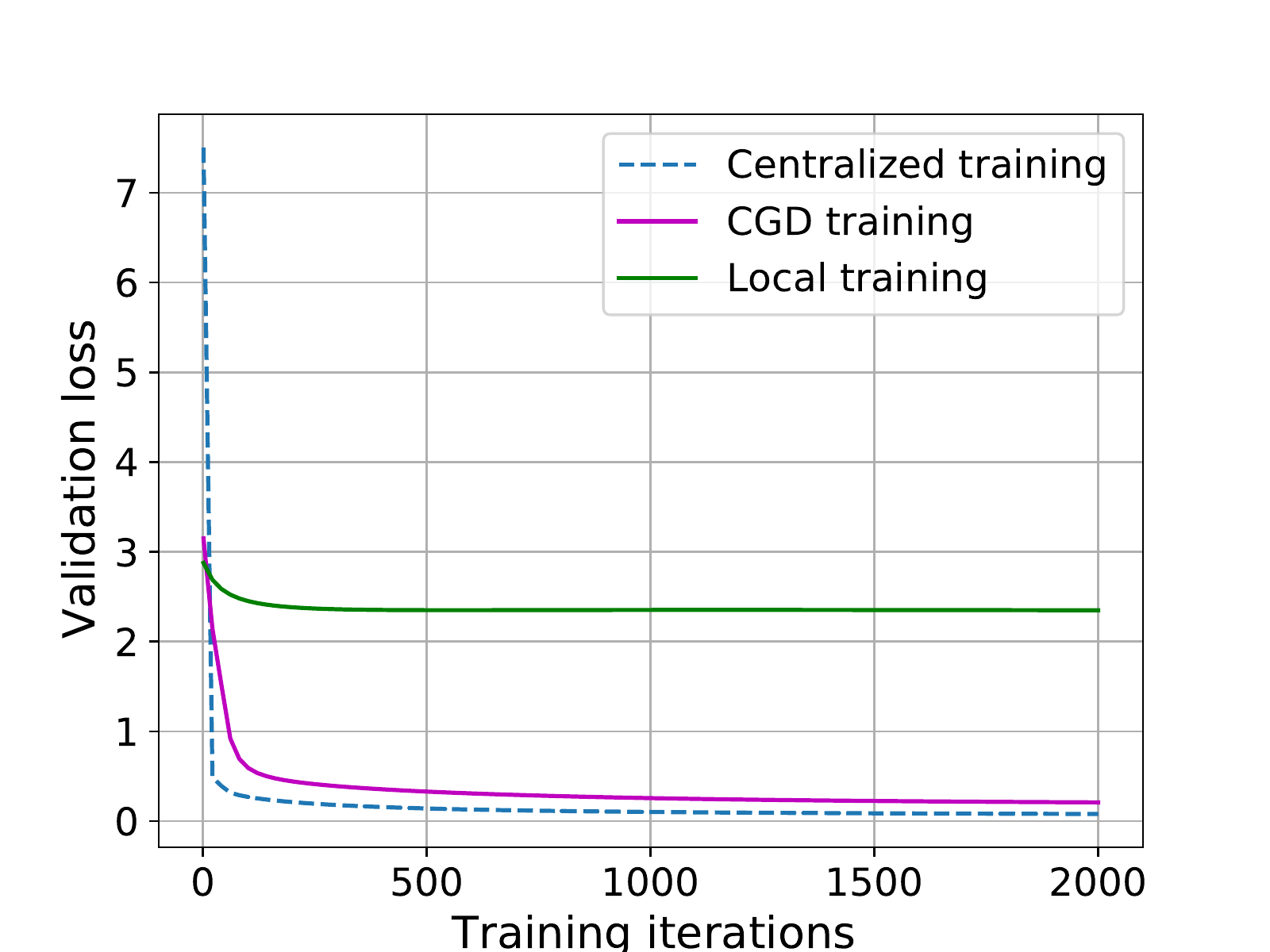} \label{fig:fixrate_1000*112_loss}}}%
	\subfloat[$m=(1000 \times 112)$]{{\includegraphics[width=4cm]{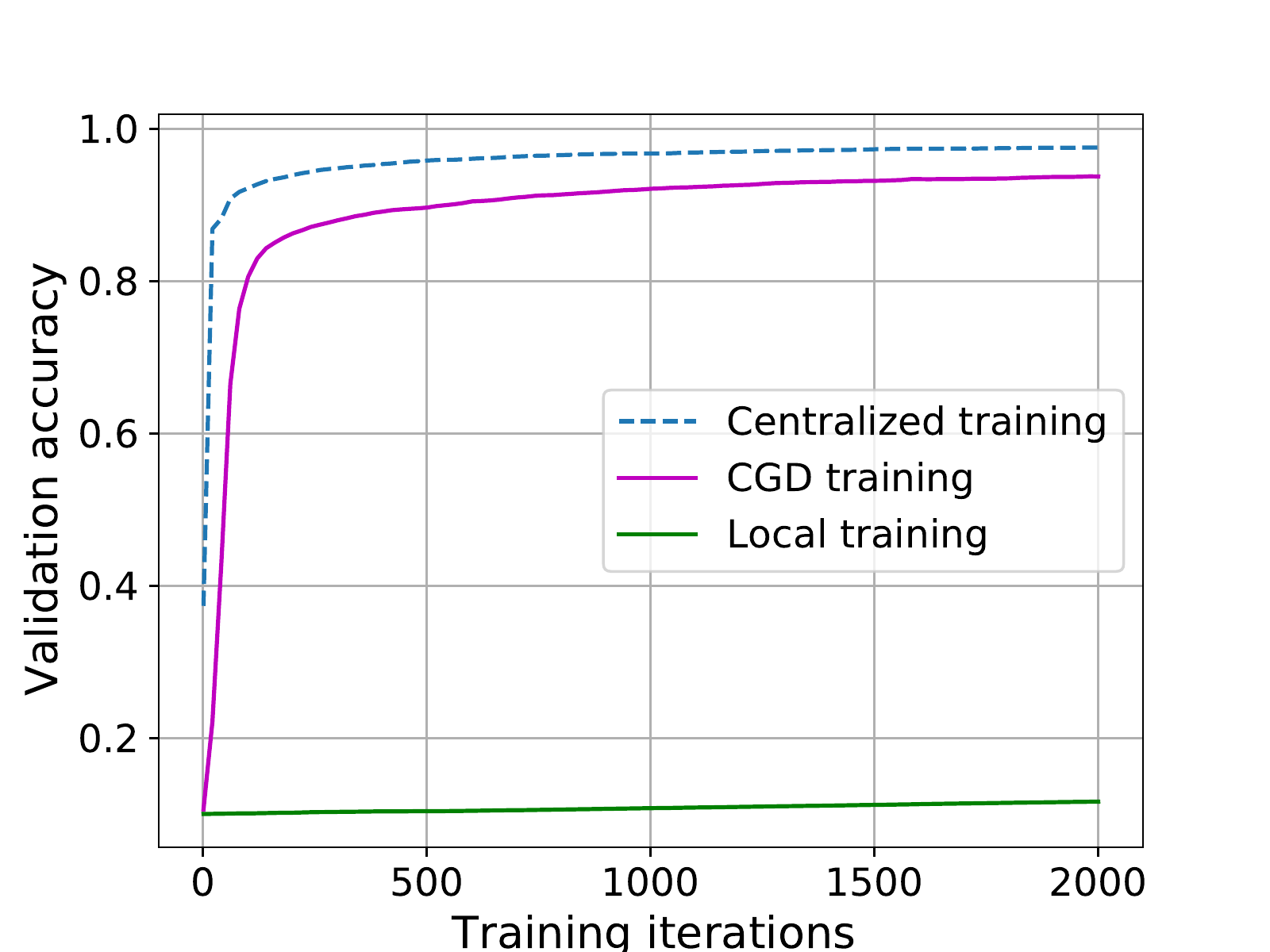} \label{fig:fixrate_1000*112_acc}}}%
	%
	\caption{Results on the validation loss and accuracy for different number of participants on the MNIST dataset.}%
	\label{fig:fixrate}
\end{figure}

We conduct our experiments with \short's default settings:
\begin{itemize}
    \item $\mu = 0$, leading to a fixed learning rate which is in line with most machine learning algorithms,
    \item $\delta = 0.1$ (cf. Equation \ref{equ:delta}), and
    \item the standard weight initialization scheme based on the Gaussian distribution of mean 0 and variance 1.
\end{itemize}
Figure \ref{fig:fixrate} summarizes the performance with varying number of participants.
In general, \short stays close to the centralized baseline in both validation loss and accuracy, and as expected, significantly outperforms the local training.
In the worst case of $m = (1000 \times 112)$ when \short contains the greatest number of participants, i.e.,  each participant owns a small proportion consisting of $60$ samples with $7$ features, \short still achieves $0.209$ and $93.74\%$ in validation loss and accuracy respectively, close to $0.081$ and $97.54\%$ of the centralized baseline.  In contrast, the performance of local training declines to $2.348$ and $11.64\%$. Table~\ref{tab:fixrate} lists the detailed performance comparison with both centralized and local trainings, and Table~\ref{tab:dpmnist} lists the comparison with DP-FL.  More results are given with varying number of participants in Appendix \ref{apx:experiment} (Figure \ref{fig:fixratesup} and Table \ref{tab:fixratesup}).

\begin{table}[t]
	\caption{Performance comparison on MNIST with default settings.} 
	\label{tab:fixrate}
	\resizebox{0.45\textwidth}{!}{	
\begin{tabular}{|l|l|l|l|l|}
	\hline
	& \multicolumn{2}{c|}{Validation loss} & \multicolumn{2}{c|}{Validation accuracy} \\ \hline
	Centralized        & \multicolumn{2}{c|}{0.081}           & \multicolumn{2}{c|}{97.54\%}             \\ \hline \hline
	& \short             & Local training           & \short               & Local training              \\ \hline
	m=($10 \times 7$)   & 0.143             & 1.283            & 96.5\%              & 54.39\%               \\ \hline
	m=($100 \times 49$)  & 0.189             & 2.076            & 94.21\%             & 23.53\%            \\ \hline
	m=($1000 \times 112$) & 0.209             & 2.348            & 93.74\%             & 11.64\%            \\ \hline
\end{tabular}
}
\end{table}

\begin{table}
	\centering
\caption{Comparison between \short and traditional FL via differential privacy in terms of the difference of validation  accuracy to the centralized baseline on MNIST.}
\label{tab:dpmnist}
	\resizebox{0.35\textwidth}{!}{
\begin{tabular}{|l|l|}
	\hline
	\multicolumn{2}{|c|}{\short}                         \\ \hline
	Participants & \shortstack[l]{Difference to the baseline}  \\ \hline
	m=($10 \times 7$)            & 1.0\%                                           \\ \hline
	m=($100 \times 49$)           & 3.3\%                                         \\ \hline
	m=($1000 \times 112$)            & 3.8\%                                           \\ \hline\hline
	\multicolumn{2}{|c|}{Traditional FL via DP~\cite{abadi2016deep}}             \\ \hline
Noise levels                                & \shortstack[l]{Difference to the baseline$^1$} \\ \hline
$\epsilon = 8$ (small noise)   & 1.3\%                      \\ \hline
 $\epsilon = 2$ (medium noise)  & 3.3\%                      \\ \hline
 $\epsilon = 0.5$ (large noise) & 8.3\%                      \\ \hline
\end{tabular}
}
\footnotesize{\qquad\qquad\qquad\qquad\qquad$^1$ The centralized baseline of validation accuracy on the MINIST in~\cite{abadi2016deep} is 98.3\%.}
\end{table}

\subsubsection{CIFAR-10}
\begin{figure}[]%
	\centering
	\subfloat[$m=(10 \times 2)$]{{\includegraphics[width=4cm]{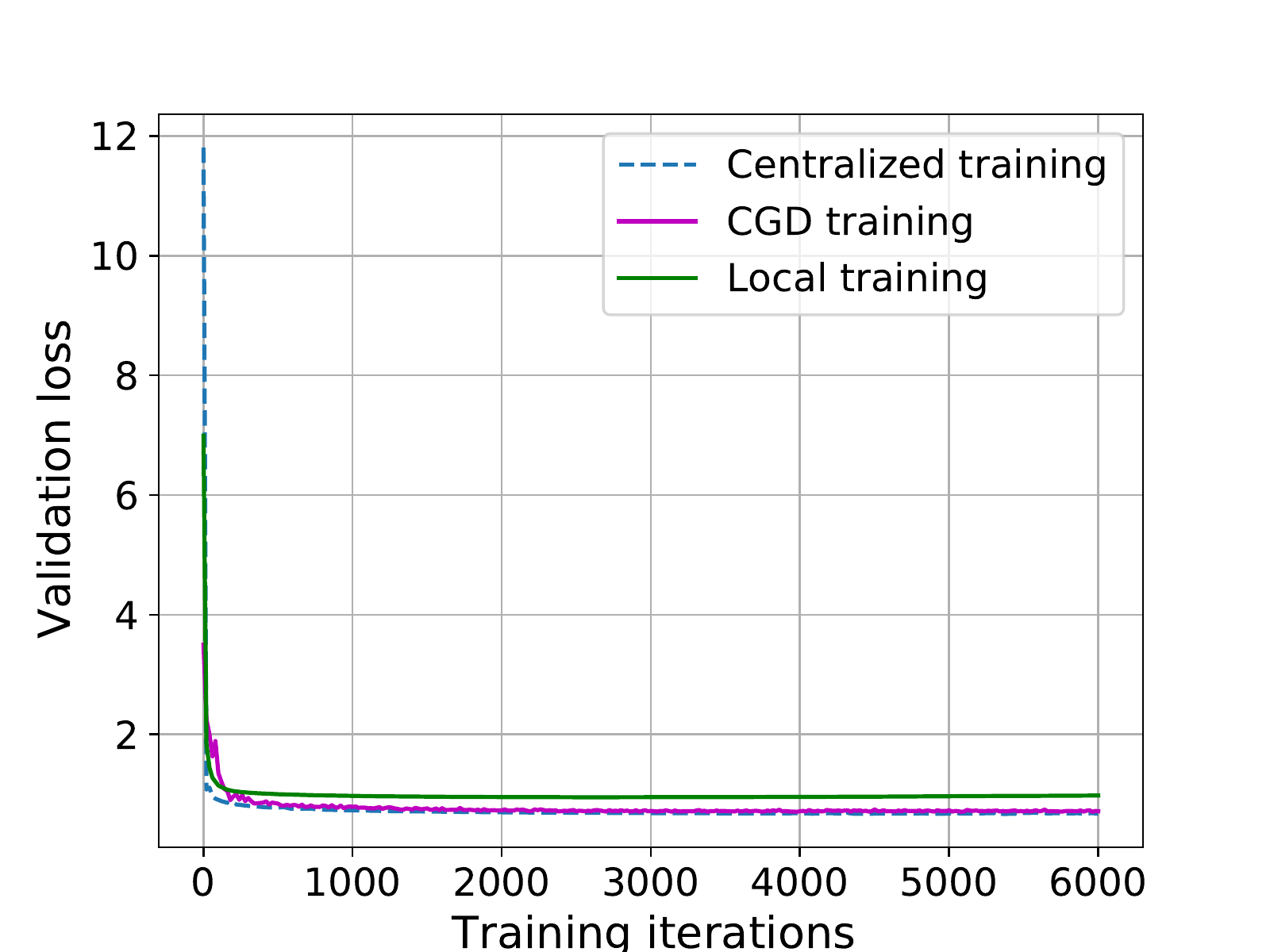} \label{fig:cifar_10_loss}}}%
	\subfloat[$m=(10 \times 2)$]{{\includegraphics[width=4cm]{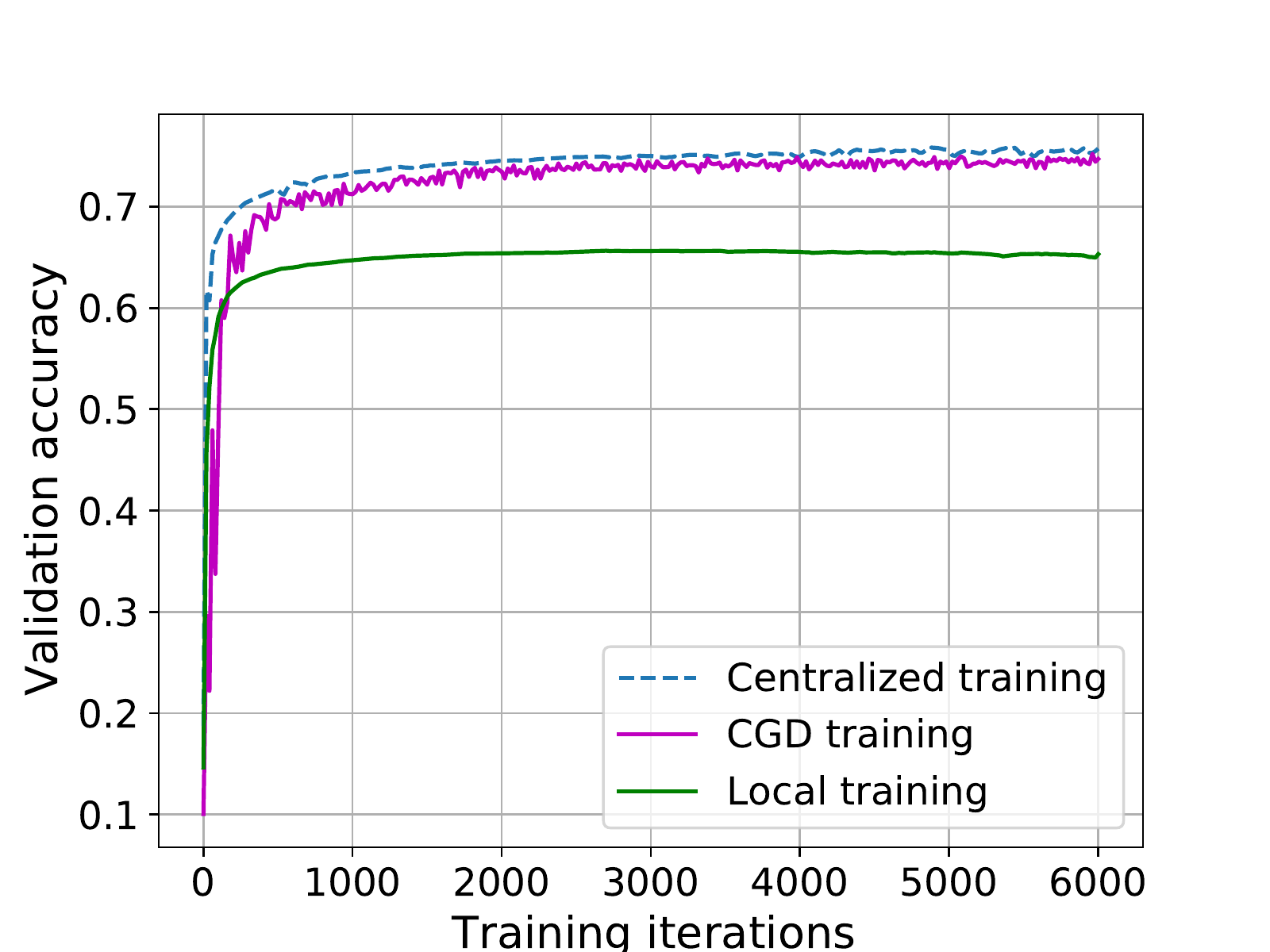} \label{fig:cifar_10_acc}}}%
	\quad
	\subfloat[$m=(100 \times 16)$]{{\includegraphics[width=4cm]{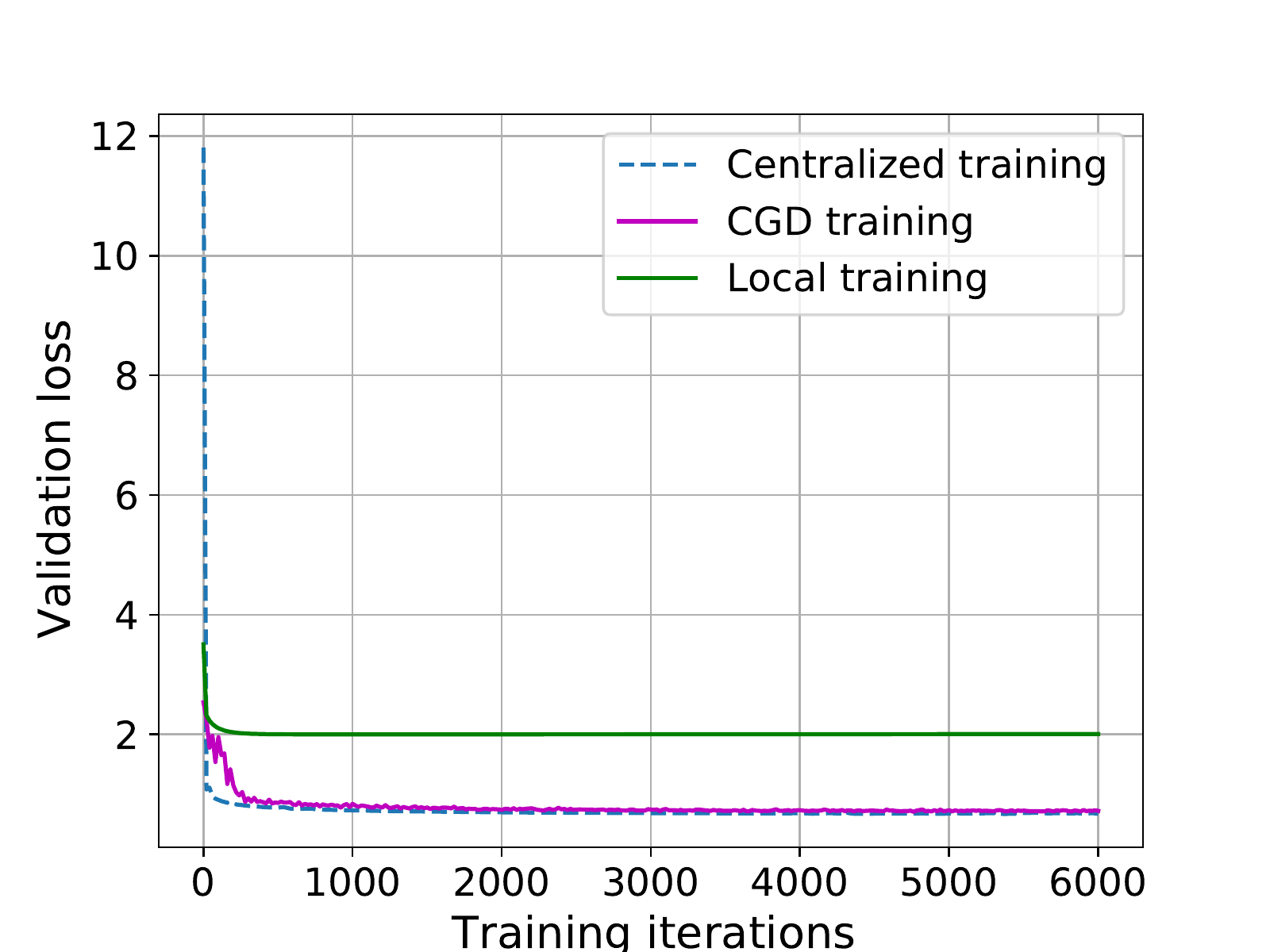} \label{fig:cifar_100_loss}}}%
	\subfloat[$m=(100\times 16)$]{{\includegraphics[width=4cm]{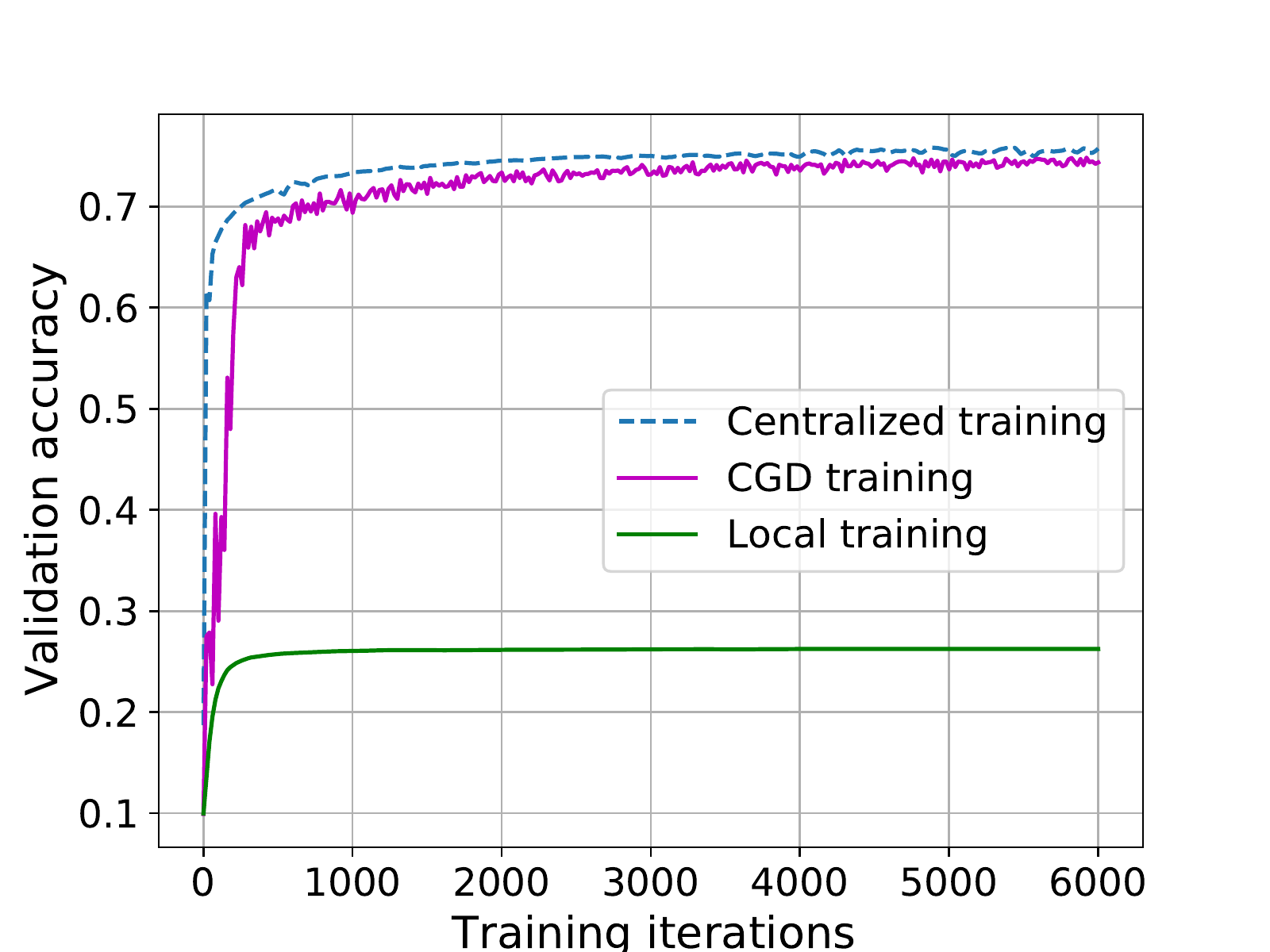} \label{fig:cifar_100_acc}}}%
	\quad
	\subfloat[$m=(1000 \times 32)$]{{\includegraphics[width=4cm]{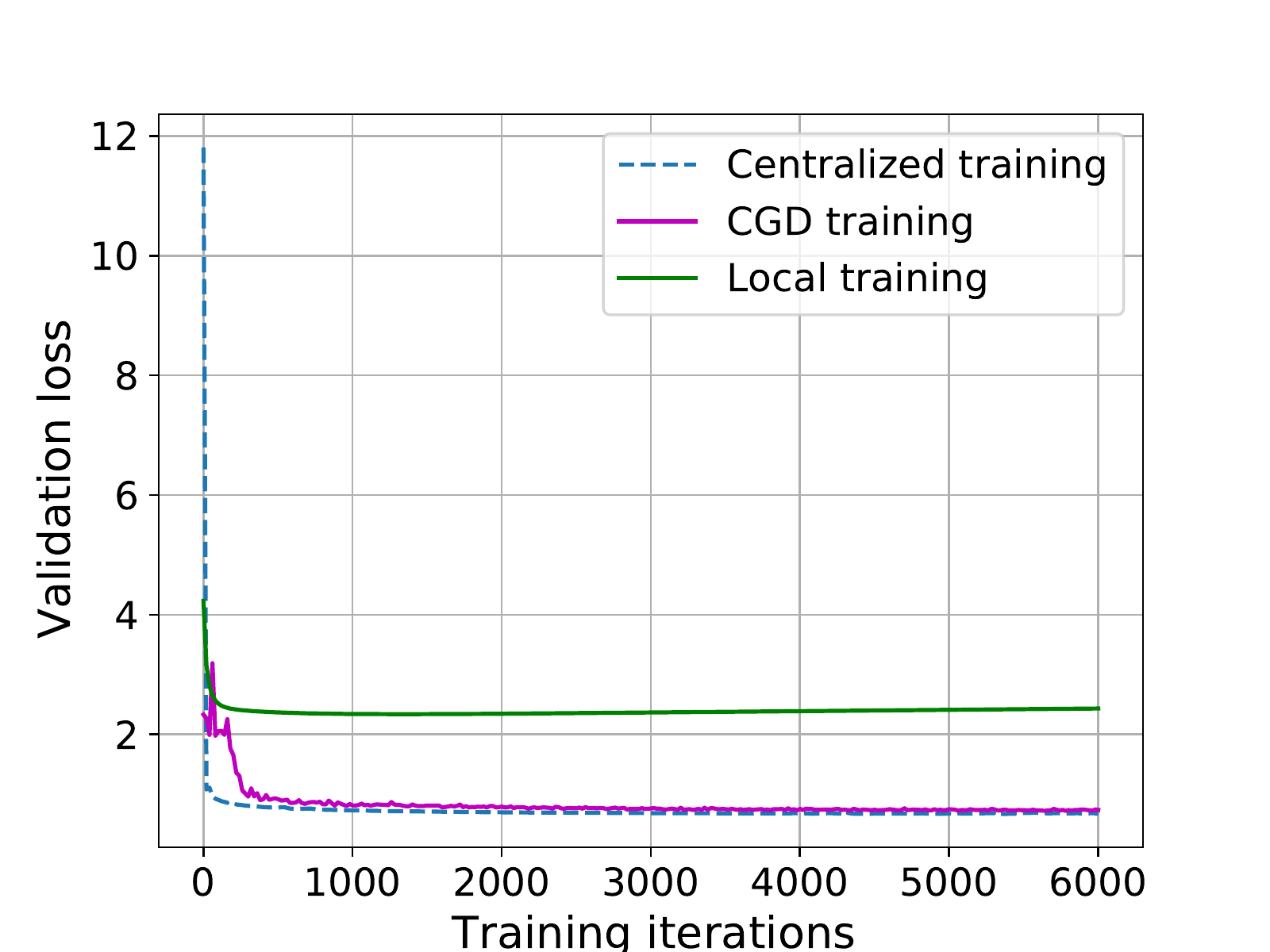} \label{fig:cifar_1000_loss}}}%
	\subfloat[$m=(1000 \times 32)$]{{\includegraphics[width=4cm]{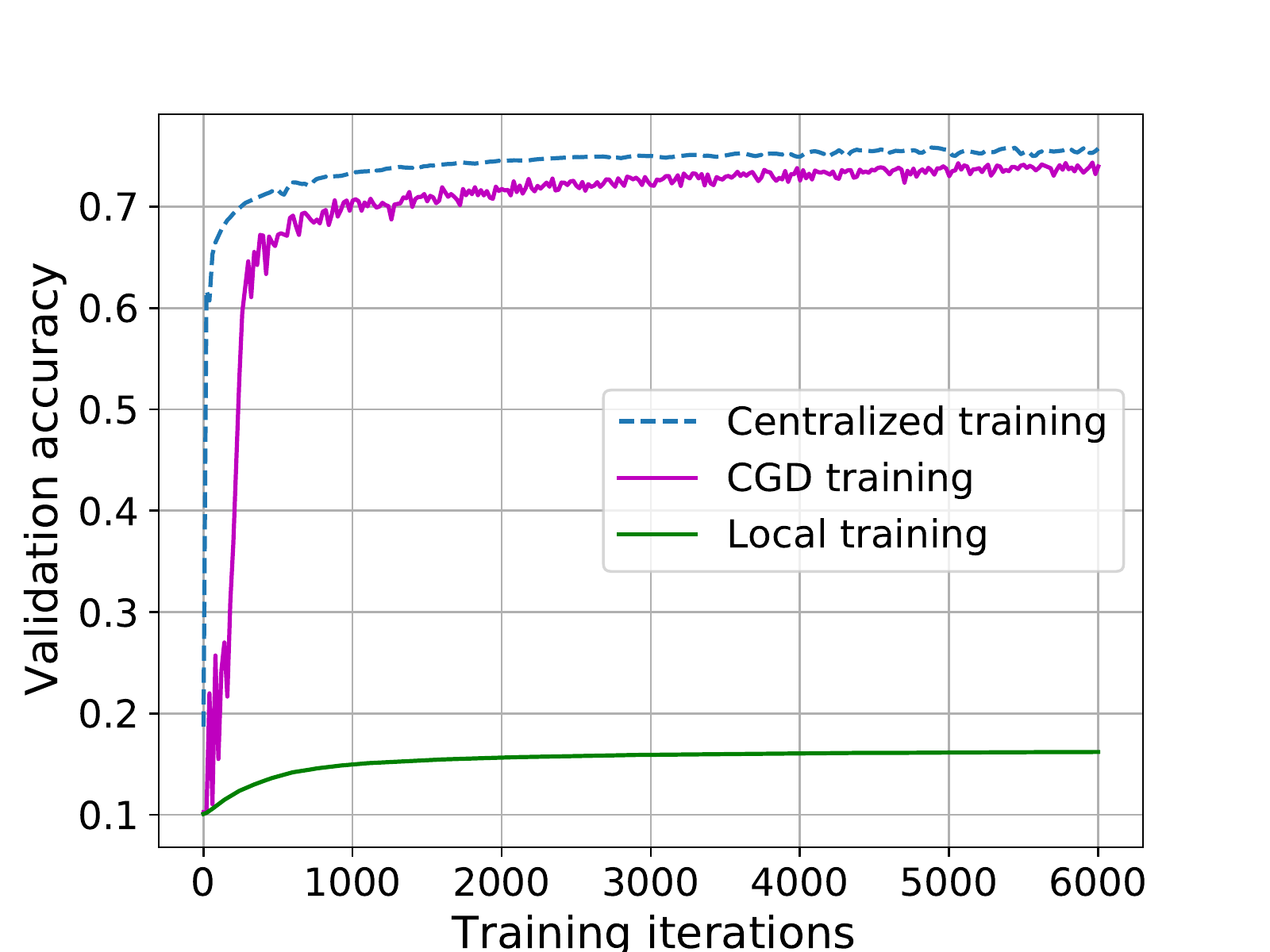} \label{fig:cifar_1000_acc}}}%
	%
	\caption{Results on the validation loss and accuracy for different number of participants on the CIFAR-10 dataset.}%
	\label{fig:cifar}
\end{figure}

Our second set of experiments are conducted on the CIFAR-10 dataset.
It consists of 60000 $32 \times 32$ colour images in 10 classes (e.g., airplane, bird, and cat), with 6000 images per class.
The images are divided into 50000 training images and 10000 test images.
We use the ResNet-56 architecture proposed by He et al.~\cite{he2016deep}.
It takes as input images of size $32 \times 32$, with the per-pixel mean subtracted.
Its first layer is $3 \times 3$ convolutions, and then a stack of $3 \times 18$ layers is used, with $3 \times 3$ convolutions of filter sizes $\{32, 16, 8\}$ respectively, and $18$ layers for each filter size.
The numbers of filters are $\{16, 32, 64\}$ in each stack. 
The network ends with an  average pooling layer, a fully-connected (FC) layer with $1024$ units, and softmax.
Our training data augmentation follows the setting in~\cite{he2016deep}.
For each training image, we generate a new distorted image by randomly flipping the image horizontally with probability $0.5$, and adding 4 amounts of padding 
to all sides before cropping the image to the size of $32 \times 32$.

Since our focus is on evaluating the proposed \short, rather than enhancing the state-of-the-art analysis on CIFAR-10,
we utilize the transferability\footnote{Transfer learning allows the analyst to take a model trained on one dataset and transfer it to another without retraining~\cite{shin2016deep}.} of convolutional layers to save the computational cost of computing per-example gradients.
We follow the experiment setting of Abadi et al.~\cite{abadi2016deep}, which treats the CIFAR-10 as the private dataset and CIFAR-100 as a public dataset.
CIFAR-100 has the same image types as CIFAR-10, and it has 100 classes containing 600 images each.
We use CIFAR-100 to train a network with the aforementioned  architecture, and  freeze the  parameters of the 
convolutional layers and retrain only the last FC layer on CIFAR-10.

Training on the entire dataset with batch update reaches the validation accuracy of $75.75\%$, which is taken as our centralized training baseline\footnote{We note that by making the network deeper or using other advanced techniques, better accuracy can be obtained, with the state-of-the-art being about 99.37\%~\cite{kolesnikov2019big}.}.
For the \short training, each participant feeds the pre-trained convolutional layers with their private data partitions, generates the input features to the FC layer, and randomly initializes the confined model parameters of the FC layer.
They then take the input 
to the FC layer as the private local training data, and train a model using Algorithm \ref{alg:cgd} with $\delta = 0.01$ and $T=6000$.

Figure \ref{fig:cifar} and Table \ref{tab:cifar} summarize our experimental results against centralized training and local training.
The results on the validation loss and accuracy are generally in line with that on the MNIST dataset.
In the worst case of $m = (1000 \times 32)$, \short achieves $0.724$ and $74.29\%$ in validation loss and accuracy respectively, which are relatively near to the centralized baseline~($0.675$ and $75.72\%$).
Table~\ref{tab:dpcifar} summarize the results against DP-FL. In line with the results on MNIST, the accuracy difference to the centralized baseline in \short is smaller than that in DP-FL.

\begin{table}
	\caption{Performance comparison on CIFAR-10.}
	\label{tab:cifar}
	\resizebox{0.45\textwidth}{!}{	
	\begin{tabular}{|l|l|l|l|l|}
		\hline
		& \multicolumn{2}{c|}{Validation loss} & \multicolumn{2}{c|}{Validation accuracy} \\ \hline
		Centralized         & \multicolumn{2}{c|}{0.675}           & \multicolumn{2}{c|}{75.72\%}             \\ \hline \hline
		& \short             & Local training           & \short               & Local training              \\ \hline
		m=($10 \times 2$)   & 0.720             & 0.980            & 74.79\%              & 65.3\%               \\ \hline
		m=($100 \times 16$)  & 0.723             & 2.008            & 74.42\%             & 26.25\%            \\ \hline
		m=($1000 \times 32$) & 0.724             & 2.434            & 74.29\%             & 16.19\%            \\ \hline
	\end{tabular}
}
\end{table}


\begin{table}
	\centering
	\caption{Comparison between \short and the traditional FL on CIFAR-10.}
	\label{tab:dpcifar}
	\resizebox{0.35\textwidth}{!}{
		\begin{tabular}{|l|l|}
			\hline
			\multicolumn{2}{|c|}{\short}                         \\ \hline
			Participants & \shortstack[l]{Difference to the baseline}  \\ \hline
			m=($10 \times 2$)            & 0.93\%                                           \\ \hline
			m=($100 \times 16$)           & 1.30\%                                         \\ \hline
			m=($1000 \times 32$)            & 1.43\%                                           \\ \hline\hline
			\multicolumn{2}{|c|}{Traditional FL via DP~\cite{abadi2016deep}}             \\ \hline
			Noise levels                                & \shortstack[l]{Difference to the baseline$^1$} \\ \hline
			$\epsilon = 8$ (small noise)   & 7\%                      \\ \hline
			$\epsilon = 4$ (medium noise)  & 10\%                      \\ \hline
			$\epsilon = 2$ (large noise) & 13\%                      \\ \hline
		\end{tabular}
	}
	\footnotesize{\qquad\qquad\qquad\qquad\qquad$^1$ The centralized baseline of validation accuracy on the CIFAR-10 in ~\cite{abadi2016deep} is 80\%.}
\end{table}

\subsection{Influencing Factors of \short} \label{subsec:factors}
In this section, we study the influence of initialization and the parameter $\mu$ on \short performance.
This study is conducted with the MNIST dataset. 

\subsubsection{Role of the initialization} \label{sec:init}
According to Theorem \ref{theorem:main}, the gap between \short solution and the centralized model
is bounded by  $\epsilon = m \| \displaystyle \mathop{\mathbb{E}}_{j\in m}(w_1^l-w_1^j)\|$, 
in which each  $w_1^l$ is decided by $\delta$ in Equation \ref{equ:delta}. Therefore, we investigate how the initialization setting affects \short's performance.

To this end, we conduct two experiments in which we mutate parameter $\delta$ while keeping other settings unchanging~(i.e., $\mu = 0$ and $T=2000$).
In the first experiment, we let all participants use the same $\delta$ in $\{0.1, 0.06, 0.01, 0.001\}$, as they are around $\frac{1}{\sqrt{60000}}$~(recall that $60000$ is the sample size of MNIST). 
In the other experiment, we let each participant randomly select its own $\delta$ based on the uniform distribution within the range from $0.001$ to $0.1$.

\begin{figure}[]%
	\centering
	\subfloat[Validation loss]{{\includegraphics[width=4.5cm]{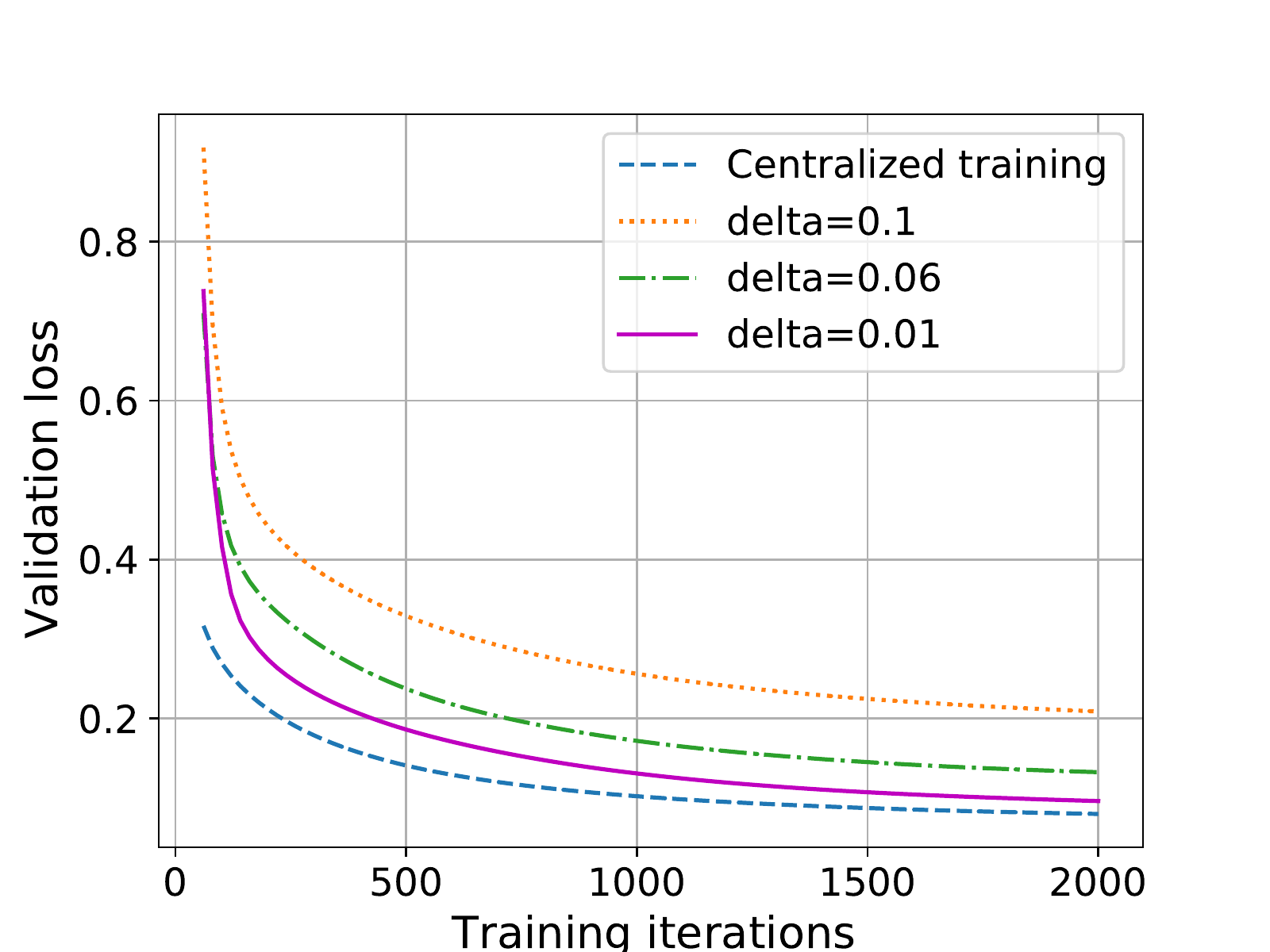} \label{fig:Init_1000_loss}}}%
	\subfloat[Validation accuracy]{{\includegraphics[width=4.5cm]{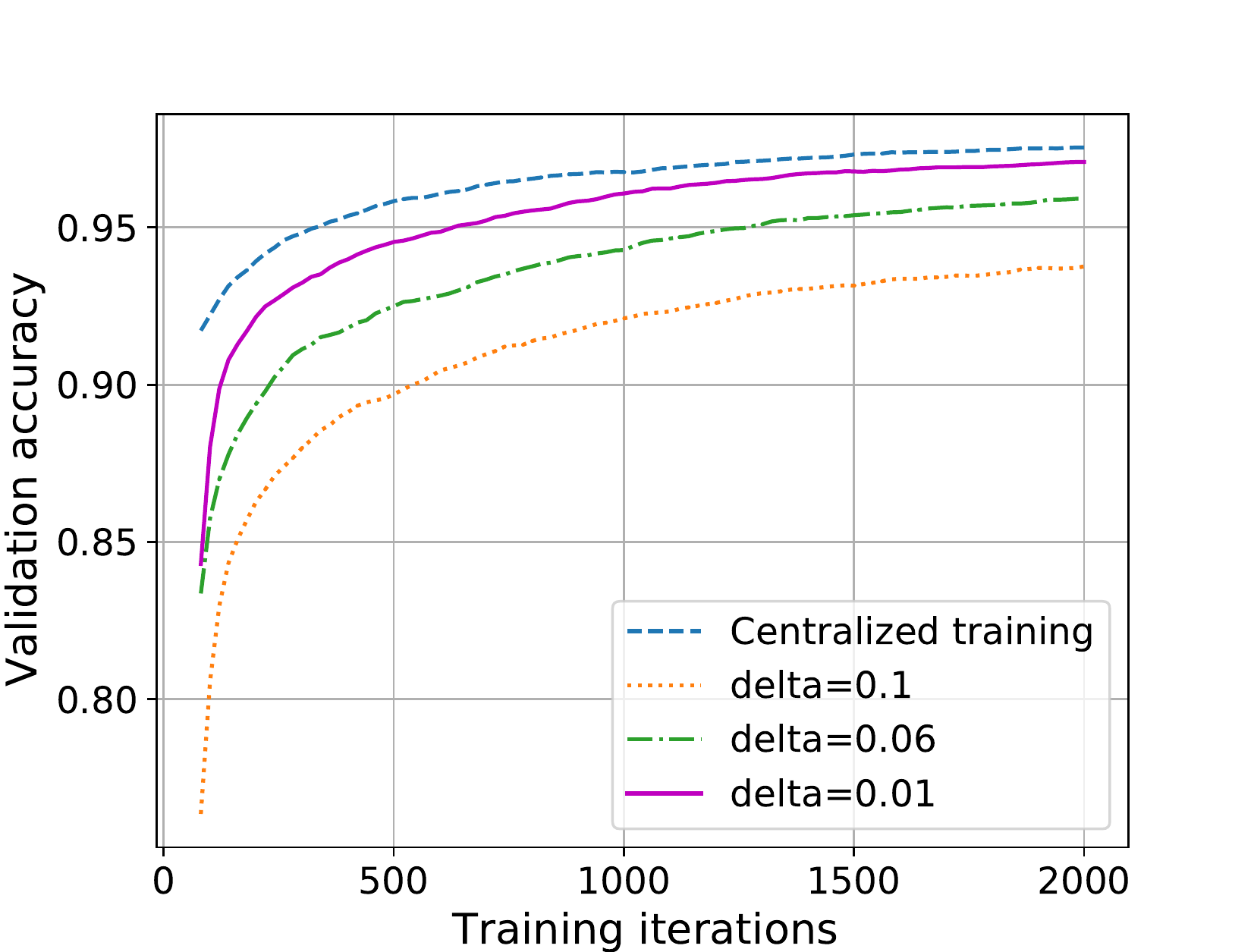} \label{fig:Init_1000_acc}}}%
	%
	\caption{Results on the validation loss and accuracy for different initialization parameter $\delta$ with $m = (1000 \times 112)$ on the MNIST dataset.}%
	\label{fig:init}
\end{figure}

The results of our first experiment are shown in Figure~\ref{fig:init} and the first five columns in Table~\ref{tab:init}.
In general, as $\delta$ decreases, \short achieves better performance.
This confirms our expectation: decreasing $\delta$ would reduce the value of $ \| \displaystyle \mathop{\mathbb{E}}_{j\in m^h}(w_1^l-w_1^j)\|$, such that the confined models are closer to the centralized optimum.
We have not observed significant difference from $\delta=0.1$, $0.06$ and $0.01$.
In the case of $m = (1000 \times 112)$, the validation accuracy and loss achieve $97.08\%$ and $0.096$ with $\delta = 0.01$ which outperform $93.74\%$ and $0.209$ in the setting of $\delta = 0.1$, but both are close to the centralized model whose validation accuracy and loss are $97.61\%$ and $0.078$ respectively.
However, $\delta$ cannot be set too small, in order to maintain numerical stability in neural network~\cite{glorot2010understanding}.
For example, when we lower $\delta$ to $0.001$, the performance starts decreasing.

\begin{figure}[]%
	\centering
	\subfloat[Validation loss]{{\includegraphics[width=4.5cm]{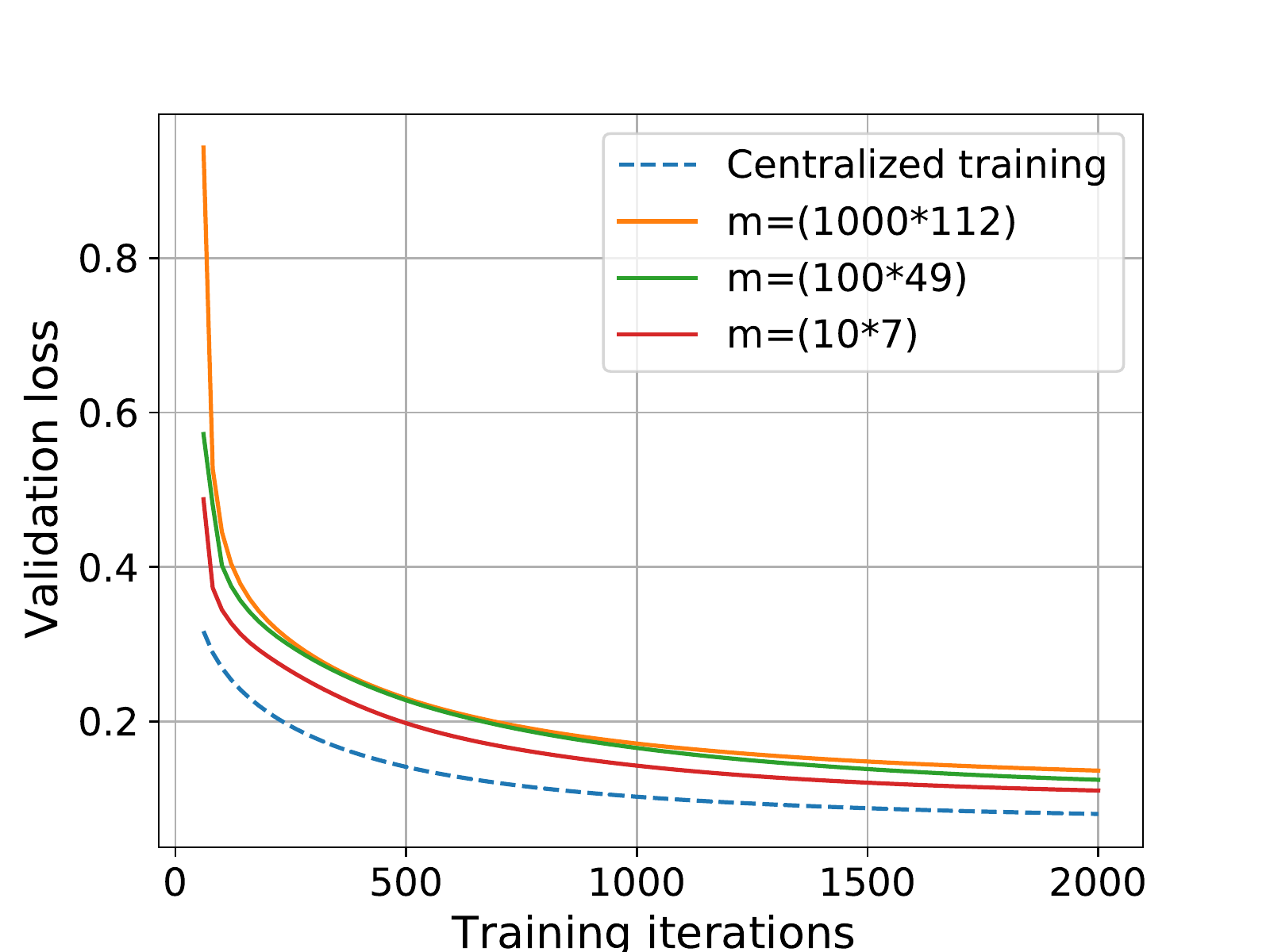} \label{fig:Init_random_loss}}}%
	\subfloat[Validation accuracy]{{\includegraphics[width=4.5cm]{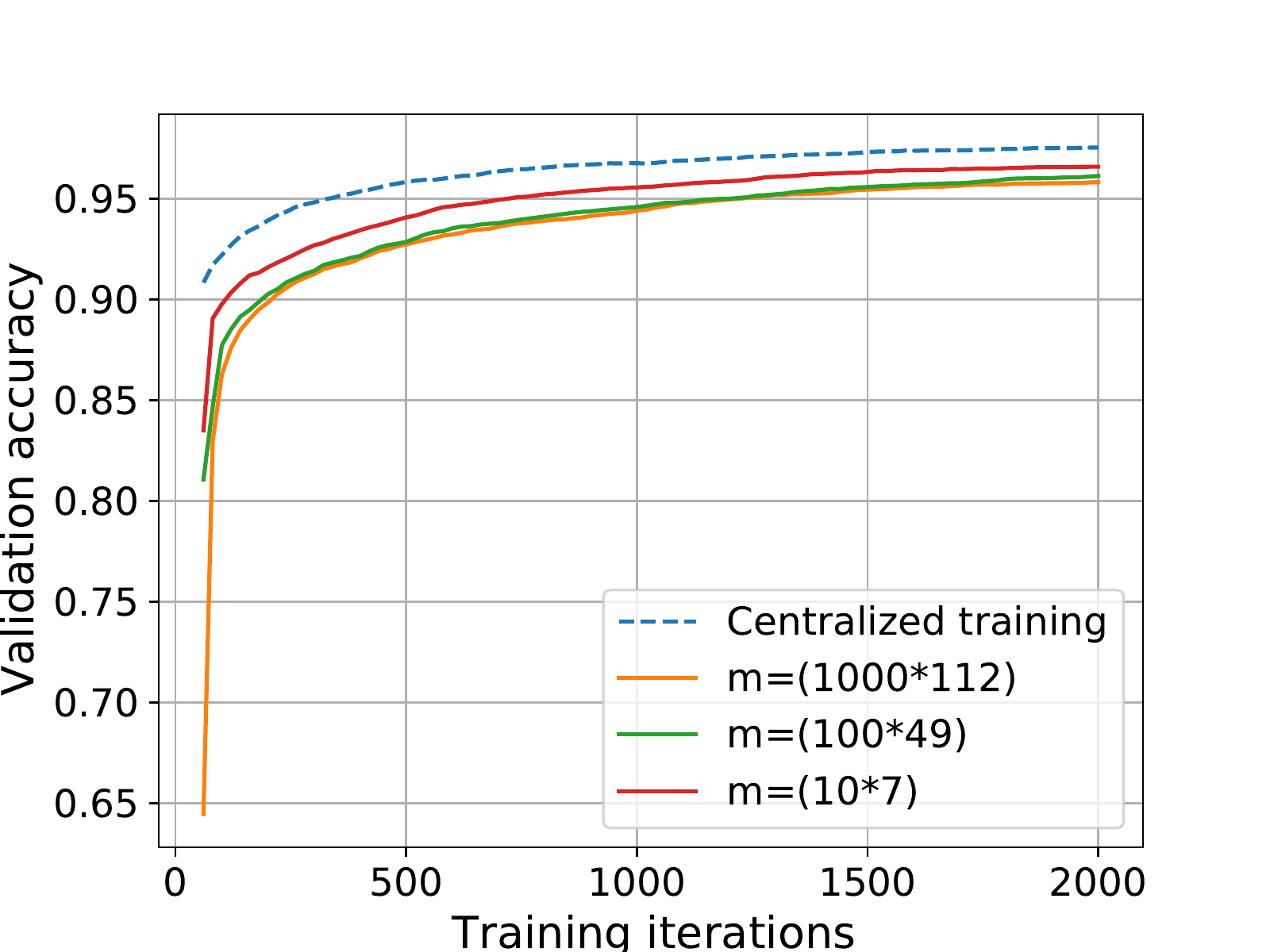} \label{fig:Init_random_acc}}}%
	%
	\caption{Results on the validation loss and accuracy for various number of participants with $\delta$ selected uniformly at random on the MNIST dataset.}%
	\label{fig:initrandom}
\end{figure}

The results of our second experiment are shown in Figure \ref{fig:initrandom} and the last column in Table~\ref{tab:init}.
When the participants uniformly randomize their $\delta$s from $0.001$ to $0.1$,
the performance of \short is still comparable with the centralized mode.
For example, with $m = (1000 \times 112)$ participants, \short achieves $95.82\%$ and $0.136$ in the validation accuracy and loss.
This suggests that \short keeps robust when the $\delta$s of its participants differ by two orders of magnitude.

\begin{table}[t]
	\centering
	\caption{Performance with different initialization parameter $\delta$ for different number of participants on the MNIST dataset.}
	\label{tab:init}
	\resizebox{0.43\textwidth}{!}{
	\begin{tabular}{|l|l|l|l|l||l|}
		\hline
		\multicolumn{6}{|c|}{Validation loss}                                                            \\ \hline
		& $\delta=0.1$   & $\delta=0.06$  & $\delta=0.01$  & $\delta=0.001$     & \multirow{2}{*}{Random}                      \\ \cline{1-5}
		Centralized & 0.080 & 0.074  & 0.078  & 0.111      &                    \\ \hline
		m=($10 \times 7$)    & 0.143 & 0.106 & 0.092 & 0.121      & 0.110                   \\ \hline
		m=($100 \times 49$)   & 0.188 & 0.125 & 0.094 & 0.129     & 0.124                    \\ \hline
		m=($1000 \times 112$)   & 0.209 & 0.132 & 0.096 & 0.117 & 0.136\\ \hline\hline
		\multicolumn{6}{|c|}{Validation accuracy}                                                        \\ \hline
		& $\delta=0.1$   & $\delta=0.06$  & $\delta=0.01$  & $\delta=0.001$     & \multirow{2}{*}{Random}                      \\ \cline{1-5}
		Centralized & 97.54\%    & 97.79\%    & 97.61\%    & 96.69\%      &                       \\ \hline
		m=($10 \times 7$)    & 95.60\%    & 96.72\%    & 97.21\%    & 96.38\%       & 96.59\%                     \\ \hline
		m=($100 \times 49$)   & 94.22\%    & 96.05\%    & 97.15\%    & 96.19\%     & 96.13\%                       \\ \hline
		m=($1000 \times 112$)   & 93.74\%    & 95.91\%    & 97.08\%    & 96.59\%     & 95.82\%                       \\ \hline
	\end{tabular}
}
\end{table}

\subsubsection{Effect of the parameter $\mu$} \label{sec:mu}
According to Theorem \ref{theorem:main}, the convergence speed of \short is affected by the parameter $\mu$.
We thus conduct an experiment to investigate this relation.
In this experiment, we tune $\mu$ while keeping $\delta$ fixed as $0.1$. 
For the stepsize~($\alpha_k$), we run \short with a fixed $\alpha$ till it (approximately) reaches a preferred point, and then continue our experiment with $\alpha_1=\alpha$.
This is to keep \short practical, as each time the stepsize is diminished, more iterations are required.
Therefore, the first $6000$ iterations is run with $\alpha= 0.01$, 
and then diminish $\alpha_k$ in each of the following $8000$ iterations.

\begin{figure}[]%
	\centering
	\subfloat[m=($100 \times 49$)]{{\includegraphics[width=5cm]{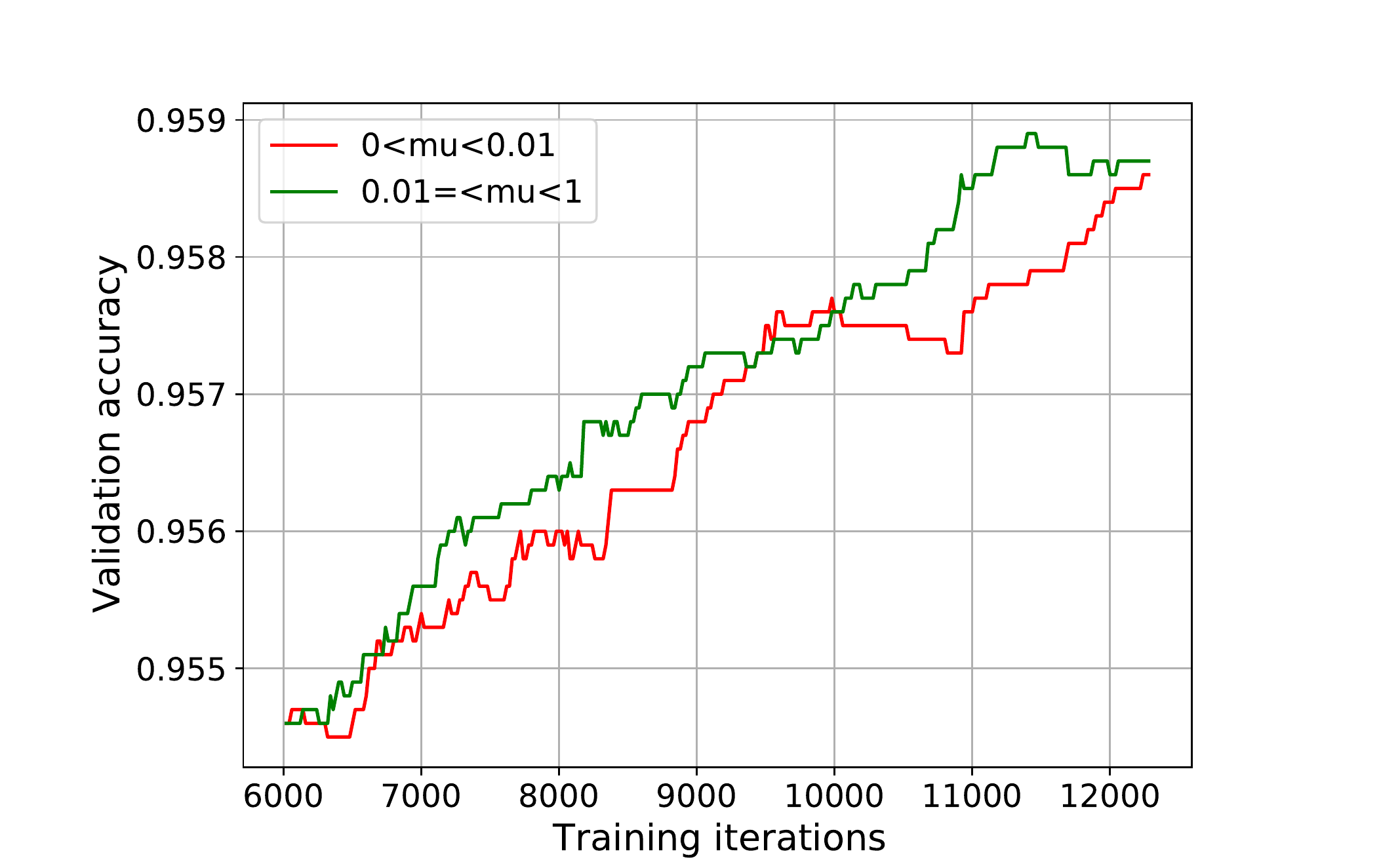} \label{fig:mu_100_acc}}}%
	\subfloat[m=($1000 \times 112$)]{{\includegraphics[width=4.5cm]{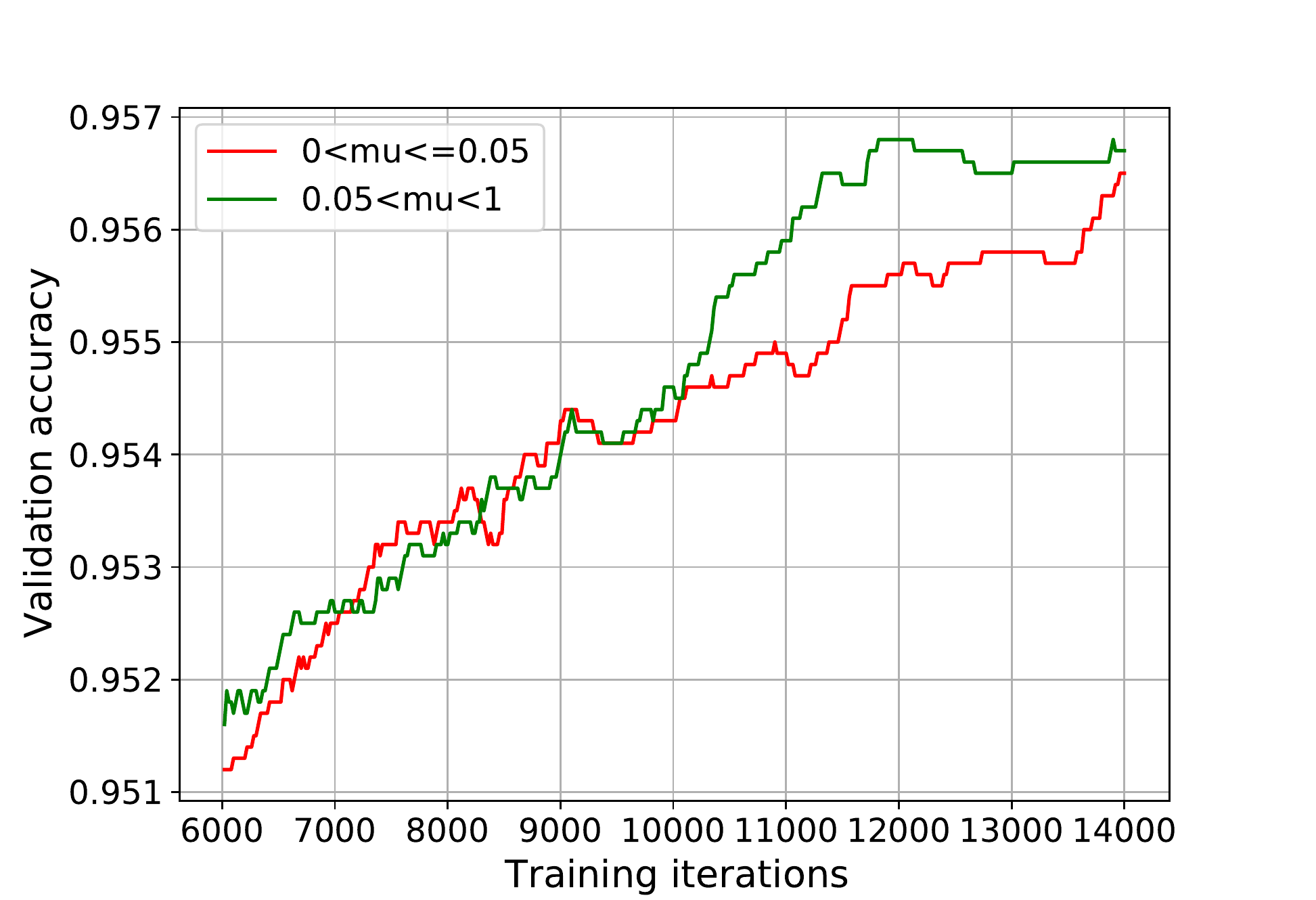} \label{fig:mu_1000_acc}}}%
	%
	\caption{Validation accuracy with different parameter $\mu$ for different number of participants on the MNIST dataset.}%
	\label{fig:mu}
\end{figure}

Figure \ref{fig:mu} demonstrates the results with varying $\mu$ for $m=(100 \times 49)$ and $m=(1000 \times 112)$.
For the case of $m=(100 \times 49)$, $\mu \geq 0.01$ gives a slightly faster speed.
\short reaches the validation accuracy of $95.9\%$ at $11401^{th}$ iteration, $2340$ iterations~(16.7\% less) faster than training with $\mu < 0.01$.
For the case of $m=1000 \times 112)$, $\mu > 0.05$ gives a faster speed.
\short reaches the validation accuracy $95.68\%$ at $11821^{th}$ iteration, $2140$ iterations~(15.28\% less) faster than training with $\mu \leq 0.05$.
Even though such slight difference is observed, our experiment suggests that the effect of tuning $\mu$ is relatively limited. The gain of the validation accuracy is only within $0.2\%$ in the cost of around $2000$ iterations.


\section{Conclusion}
We have presented \short, a novel optimization algorithm for learning confined models to enhance privacy for federated learning.
Privacy preservation is achieved against an honest-but-curious adversary even though the majority ($m-2$ out of $m$) of participants are corrupted. 
We formally proved the convergence of \short optimization. In our experiments, we achieved  validation accuracy of 97.01\% with ($1000 \times 112$) participants on the MNIST dataset, and 74.4\%   with ($1000 \times 32$) participants on the CIFAR-10 dataset.  Both are comparable to the performance of centralized training.  

A number of future work directions are of interest. In particular, we see new research opportunities in applying our techniques to asynchronous FL, allowing different participants to be at different iterations of model updates up to a bounded delay. We are also considering other types of deep networks, for example, unsupervised neural networks such as autoencoders. 

\bibliographystyle{ACM-Reference-Format}
\bibliography{CCS21}

\begin{appendices}
	

	\section{Additive secret sharing scheme}
	\label{apx:secret}
	Secret sharing schemes aim to securely distribute secret values amongst a group of participants. \short employs the secret sharing scheme proposed by \cite{bogdanov2008sharemind}, which uses \emph{additive sharing} over $\mathbb{Z}_{2^{32}}$.
	In this scheme, a secret value $srt$ is split to $s$ shares $E^1_{srt},...,E^s_{srt} \in \mathbb{Z}_{2^{32}}$  such that
	\begin{equation}
		E^1_{srt}+E^2_{srt}+...+E^s_{srt} \equiv srt \mod 2^{32},
	\end{equation}
	and any $s-1$ elements $E^{i_1}_{srt},...,E^{i_{s-1}}_{srt}$ are uniformly distributed.
	This prevents any participant who has part of the shares from deriving the value of $srt$, unless all participants join their shares.
	
	In addition, the scheme has a homomorphic property that allows efficient and secure addition on a set of secret values $srt_1,...,srt_s$ held by corresponding participants $S_1,...,S_s$.
	To do this, each participant $S_i$ executes a randomised sharing algorithm $Shr(srt_i, S)$ to split its secret $srt_i$ into shares $E^{1}_{srt_i},...,E^{s}_{srt_i}$, and distributes each $E^{j}_{srt_i}$ to the participant $S_j$.
	Then, each $S_i$  locally adds the shares it holds $E^{i}_{srt_1},...,E^{i}_{srt_s}$ and produces $\sum_{j=1}^sE^{i}_{srt_j}$ (denoted by $E^i$ for brevity). After that, a reconstruction algorithm $Rec(\{(E^{i}, S_i)\}_{S_i \in S})$, which takes $E^i$ from each participant and add them together, can be executed by an aggregator to reconstruct the $\sum_{i=1}^s srt_i$ without revealing any secret addends $srt_i$.
	
	\section{Simulation Paradigm}
	\label{apx:sim}
	In simulation paradigm (a.k.a., the real/ideal model)~\cite{goldreich2019play}, the security of a protocol is proved by comparing what an adversary can do in a real protocol execution   to what it can do in an ideal scenario, which is secure by definition. 
	Formally, a protocol $\mathcal{P}$ securely computes a functionality $\mathcal{F}_p$, if for every adversary $\mathcal{A}$ in the real model, there exists an adversary $\mathcal{S}$ in the ideal model, such that the view of the adversary from a real execution $\mathit{VIEW_{real}}$ is indistinguishable from the view of the adversary from an ideal execution $\mathit{VIEW_{ideal}}$. 
	The adversary $\mathcal{S}$ in the ideal model, is called the simulator. 
	An indistinguishability between $\mathit{VIEW_{real}}$ and $\mathit{VIEW_{ideal}}$ guarantees that the adversary can learn nothing more than their own inputs and the information required by $\mathcal{S}$  for the simulation. 
	In other words, the information required by $\mathcal{S}$ for the simulation is the only information that can leak to adversary $\mathcal{A}$ from the real execution.
	
	Let $c$ denote the set of corrupted parties. The simulator performs the following operations:
\begin{itemize}
	\item Generate dummy inputs $\{\eta^l\}$  for each honest party $l \notin c$ and receives the actual inputs $\{x^l\}$  of corrupted parties $l \in c$;
	\item Run $\mathcal{P}$ over $\{\eta^l\}$ ($l \notin c$ ) and $\{x^l\}$ ($l \in c$) and add all messages sent/received by corrupted parties to $\mathit{VIEW_{ideal}}$;
	\item Send the inputs of corrupted parties $\{x^l\}$ ($l \in c$)  to the trusted third party;
	\item Receive the outputs of corrupted users $\{y^l\}$ ($l \in c$) from the trusted third party and add them to $\mathit{VIEW_{ideal}}$.
\end{itemize}	
	
	Meanwhile, a real instance of $\mathcal{P}$ is executed with actual inputs for all parties, and $\mathit{VIEW_{real}}$ is created by gathering inputs of corrupted parties, messages sent/received by corrupted parties during the protocol and their final outputs. Once the simulation is finished, the security is proved by showing that $\mathit{VIEW_{ideal}}$ is indistinguishable from $\mathit{VIEW_{real}}$.

	
	\section{Supplementary Experimental Results}
	\label{apx:experiment}
	\begin{figure}[H]%
		\centering
		\subfloat[$m=(10 \times 1)$]{{\includegraphics[width=4cm]{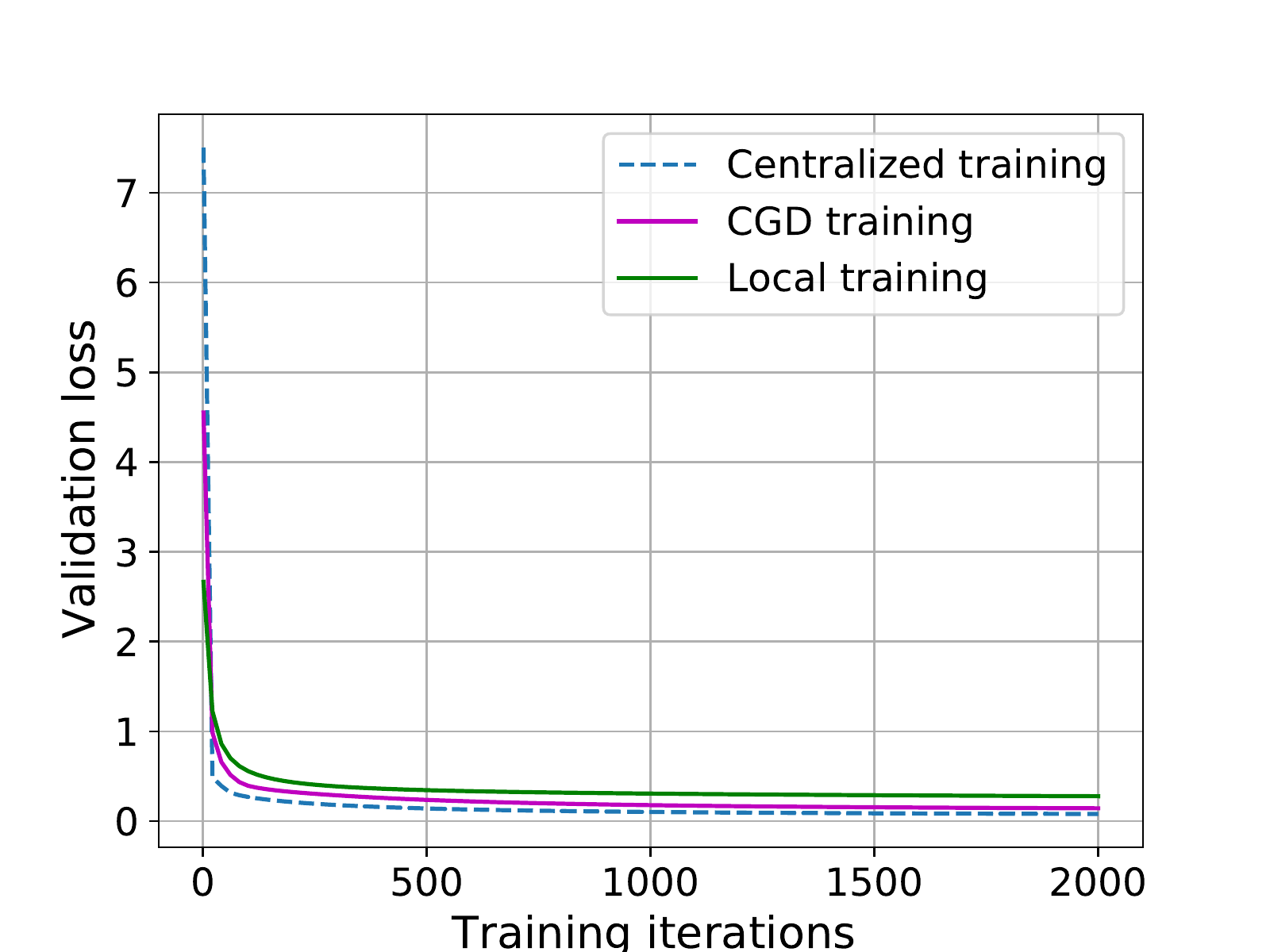} \label{fig:fixrate_10*1_loss}}}%
		\subfloat[$m=(10 \times 1)$]{{\includegraphics[width=4cm]{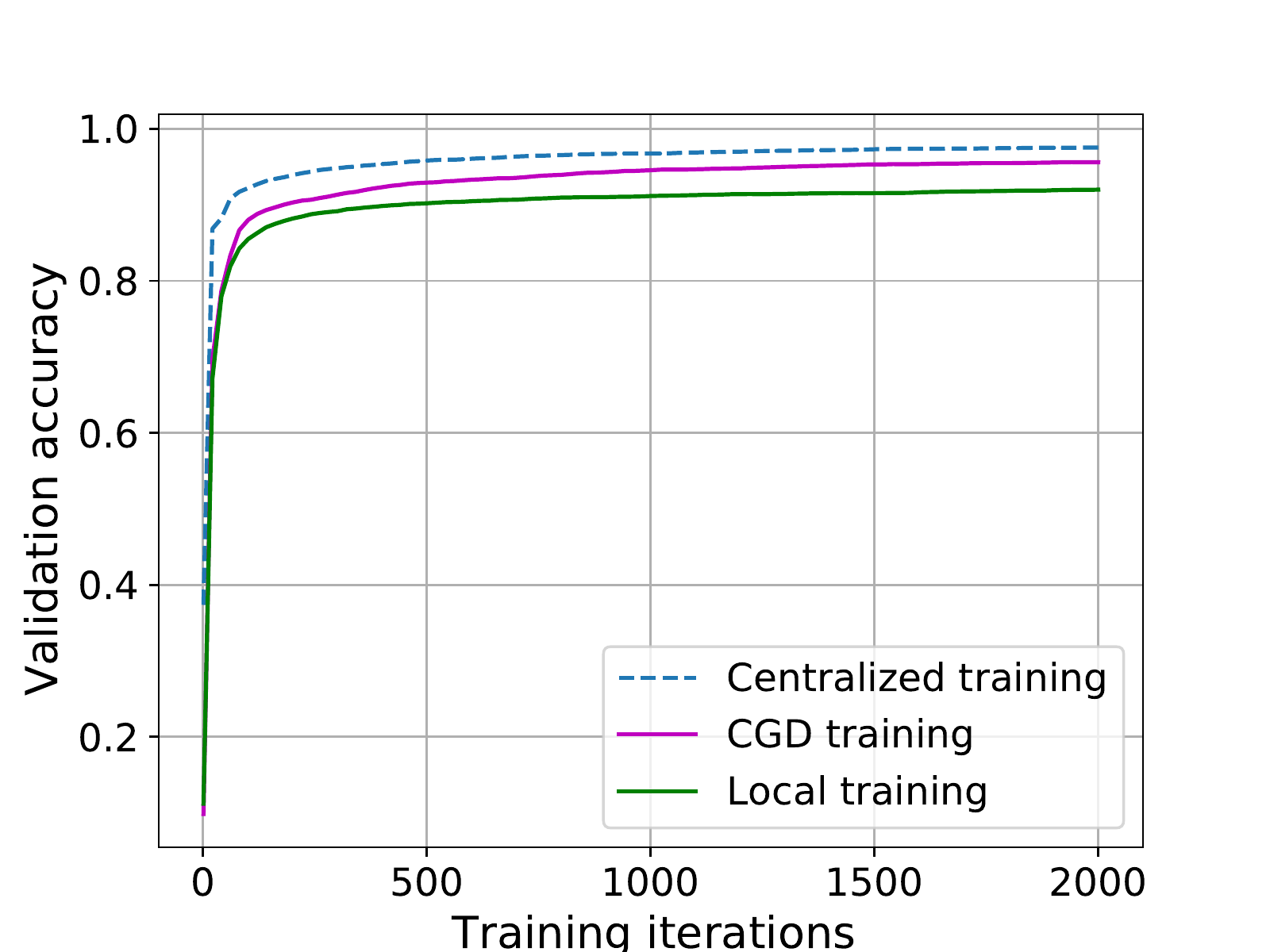} \label{fig:fixrate_10*1_acc}}}%
		\quad
		\subfloat[$m=(100 \times 1)$]{{\includegraphics[width=4cm]{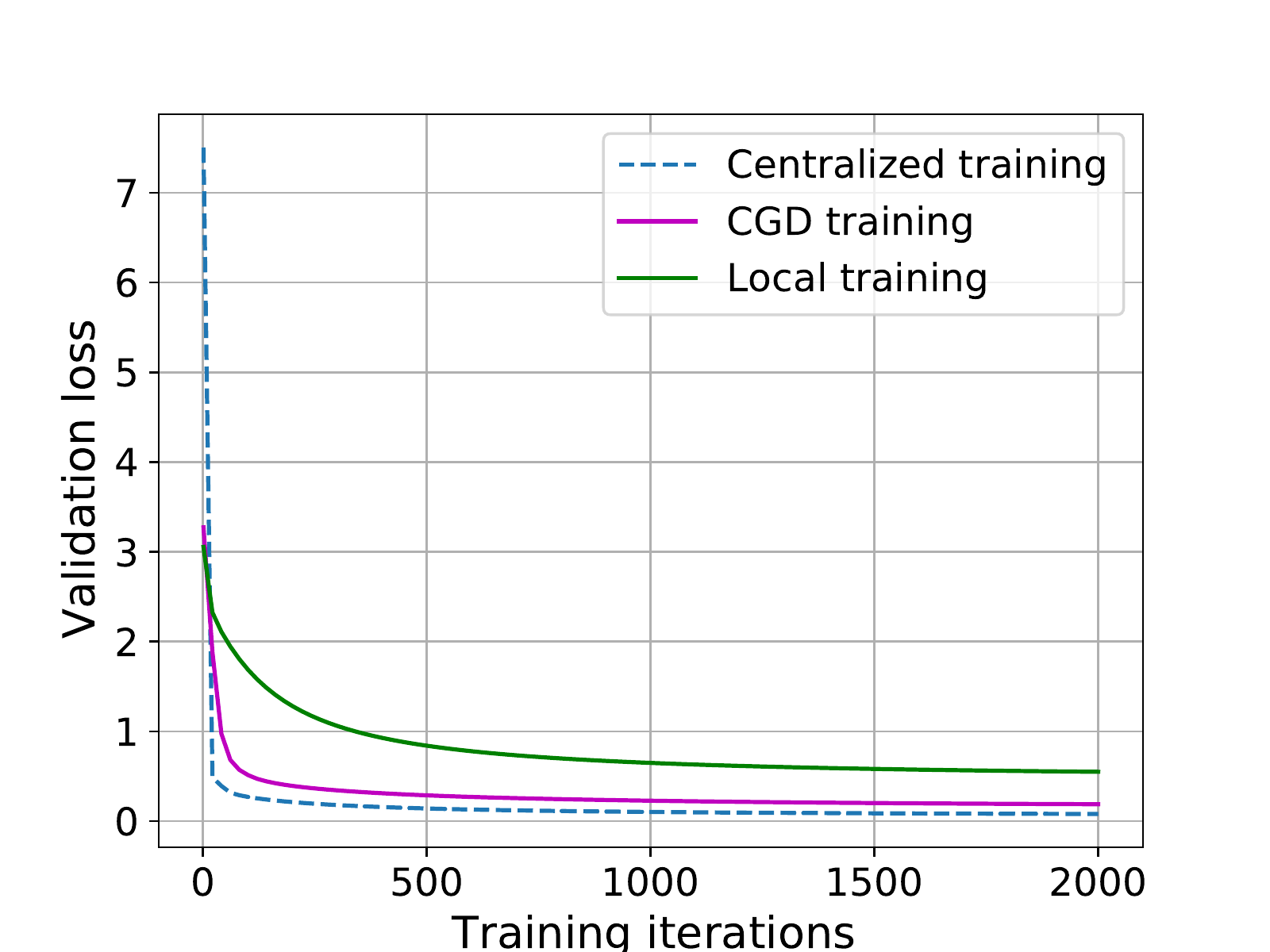} \label{fig:fixrate_100*1_loss}}}%
		\subfloat[$m=(100 \times 1)$]{{\includegraphics[width=4cm]{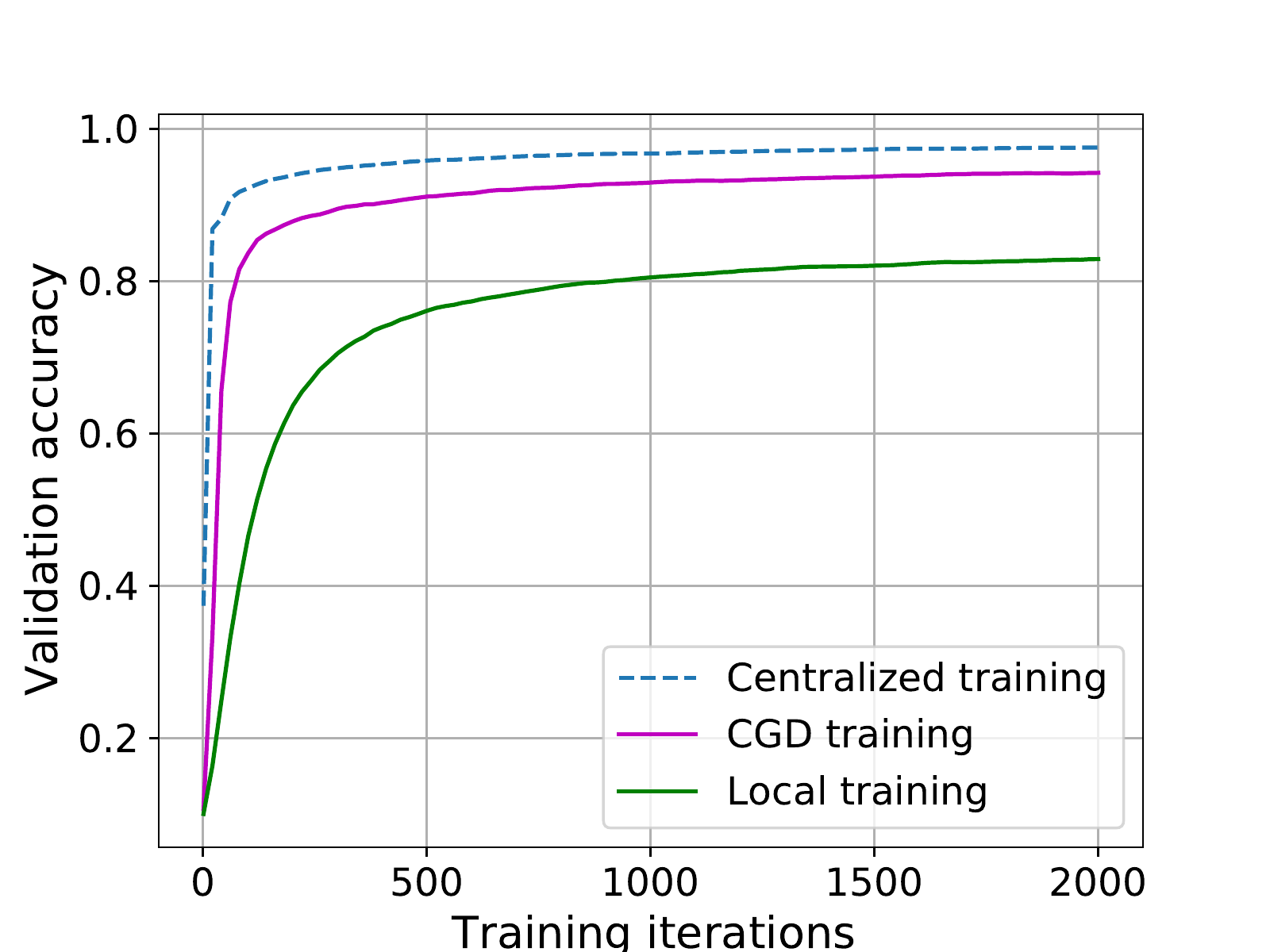} \label{fig:fixrate_100*1_acc}}}%
		\quad
		\subfloat[$m=(1000 \times 1)$]{{\includegraphics[width=4cm]{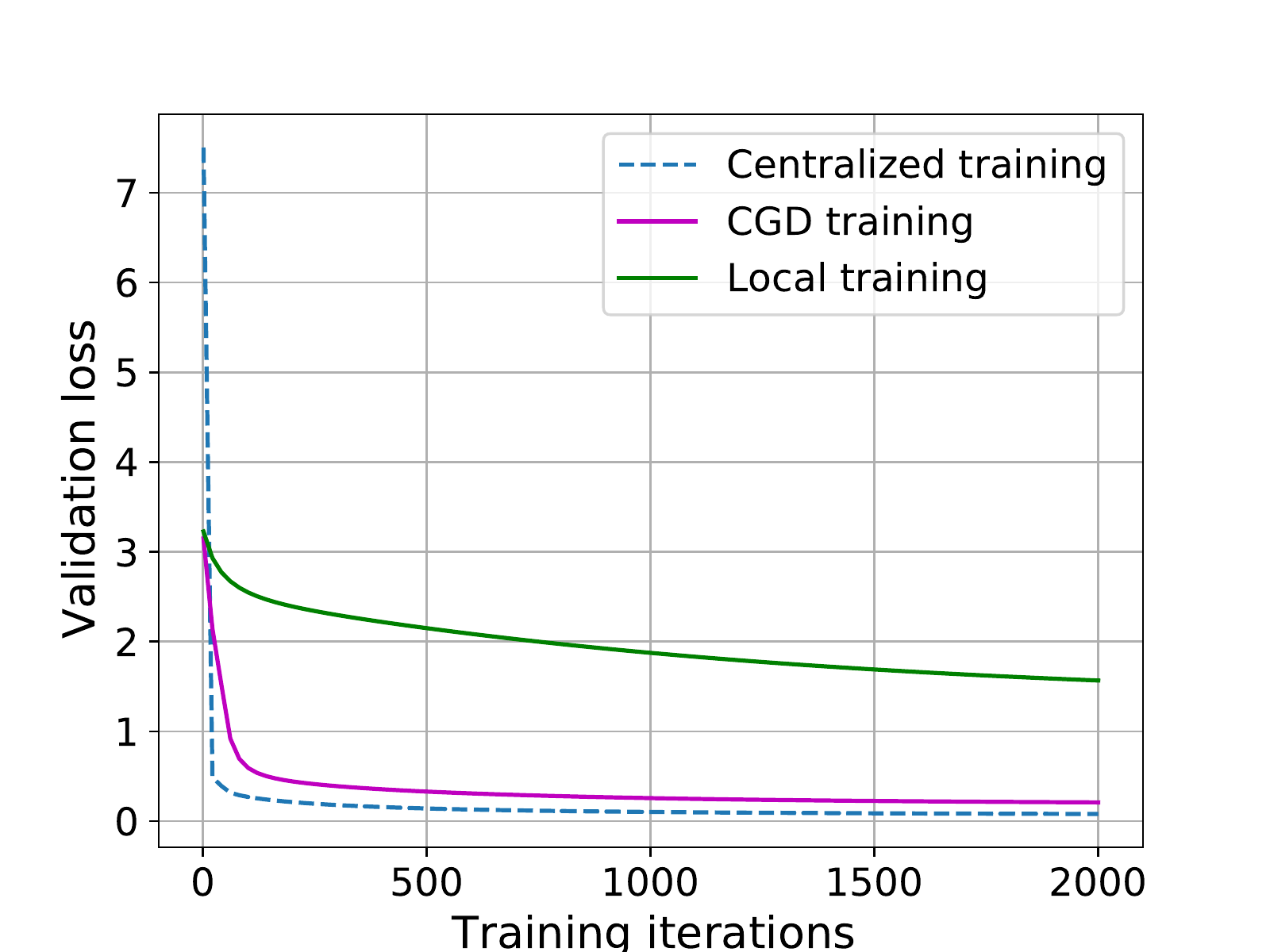} \label{fig:fixrate_1000*1_loss}}}%
		\subfloat[$m=(1000 \times 1)$]{{\includegraphics[width=4cm]{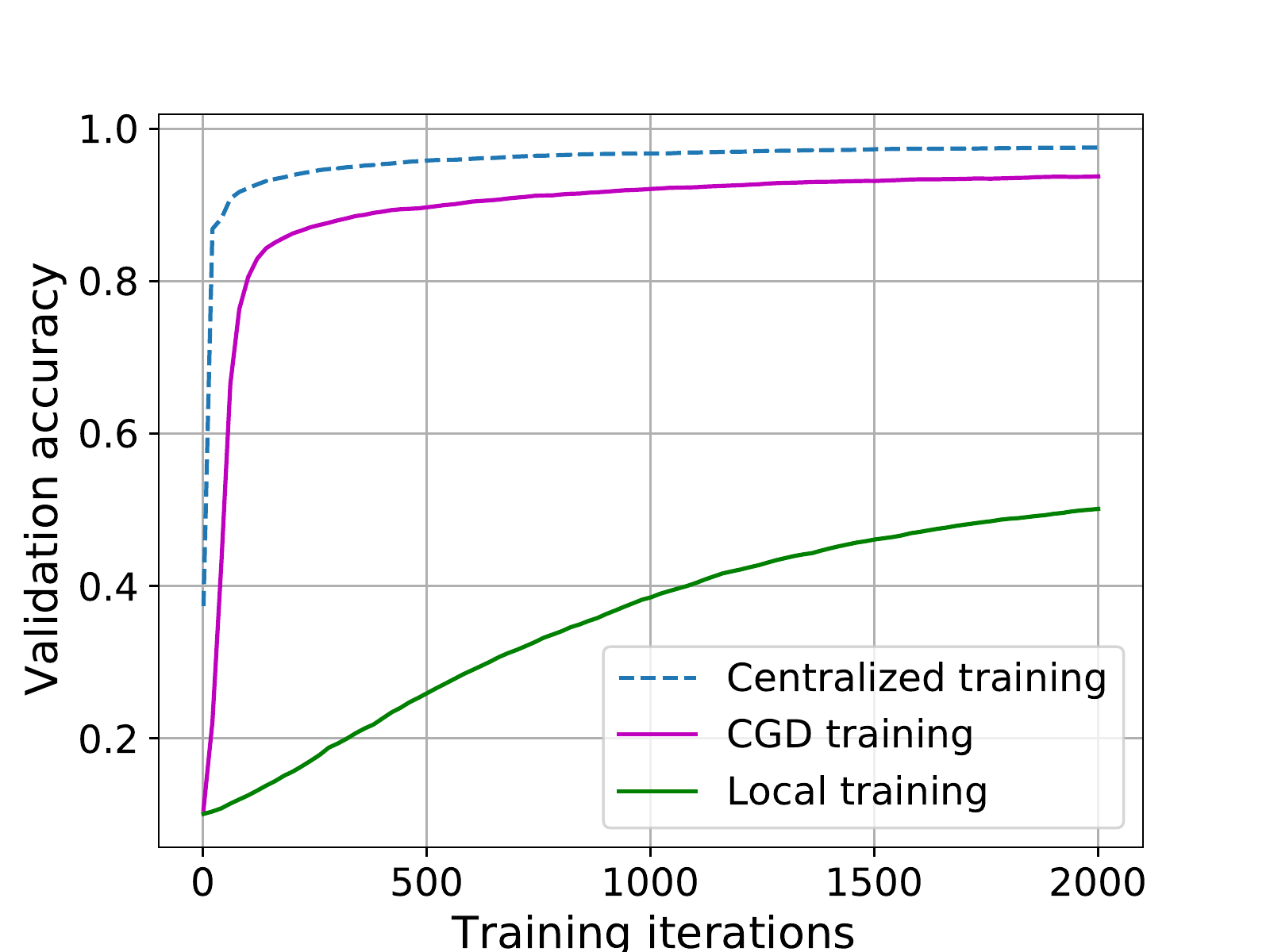} \label{fig:fixrate_1000*1_acc}}}%
		%
		\caption{Supplementary experimental results on the validation loss and accuracy for different number of participants on the MNIST dataset in the default setting.}%
		\label{fig:fixratesup}
	\end{figure}
	
\begin{table}[H]
	\caption{Supplementary experimental results on the validation loss and accuracy for different number of participants on the MNIST dataset in the default setting.} 
	\label{tab:fixratesup}
	\begin{tabular}{|l|l|l|l|l|}
		\hline
		& \multicolumn{2}{c|}{Validation loss} & \multicolumn{2}{c|}{Validation accuracy} \\ \hline
		Centralized        & \multicolumn{2}{c|}{0.081}           & \multicolumn{2}{c|}{97.54\%}             \\ \hline \hline
		& \short             & Local training           & \short               & Local training              \\ \hline
		m=($10 \times 1$)   & 0.143             & 0.278            & 95.62\%              & 92.01\%               \\ \hline
		m=($100 \times 1$)  & 0.188             & 0.549            & 94.22\%             & 82.89\%            \\ \hline
		m=($1000 \times 1$) & 0.208             & 1.566            & 93.76\%             & 50.11\%            \\ \hline
	\end{tabular}
\end{table}

\end{appendices}

\end{document}